\documentclass{article}

\usepackage{arxiv}

\usepackage[utf8]{inputenc} % allow utf-8 input
\usepackage[T1]{fontenc}    % use 8-bit T1 fonts
\usepackage{hyperref}       % hyperlinks
\usepackage{url}            % simple URL typesetting
\usepackage{booktabs}       % professional-quality tables
\usepackage{amsfonts}       % blackboard math symbols
\usepackage{nicefrac}       % compact symbols for 1/2, etc.
\usepackage{microtype}      % microtypography
\usepackage{lipsum}         % Can be removed after putting your text content
\usepackage{graphicx}
\usepackage{doi}
\usepackage{appendix}

\usepackage[T1]{fontenc}    % Font encoding
\usepackage{csquotes}       % Include csquotes

\title{
Interpretable and efficient data-driven \\ Discovery and Control of distributed systems %Andrea's proposal
}

\newif\ifuniqueAffiliation
\uniqueAffiliationtrue

\ifuniqueAffiliation % Standard variant of author block

\author{ \href{https://orcid.org/0009-0003-0627-7051}{
  \includegraphics[scale=0.06]{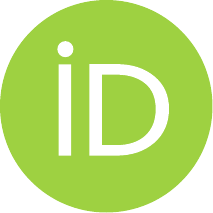}\hspace{1mm}Florian Wolf
    }\thanks{Work carried out while visiting MOX-Department of Mathematics, Politecnico di Milano, Italy.} \\
  Department of Mathematics\\
  Technical University of Darmstadt\\
  Darmstadt, Germany\\
	\texttt{mail@florian-wolf.cc} \\
	\And
	\href{https://orcid.org/0000-0003-3676-3619}{
    \includegraphics[scale=0.06]{figures/orcid.pdf}\hspace{1mm}Nicolò Botteghi} \\
    Department of Applied Mathematics\\
    University of Twente\\
    Enschede, The Netherlands\\
	\texttt{n.botteghi@utwente.nl} \\
	\And
	\href{https://orcid.org/0000-0002-3716-686X}{
    \includegraphics[scale=0.06]{figures/orcid.pdf}\hspace{1mm}Urban Fasel} \\
    Department of Aeronautics\\
    Imperial College London\\
    London, United Kingdom\\
	\texttt{u.fasel@imperial.ac.uk} \\
	\And
	\href{https://orcid.org/0000-0001-8277-2802}{
    \includegraphics[scale=0.06]{figures/orcid.pdf}\hspace{1mm}Andrea Manzoni} \\
    MOX -- Department of Mathematics\\
    Politecnico di Milano\\
    Milan, Italy\\
	\texttt{andrea1.manzoni@polimi.it}
}
\else
\usepackage{authblk}

\newbox{\orcid}\sbox{\orcid}{\includegraphics[scale=0.06]{orcid.pdf}} 
\author[1]{%
	\href{https://orcid.org/0000-0000-0000-0000}{\usebox{\orcid}\hspace{1mm}David S.~Hippocampus\thanks{\texttt{hippo@cs.cranberry-lemon.edu}}}%
}
\author[1,2]{%
	\href{https://orcid.org/0000-0000-0000-0000}{\usebox{\orcid}\hspace{1mm}Elias D.~Striatum\thanks{\texttt{stariate@ee.mount-sheikh.edu}}}%
}
\affil[1]{Department of Computer Science, Cranberry-Lemon University, Pittsburgh, PA 15213}
\affil[2]{Department of Electrical Engineering, Mount-Sheikh University, Santa Narimana, Levand}
\fi

\renewcommand{\shorttitle}{Wolf et al. AE+SINDy-C}

\usepackage{xcolor}
\definecolor{darkblue}{rgb}{0,0,.5}
\definecolor{darkred}{rgb}{0.5,0,0} % maybe 0.6
\definecolor{darkorange}{rgb}{0.8, 0.4, 0}

\hypersetup{
  pdftitle={Interpretable and Efficient Data-driven Discovery and Control 
  of Distributed Systems },
  pdfsubject={Model-based Reinforcement Learning},
  pdfauthor={Florian Wolf, Nicolò Botteghi, Urban Fasel, Andrea Manzoni},
  pdfkeywords={Reinforcement Learning, Sparse Identification of
  	Nonlinear Dynamics, Autoencoder},
    colorlinks=true,          % Enables colored links
    linkcolor=darkblue,       % Color for internal references (e.g., \ref, \pageref)
    citecolor=darkblue,        % Color for citation links (e.g., \cite)
    urlcolor=darkblue         % Color for URLs (e.g., \url)
}

\usepackage[backend=biber, style=alphabetic, citestyle=alphabetic-verb]{biblatex}
\bibliography{references_AE_SINDy_Wolf_Manzoni}

\usepackage{enumitem}
\usepackage{tikz}
\usetikzlibrary{shapes.geometric, positioning, calc, fit, backgrounds, shapes.symbols}
\usepackage{xcolor}
  
\definecolor{darkRed}{RGB}{92, 0, 29}
\definecolor{lightRed}{RGB}{186, 49, 51}
\definecolor{darkBlue}{RGB}{32, 72, 127}
\definecolor{lightBlue}{RGB}{157, 202, 224}
\definecolor{superLightBlue}{RGB}{214, 231, 240}

\newcommand{\lightRed}[1]{\textcolor{lightRed}{#1}}
\newcommand{\darkBlue}[1]{\textcolor{darkBlue}{#1}}

\usepackage{algorithm}
\usepackage{algpseudocode}

\usepackage{amsmath}
\usepackage{mathrsfs}
\usepackage{amsfonts}

\usepackage{graphicx}
\usepackage{subcaption}
\usepackage{booktabs}
\usepackage{caption} 
\usepackage{cleveref}

\newcommand{\norm}[1]{\left\lVert#1\right\rVert}
\def\RR{{\mathbb{R}}}
\def\NN{{\mathbb{N}}}
\def\EE{{\mathbb{E}}}
\def\phi{{\varphi}}
\def\epsilon{{\varepsilon}}
\def\kDyn{{k_{\text{dyn}}}}

\algrenewcommand \algorithmicensure{\textbf{Output:}}

\renewcommand{\vec}[1]{\boldsymbol{\mathrm{#1}}}

\newcommand{\dimX}{{N_x}}
\newcommand{\dimObsX}{{N_x^{\mathsf{Obs}}}}
\newcommand{\dimRedX}{{N_x^{\mathsf{Lat}}}}

\newcommand{\dimU}{{N_u}}
\newcommand{\dimObsU}{{N_u}}
\newcommand{\dimRedU}{{N_u^{\mathsf{Lat}}}}

\newcommand{\observed}{\mathsf{Obs}}
\newcommand{\latent}{\mathsf{Lat}}

\DeclareMathOperator*{\argmin}{arg\,min}

\newcommand{\AeSindy}{\mbox{AE+SINDy}}
\newcommand{\SindyC}{\mbox{SINDy-C}}
\newcommand{\ourAlgo}{\mbox{AE+SINDy-C}}
\newcommand{\rlLib}{\mbox{\emph{Ray RLLib}}}

\newcommand*\dd{\: \mathrm{d}}

\def\Uniform{{\mathcal{U}([-1,1]^\dimX)}}

\begin{document}
\maketitle

\begin{abstract}
Effectively controlling systems governed by Partial Differential Equations (PDEs) is crucial in several fields of Applied Sciences and Engineering. These systems usually yield significant challenges to conventional control schemes due
to their nonlinear dynamics, partial observability, high-dimensionality once discretized, distributed nature, and the requirement for low-latency feedback control. Reinforcement Learning (RL),
particularly Deep RL (DRL), has recently emerged as a promising control paradigm for such systems,
demonstrating exceptional capabilities in managing high-dimensional, nonlinear dynamics.
However, DRL faces challenges including sample inefficiency, robustness issues, and an overall lack of interpretability.
To address these issues, we propose a data-efficient, interpretable, and scalable Dyna-style
Model-Based RL framework for PDE control, combining the Sparse Identification of Nonlinear
Dynamics with Control (\SindyC{}) algorithm and an autoencoder (AE) framework for the sake of dimensionality reduction of PDE states and actions. This novel approach enables fast rollouts, reducing the need for extensive environment interactions, and provides an
interpretable latent space representation of the PDE forward dynamics.
We validate our method on two PDE problems describing fluid flows -- namely, the 1D Burgers equation and 2D Navier-Stokes equations -- comparing it against a model-free baseline, and carrying out an extensive analysis of the learned dynamics.
%\\\\\emph{The code for \ourAlgo{} will be made publicly available upon acceptance.}
\end{abstract}

\keywords{Model-Based Deep Reinforcement Learning \and Sparse Identification of Nonlinear Dynamics \and Autoencoder}

\noindent\textbf{Highlights}

The novel \ourAlgo{} framework:
\begin{itemize}
	\item enables data-efficient control of high-dimensional discretized PDEs by combining \SindyC{} with an autoencoder framework for dimensionality reduction of states and actions;
	\item is validated in the problem of controlling nonlinear time-dependent PDEs in partially and fully observable setting and it is compared with a model-free baseline;
	\item provides interpretable dynamics
	with closed-form representation in the low-dimensional surrogate space that is thoroughly analyzed.
\end{itemize}

%\newpage
%\tableofcontents
\newpage

\section{Introduction}

Feedback control for complex physical systems is essential in many fields of Engineering and Applied Sciences, 
which are typically governed by Partial Differential Equations (PDEs). In these cases, the state of the systems is often challenging
or even impossible to observe completely, the systems exhibit nonlinear dynamics, and require low-latency feedback
control \cite{Brunton_2020, Peitz_2020, BrownianMot2020}.
Consequently, effectively controlling these systems is a computationally intensive task. For instance, significant efforts have been devoted in the last decade to the investigation of optimal control problems governed by PDEs
\cite{hinze2008optimization, manzoni2022optimal}; however, classical feedback control strategies face limitations with such highly complex dynamical systems. For instance, (nonlinear) \emph{model predictive control} (MPC) \cite{NonlinearMPC2017} has emerged as an effective and important control paradigm. MPC utilizes an internal model of the dynamics to create a feedback loop and provide optimal controls, resulting in a difficult trade-off between model accuracy and computational performance. Despite its impressive success in disciplines such as robotics
\cite{williams2018information} and controlling PDEs \cite{MPCPDE2014},
MPC struggles with real-time applicability in providing low-latency actuation,
due to the need for solving complex optimization problems.

In recent years, \emph{reinforcement learning} (RL), particularly \emph{deep reinforcement learning}
(DRL) \cite{Sutton1998}, an extension of RL relying on \emph{deep neural networks} (DNN), has gained
popularity as a powerful and real-time applicable control paradigm. Especially in the context of solving
PDEs, DRL has demonstrated outstanding capabilities in controlling complex and high-dimensional dynamical
systems at low latency \cite{yousif_optimizing_2023, peitz_distributed_2023, botteghi_parametric_2024,
FlowPDEsVinuesa2024}.
Additionally, recent community contributions of benchmark environments such as
\emph{ControlGym}\footnote{\url{https://github.com/xiangyuan-zhang/controlgym}}
\cite{zhang_controlgym_2023},
\emph{PDEControlGym}\footnote{\url{https://github.com/lukebhan/PDEControlGym}} \cite{bhan2024pde},
and \emph{HydroGym}\footnote{\url{https://github.com/dynamicslab/hydrogym}}, all wrapped under the
uniform \emph{Gym} \cite{OpenAIGym2016} interface, highlight the importance of DRL in the context
of controlling PDEs. Despite its impressive and versatile success, DRL faces three major challenges:
\begin{enumerate}
	\item \textbf{Data usage:} DRL algorithms are known to be sample inefficient, requiring a large number of environment interactions, resulting in long training times and high computational requirements \cite{NeuralArchitectureSearchWithDRL2016}. This is particularly problematic in the context of controlling PDEs, as generating training data often requires either long simulations 
	\cite[section B.3]{zolman_sindy-rl_2024} or would necessitate
    very expensive real-world experiments 
    \cite{OpenAIRubicsCube208}.
	\item \textbf{Robustness of the control performance:} In high-dimensional state-action spaces,
	as in the context of (discretized) PDEs, it is challenging to generate sufficient training data
	for the agent to adequately cover the state-action space. This is crucial for ensuring a reliable
	and trustworthy controller, impeding the application to safety-critical use-cases like
	nuclear fusion \cite{FusionControl2018} or wind energy \cite{WindControl2017}.
	\item \textbf{Black-Box model:} DRL methods suffer from a lack of interpretability of the
	learned policy and the underlying (unknown) nonlinear dynamics. In the field of controlling PDEs,
	significant effort is spent to simultaneously understand the dynamical system \cite{botteghi_parametric_2024,
    alla2023onlineidentificationcontrolpdes}.
\end{enumerate}

A popular approach to address the aforementioned problem of 
sample inefficiency is the use of model-based algorithms,
specifically \emph{model-based reinforcement learning} (MBRL)
\cite{SUTTON1990216, Deisenroth2011MBRL, NEURIPS2018_3de568f8, Clavera2018MBRL, BenchmarkingMBRL2019,  Hafner2020MBRL}. One of its
various forms is the \emph{Dyna-style} MBRL algorithm \cite{DynaSutton1991}, which iteratively collects
samples from the full-order environment to learn a surrogate dynamics model. The agent then alternates between
interacting with the surrogate model and the actual environment, significantly reducing the amount of
required training data, allowing faster rollouts. 
Recent contributions in MBRL for PDEs include the use of convolutional long-short term memory with actuation networks % by Werner et al. 
\cite{Werner2023LearningAM} and Bayesian linear regression for
model identification in the context of the State-Dependent Riccati Equation control approach
% by Alla et al. 
\cite{alla2023onlineidentificationcontrolpdes}.

\emph{Sparse dictionary learning} are data-driven methods that seek to approximate a nonlinear function using a sparse linear
combination of candidate dictionary functions, e.g., polynomials of degree $d$ or trigonometric functions. In the context of identification of dynamical systems, we sparse dictionary learning is used by the \emph{sparse identification of nonlinear systems} (SINDy) method (cf. \cref{sec:2BackgroundSindy}). SINDy is a very powerful method to identify a parsimonious, i.e., with a limited number of terms, dynamics model and resolve the issue of lacking interpretability, for example, of deep neural networks. The
\emph{PySINDy}\footnote{\url{https://github.com/dynamicslab/pysindy}} package \cite{kaptanoglu2021pysindy}
makes implementations easy and fast. With its active community, SINDy has been extended to various
forms, such as ensemble versions \cite{Fasel_2022Ensemble},
uncertainty quantification
for SINDy \cite{hirsh2021sparsifyingpriorsbayesianuncertainty},
Bayesian autoencoder-based SINDy
\cite{gao2022bayesianautoencodersdatadrivendiscovery}
% \textcolor{red}{I would add UQ-SINDy, Hirsh et al, 2022 and Bayesian SINDy, Gao et al, 2024}
 and applied to a variety of different
applications, including turbulence closures \cite{SindyClosures2020}, PDEs
\cite{rudy2016datadrivendiscoverypartialdifferential}, continuation methods \cite{conti_reduced_2023} and variational architectures \cite{conti2024venivindyvicivariational}.
%\textcolor{red}{I would add Veni-Vindy-Vici, Conti et al, arxiv 2024}. 
However, SINDy does not scale
to high dimensional systems, which is why 
we learn the latent representation of the high-dimensional states and actions of the PDE with two autoencoders (AEs). This dimensionality-reduction step allows for using SINDy, and
improves the robustness (we learn a low-dimensional representation of the PDE states) of the control policies \cite{lesort2018state, botteghi2022unsupervised}.

%\newpage
\begin{figure}[t!b]
    \centering
    \includegraphics[width=\textwidth]{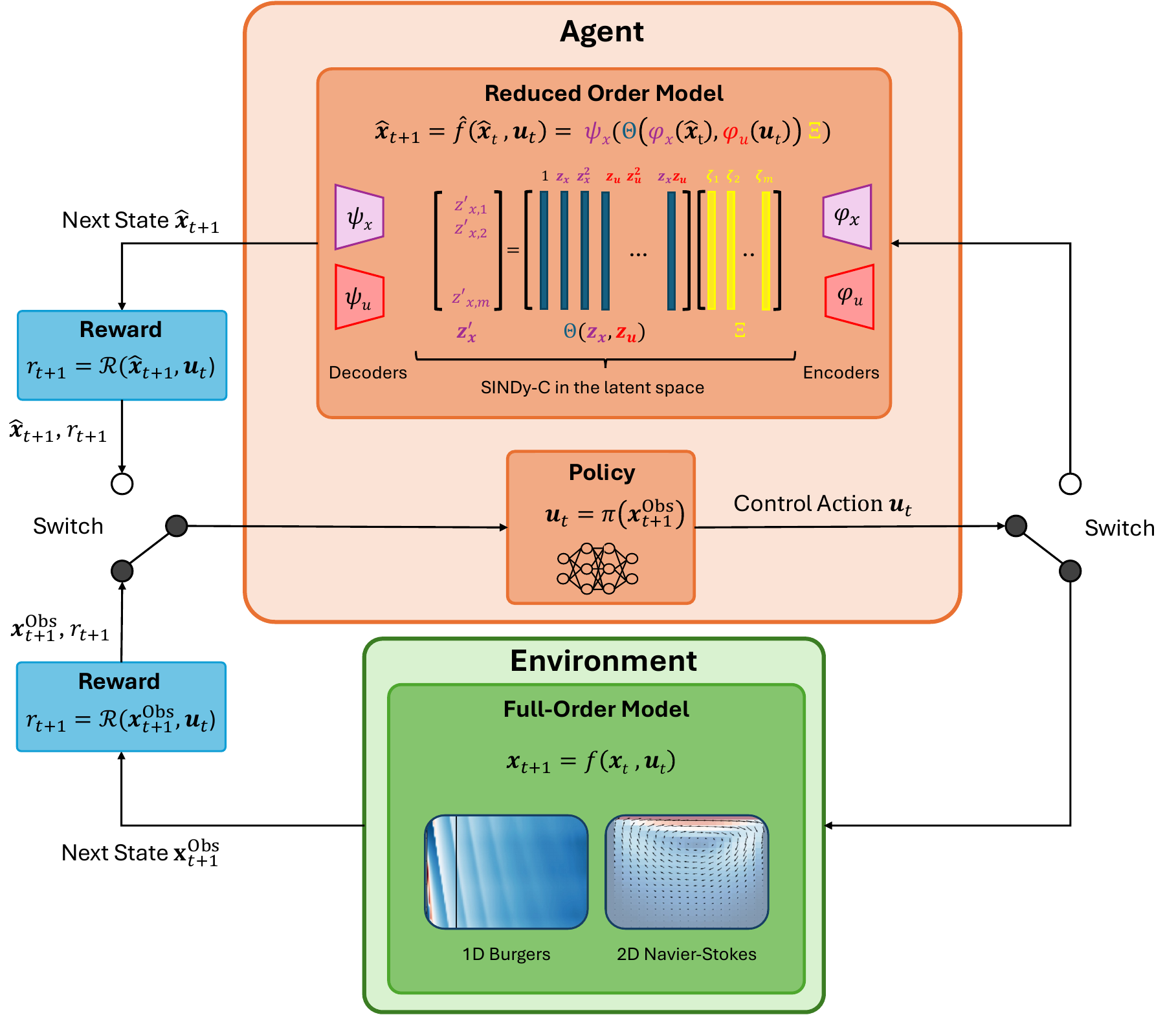}
    \caption{A general overview of the RL training loop. In dyna-style
    algorithms we choose if the agent interacts with the full-order model, requiring
    (expensive) environment rollouts or the learned surrogate, i.e. reduced order, model, providing
    fast approximated rollouts. In this work, we focus on the setting where the full-order reward is (analytically) known and only the dynamics are approximated.
    In general, the observed state is computed by $\RR^{\dimObsX} \ni \vec{x}^{\observed}_{t+1} = C \cdot \vec{x}_{t+1}$.
    In the partially observable (PO) case the projection matrix $C \in \{0,1\}^{\dimX \times \dimObsX}$ is structured with a single 1 per row and zero elsewhere, i.e. $\dimObsX \ll \dimX$. In the fully observable case $C \equiv \mathrm{Id}_{\RR^\dimX}$, i.e. $\dimObsX = \dimX$.} \label{fig:TikzSchemeRLSurrogateModel}
\end{figure}

\paragraph{Our contribution} Our work can be summarized as follows:
\begin{itemize}
	\item With \ourAlgo{}, we present a novel combination of the AE framework and the \SindyC{} algorithm, incorporating controls into the AE framework as a Dyna-style MBRL method for controlling PDEs.
	\item \ourAlgo{} enables fast rollouts, significantly reducing the required number of full-order environment interactions and provides an interpretable, low-dimensional latent %space 
 representation of the dynamical system.
	\item We demonstrate the feasibility of our approach by solving two different fluid-flow PDEs, comparing it to a state-of-the-art model-free baseline algorithm. 
    \item We provide an extensive analysis of the learned
	dynamics and numerical ablation studies.
\end{itemize}

In particular, our work relies on two SINDy \cite{brunton2016sparseidentificationnonlineardynamics} extensions:
\begin{enumerate}
	\item For control applications, SINDy has been extended to work with controls, namely \emph{\SindyC{}},
	MPCs \cite{Kaiser_2018,SINDyCFasel2021}, and importantly, in the context of RL in the work of Arora et al.
	\cite{arora2022modelbasedreinforcementlearningsindy}, and more recently
	by Zolman et al. \cite{zolman_sindy-rl_2024} (see \cref{sec:ZolmanSINDy-RL} for more details).
	\item The work of Champion et al. \cite{champion_data-driven_2019} enables data-driven discovery
	using SINDy and an AE framework for high-dimensional dynamical systems
	(see \cref{sec:ChampionSINDyAE} for more details), which has been extended to the continuation of
	parameter-dependent PDE solutions by Conti et al. \cite{conti_reduced_2023}.
\end{enumerate}

The remainder of this paper is structured as follows: \Cref{sec:Section2Background} provides a brief
overview of the general problem setting, related with control strategies relying on DRL, and the two aforementioned SINDy extensions upon which \ourAlgo{} is based. \Cref{sec:Section3OurAlgo} explains the proposed method in
detail and highlights the usage of \ourAlgo{} in the context of DRL for controlling PDEs.
In \cref{sec:NumericalExperiments}  we explain the two PDE benchmark cases and analyze the numerical results
in detail. \Cref{sec:DiscussionAndFutureWork} provides an overview of possible interesting
directions for follow-up work and variations of \ourAlgo{}. All details about the used environments,
DRL, and the training of \ourAlgo{} can be found in the
\cref{sec:Appendix}.

\section{Background and Related Work}
\label{sec:Section2Background}
\subsection{Problem setting}
In this work, we address the problem of controlling a nonlinear, distributed,
%\textcolor{red}{distributed}, 
dynamical system described by the equation:
\begin{align*}
\frac{\mathrm{d}}{\mathrm{d}t} \vec{x}(t) = f(\vec{x}(t), \vec{u}(t)),
\end{align*}
where the state $\vec{x}(t) \in \RR^\dimX$ and control input $\vec{u}(t) \in \RR^\dimU$ 
are continuous variables, \textcolor{black}{with potentially very large dimensions $\dimX$ and $\dimU$, respectively.}
%\textcolor{red}{We do not address any parametrized system: I would avoid at all to introduce parameters...} 
%and $\vec{\mu} \in \RR^l$ is a vector of system parameters.\footnote{
%    Although potentially variable, in this work we only consider fixed systems parameters, but we refer to \cref{sec:FutureWork} for more details on possible parameter-dependent extensions.} 
The function $f: \RR^\dimX \times \RR^\dimU \times \RR^l \to \RR^\dimX$ is assumed to
be unknown, but we can observe a time-discrete evolution of the system, resulting in a sequence of
(partially or fully) observable measurements $\vec{m}_t, \vec{m}_{t+1}, \ldots,
\vec{m}_{t+H}$ of
the state over a horizon $H \in \NN_+$. A system is fully-observable (FO) if the observations 
$\vec{m}_i \in \RR^\dimX$ allow to observe each (full) state of the system directly, i.e., $\dimObsX = \dimX$ and $\vec{m}_i=\vec{x}_i$. On the other hand,
a system is partially-observable (PO) if only a limited number of sensors is available,
resulting in a lower-dimensional observation space $\vec{m}_i \in \RR^\dimObsX$, where
$\dimObsX \ll \dimX$ that does not capture the full state of the PDE.
In this work, we consider both FO and PO systems. Similarly, we assume a limited number of actuators are
available to control the system. The combination of nonlinear system dynamics with a limited number of state sensors and
control actuators entails a complex and challenging problem in PDE control.

\subsection{Reinforcement Learning}
Reinforcement learning (RL)
is a general framework for solving sequential decision-making processes. RL has been applied to a variety
of different tasks, including natural language processing for dialogue generation
\cite{openai2024gpt4technicalreport}
and text summarization \cite{li2016deep}, computer vision for object detection
and image classification \cite{mnih2013playing}, robotics for autonomous control and
manipulation \cite{levine2016end}, finance for portfolio management and
trading strategies \cite{jiang2017deep}, and game playing for mastering complex
strategic games \cite{silver2016mastering}.
These applications highlight the broad versatility and significant impact of
RL across multiple fields.

RL is the subset of machine learning that focuses on training agents to make 
decisions by interacting with an environment.
The RL framework is typically modeled as a Markov Decision Process (MDP), defined by the
tuple $\mathcal{M} := (\mathcal{X}, \mathcal{U}, \mathcal{P}, \mathcal{R}, \gamma)$, where
$\mathcal{X} \subseteq \RR^\dimX$ is the set of observable states,
$\mathcal{U} \subseteq \RR^\dimU$ is the set of actions,
$\mathcal{P}: \mathcal{X} \times \mathcal{X} \times \mathcal{U} \to [0, 1]$ is the transition
probability kernel with $\mathcal{P}\left(\vec{x}^\prime, \vec{x}, \vec{u}\right)$ representing
the probability of reaching the state $\vec{x}^\prime \in \mathcal{X}$ while being in the state
$\vec{x} \in \mathcal{X}$ and applying action $\vec{u} \in \mathcal{U}$, $\mathcal{R}: \mathcal{X}
\times \mathcal{U} \to \RR$ is the reward function, and $\gamma \in (0, 1]$ is the discount
factor. 

The goal of an RL agent is to learn an optimal policy $\pi$ 
that maps states $\vec{x}$ of the environment to actions $\vec{u}$ in a way that maximizes the expected cumulative reward
over time a control horizon $H$:
\begin{align}
R_H = \EE\left[\sum_{t=0}^{H} \gamma^t r_t \right], \label{eq:CumulativeReward}
\end{align}
with rewards $r_t$ collected at timesteps $t$ and a control horizon $H$ being finite or
infinite -- in this work we focus on finite control horizons.
RL agents often optimize the policy by approximating either the value function $V: \mathcal{X} \to \RR$
or the action-value function $Q: \mathcal{X} \times \mathcal{U} \to \RR$ to quantify, and optimize the
cumulative reward \cref{eq:CumulativeReward} for a state $\vec{x} \in \mathcal{X}$ 
or a state-action pair $(\vec{x}, \vec{u}) \in \mathcal{X} \times
\mathcal{U}$,
respectively. When dealing with high-dimensional state spaces, continuous actions, and nonlinear dynamics,
the estimation of the value function becomes a very challenging and data-inefficient optimization problem.

In general, we distinguish between \emph{model-free} and \emph{model-based} RL algorithms.
Model-free algorithms do not assume any explicit model of the system dynamics, and aim at optimizing
the policy directly by interacting with the environment. This approach offers more flexibility and
is more robust to model inaccuracies but usually requires a large number of interactions to achieve
good performance, making it difficult to apply in cases of data sparsity or expensive interactions.
On the other hand, model-based RL algorithms internally create a model of the environment's dynamics,
i.e., the transition probability kernel $\mathcal{P}$ and the reward function $\mathcal{R}$.\footnote{
Note that in this work we only consider the case of a known full-order reward functions.}
The agent leverages this model to simulate the environment and plan its actions accordingly, typically
resulting in more sample-efficient training.
We focus on Dyna-style \cite{DynaSutton1991} 
RL algorithms which learn a surrogate model of the system dynamics,
allowing the agent to train on an approximation of the environment and thus requiring 
fewer data.

A general scheme of the RL cycle we used for training, including a Dyna-style surrogate dynamics
model of the environment is shown in \cref{fig:TikzSchemeRLSurrogateModel}.
Since this work focuses on Dyna-style RL algorithms with the surrogate model being the center 
of this work, we use the proximal-policy optimization (PPO) algorithm \cite{PPO2017}
as a state-of-the-art actor-critic algorithm.
Actor-critic methods learn both the value-function, i.e., the critic, and the policy, i.e., the actor, at the same time and have shown very promising results in the last years.

\subsection{SINDy: Sparse Identification of Nonlinear Dynamics}
\label{sec:2BackgroundSindy}

We briefly review two versions of \emph{sparse identification of nonlinear dynamics}
(SINDy) used in the SINDy-RL \cite{zolman_sindy-rl_2024} and AE+SINDy
\cite{champion_data-driven_2019} frameworks. SINDy \cite{brunton2016sparseidentificationnonlineardynamics} is an extremely versatile and popular dictionary learning method, that is, a data-driven algorithm aiming at approximating a
nonlinear function through a dictionary of user-defined functions. In particular, SINDy assumes that this approximation takes the form of a sparse 
linear combination of (potentially nonlinear)
candidate functions, such as polynomials of degree $p$ or trigonometric functions.
%Sparse dictionary learning methods perform symbolic regression
%to provide interpretable dynamics representation while %enforcing sparsity
%to balance model complexity and approximation quality. \textcolor{red}{Not clear: SINDy is not based on symbolic regression - for instance, D-CODE by Qian et al, ICLR 2022, is based on symbolic regression. Why do we need to mention it? How Nick's paper provides more details on that?} We refer to
%\cite{zolman_sindy-rl_2024} for more details and
%a comprehensive overview.

In its general formulation, we are given a batch of $N$ data points
$(\vec{x}_1, \vec{y}_1), \ldots, (\vec{x}_N, \vec{y}_N)$ with
$\vec{x}_k \in \RR^m$ and $\vec{y}_k \in \RR^n$ governed by the relationship
$\vec{y}_k = f(\vec{x}_k)$, for $k=1,\ldots, N$. With
\begin{align*}
    \vec{X} &= \left[\vec{x}_1, \ldots, \vec{x}_N\right]^\top \in \RR^{N \times m}, \\
    \vec{Y} &= \left[\vec{y}_1, \ldots, \vec{y}_N\right]^\top \in \RR^{N \times n},
\end{align*}
and a set of $d$ candidate functions
$\vec{\Theta}(\vec{X}) = \left[\theta_1(\vec{x}), \ldots, \theta_d(\vec{x})\right]
\in \RR^{N \times d}$,
we aim to find a representation of the form
\begin{align}
    \vec{Y} = \vec{\Theta}(\vec{X}) \cdot \vec{\Xi} \label{eq:SINDyGenericDynamicsFormulation},
\end{align}
where $\mathbf{\Xi} \in \RR^{d \times n}$ are the coefficients to be fit. These are usually
trained in a Lasso-style optimization since we assume sparsity, i.e., the system
dynamics can be sufficiently represented by a small subset of terms in the library.
Overall, we optimize (a variation of) the following loss:
\begin{align}
    \vec{\Xi} = \argmin_{\hat{\vec{\Xi}}} \norm{\vec{Y} - \vec{\Theta}(\vec{X})
    \hat{\vec{\Xi}}}_F + \mathcal{L}(\hat{\vec{\Xi}}), \label{eq:SINDyGeneralOptimizationLoss}
\end{align}
with a regularization term $\mathcal{L}$ promoting sparsity, i.e., usually
$\mathcal{L}(\vec{\Xi}) = \norm{\vec{\Xi}}_1$. While in
\cite{champion_data-driven_2019} and \cite{zolman_sindy-rl_2024} the loss
\cref{eq:SINDyGeneralOptimizationLoss} is trained by an $L^2$ penalty
and sequential thresholding least squares (STLS), we use \emph{PyTorch}'s automatic differentiation
framework \cite{paszke2017automatic} and optimize the $\norm{\cdot}_1$-loss directly.

\subsubsection{SINDy-RL}
\label{sec:ZolmanSINDy-RL}
%In their work, Zolman et al. \cite{zolman_sindy-rl_2024,arora2022modelbasedreinforcementlearningsindy} present the use of SINDy combined with a reinforcement learning (RL) algorithm for Dyna-style RL. 
SINDy-RL is a very recent extension of SINDy that combines SINDy with a 
RL algorithm for Dyna-style RL \cite{zolman_sindy-rl_2024,arora2022modelbasedreinforcementlearningsindy}.
This approach involves using SINDy to 
learn a surrogate model of the environment and then train the RL agent using this approximation of
the full-order model. This method results in a drastically-reduced number of necessary full-order
interactions and very fast convergence. Specifically, Zolman et al. \cite{zolman_sindy-rl_2024} work with the discrete
\SindyC{} formulation, i.e., including controls,
for the discovery task $\vec{x}_{k+1} = f(\vec{x}_k, \vec{u}_k)$,
where $\vec{X} = (\vec{x}(t_k), \vec{u}(t_k))_{k=1, \ldots, N}$ and $\vec{Y} = (\vec{x}(t_{k+1}))_{k=1,\ldots, N}$ in the notation
of \cref{eq:SINDyGenericDynamicsFormulation}. Note that since in our work we focus on enabling efficient surrogate representations for controlling distributed systems, 
we do not further review the reward approximation nor the policy approximation aspects addressed in  \cite{zolman_sindy-rl_2024}.

Despite significantly reducing the number of full-order interactions, not requiring derivatives of
(potentially) noisy measurements, and incorporating controls in the surrogate model, so far SINDy-RL lacks the scalability to high-dimensional systems (see
\cite[section 6]{zolman_sindy-rl_2024}). 
While it shows convincing results for state spaces with up to eight dimensions and action spaces with
up to two dimensions, controlling distributed systems  require much larger state and action spaces to be
solved and controlled sufficiently accurately.

\subsubsection{\AeSindy{}: Autoencoder Framework for Data-driven Discovery in
    High-dimensional Spaces}
\label{sec:ChampionSINDyAE}

In their work \cite{champion_data-driven_2019}, Champion et al. rely on the classical SINDy formulation for the discovery task in the latent space. 
%$\dot{\vec{x}} = f(\vec{x})$, with $\dot{\vec{x}} = \frac{\mathrm{d}}{\mathrm{d}t}\vec{x}$. In the notation of \cref{eq:SINDyGenericDynamicsFormulation},this corresponds to $\vec{x}_k = \vec{x}(t_k)$ and $\vec{y}_k = \dot{\vec{x}}(t_k)$.
Indeed, they generalize %Their work effectively generalizes 
the concept of data-driven dynamics discovery to 
high-dimensional dynamical systems by combining the classical SINDy formulation with an
AE framework \cite{ReducingDimHinton2006,
BengioReviewUnsupervisedLearning2012,Goodfellow-et-al-2016} to allow for nonlinear dimensionality reduction. 
Hence, instead of directly working with high-dimensional systems, the SINDy algorithm is applied to the compressed, lower-dimensional representation of the system with
$\dimRedX \ll \dimX$.

Given the encoder $\phi(\;\cdot\;, \mathbf{W}_\phi): \RR^\dimX \to \RR^\dimRedX$ and the decoder network
$\psi(\;\cdot\;, \mathbf{W}_\psi): \RR^\dimRedX \to \RR^\dimX$, as well as the latent representation
$\vec{z}(t) = \phi(\vec{x}(t)) \in \RR^\dimRedX$, the following loss is minimized:
\begin{align}
    \begin{aligned}
\min_{\vec{W}_{\phi}, \vec{W}_{\psi}, \vec{\Xi}}\quad &\underbrace{\left\| \vec{x} - \psi(\vec{z}) \right\|_2^2}_{\text{reconstruction loss}} 
    + \underbrace{\lambda_1 \left\| \dot{\vec{x}} - (\nabla_{\vec{z}} \psi(\vec{z})) \left( 
            \vec{\Theta}(\vec{z}^\top) \vec{\Xi} \right) \right\|_2^2}_{\text{SINDy loss in } \dot{\vec{x}}}\\
    &+ \underbrace{\lambda_2 \left\| \dot{\vec{z}} - \vec{\Theta}(\vec{z}^\top)
        \vec{\Xi} \right\|_2^2}_{\text{SINDy loss in the latent space}} 
    + \underbrace{\lambda_3 \left\| \vec{\Xi} \right\|_1}_{\text{sparse regularization}},
    \end{aligned}
    \label{eq:LossFunctionChampionAE}
\end{align}
with $\dot{\vec{z}} = (\nabla_{\vec{x}} \vec{z}) \dot{\vec{x}}$.
The SINDy loss in $\dot{\vec{x}}$, i.e., the \emph{consistency loss}, ensures that the time
derivatives of the prediction matches the input time derivative $\dot{\vec{x}}$; we refer to fig. 1 in \cite{champion_data-driven_2019} for more
details.\footnote{The \AeSindy{} algorithm could be interpreted as the upper half of
\cref{fig:TikzArchitectureAutoEncoder} with a different loss function.}
As described in the supplementary material of \cite{champion_data-driven_2019},
due to its non-convexity and to obtain a parsimonious dynamical model,
the loss in \cref{eq:LossFunctionChampionAE} is optimized via
STLS.

AE+SINDy provides very promising results for (re-)discovering the underlying true
low-dimensional dynamics representation of high-dimensional dynamical systems. Further extensions of the AE+SINDy framework have been recently proposed in \cite{conti_reduced_2023} and \cite{Bakarji2022PDEDiscovering}.
%\textcolor{red}{Bakarji et al, rspa.2023.0422}. 
However, to be applicable in a control setting with (potentially) noisy data, AE+SINDy suffers from the necessity of requiring derivatives of the observed
data and the inability to include controls in the SINDy framework. In the next section, we will combine and generalize the approaches of SINDy-RL
and \AeSindy{} to develop \ourAlgo{} within an RL setting.

\section{AE+SINDy-C: Low-dimensional Sparse Dictionary Learning for Simultaneous Discovery and Control of Distributed Systems}% \textcolor{red}{simultaneous discovery and control of distributed systems? (better, but then I would remove RL)}}
\label{sec:Section3OurAlgo}

\subsection{Latent Model Discovery} %Learning of Latent Dynamics 

To efficiently scale SINDy to high-dimensional state spaces, incorporate controls, and eliminate reliance on derivatives, we introduce \ourAlgo{}. This approach aims to
accelerate DRL for control tasks involving distributed -- potentially large-scale -- systems, by combining the derivative-free Dyna-style DRL training in the SINDy-C case, as demonstrated by
\cite{zolman_sindy-rl_2024,arora2022modelbasedreinforcementlearningsindy},
with the scalable approach to high-dimensional state- and action-spaces from \cite{champion_data-driven_2019}.

Our method approximates the environment dynamics to speed up computationally-expensive simulations, significantly reducing the need for interactions with
the full-order model. Additionally, it yields an interpretable and parsimonious dynamics
model within a low-dimensional surrogate space. The dimensions of the latent spaces,
$\dimRedX$ and $\dimRedU$, representing state and control respectively, can be
set equal to the intrinsic dimension of the dynamical system when this is known a
priori. Alternatively, these dimensions can be tuned to discover the intrinsic dimension
of an unknown system, solely based on available measurements. Our experiments
in \cref{sec:NumericalExperiments} demonstrate that in some cases \ourAlgo{}
is able to independently discover the underlying intrinsic dimensions for the
control space.

We operate in the (discrete-time) SINDy-C setting, as described in
\cref{sec:ZolmanSINDy-RL}, with $\vec{x}_k = (\vec{x}(t_k), \vec{u}(t_k))$
and $\vec{y}_k = \vec{x}(t_{k+1})$. Here, $\vec{x}(t_k) \in \RR^\dimObsX$ where
$\dimObsX = \dimX$ in the fully observable case, and $\dimObsX \ll \dimX$ in the
partially observable case, respectively, and $\vec{u}(t_k) \in \RR^\dimU$.\footnote{Compared
to \cref{fig:TikzSchemeRLSurrogateModel}, we simplify the $\vec{x}_t^\observed$ 
notation by omitting the superscript.} Overall, we are interested in the discovery task
$\vec{x}_{k+1} = f(\vec{x}_k, \vec{u}_k)$. Instead of applying \SindyC{} directly on our
measurements, we follow the idea of Champion et al. \cite{champion_data-driven_2019} to first compress the states and actions,
then apply \SindyC{} in the low-dimensional latent space, predict the next state,
and subsequently decompress the obtained prediction back to the observation space.

\begin{figure}
    \centering
    \input{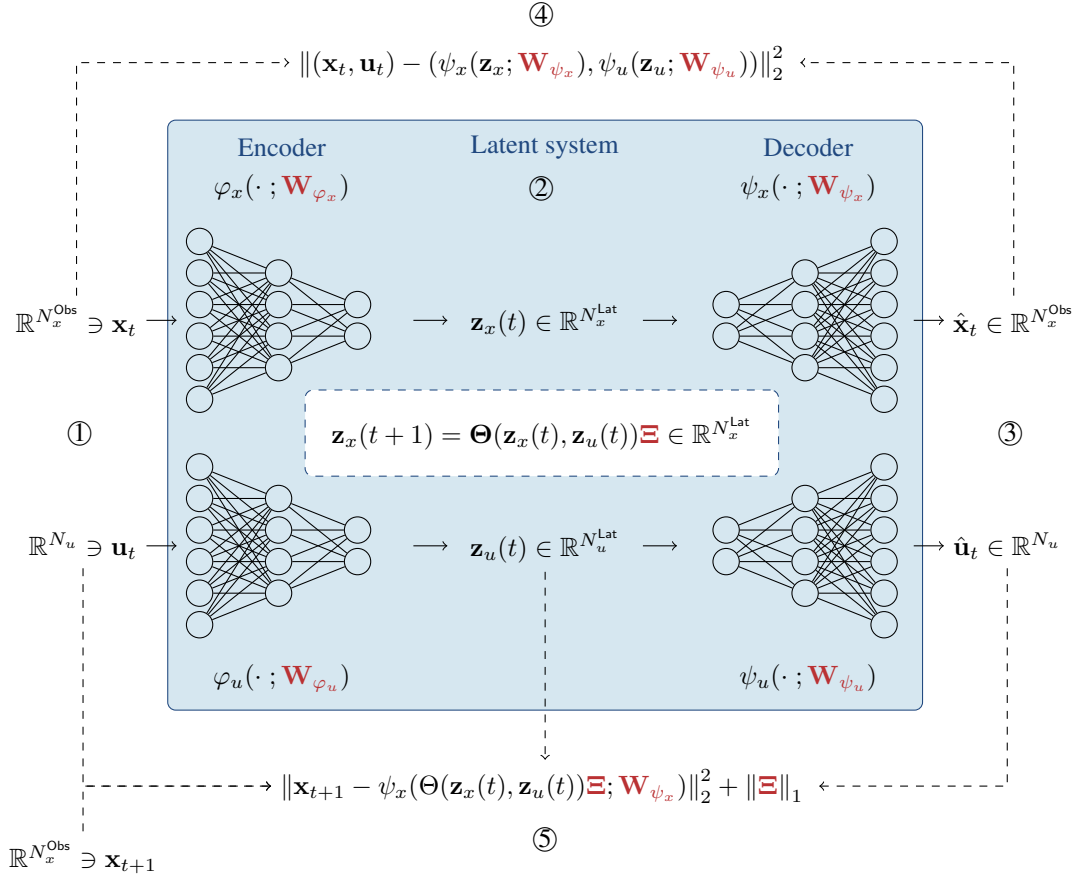}
    \caption{AE architecture and loss function used during the training stage.
    \lightRed{Trainable parameters} are highlighted in red. The different stages of
    the training scheme can be listed as follows. (1) the current state $\vec{x}_t$,
    applied control $\vec{u}_t$, and the next state $\vec{x}_{t+1}$ are provided as input
    data. (2) After compressing both the current state and the control vector, the
    \SindyC{} algorithm is applied in the latent space, yielding a low-dimensional
    representation of the prediction for the next state. (3) The latent space
    representations of the current state, the control, and the next state prediction are decoded.
    (4) The classical AE loss and a regularization term to promote sparsity are computed.
    (5) The \SindyC{} loss is computed. The figure is inspired by 
    \cite[fig. 1]{conti_reduced_2023}.}
    \label{fig:TikzArchitectureAutoEncoder}
\end{figure}

\paragraph{Offline Training} As already described in the previous section, the overall
framework is shown in \cref{fig:TikzArchitectureAutoEncoder} and can be divided into
three steps:
\begin{enumerate}[label=(\roman*)]
    \item \textbf{Encoding:} The observed state $\vec{x}_t \in \RR^\dimObsX$ and the current control
        $\vec{u}_t \in \RR^\dimU$ are individually compressed by two separated encoder networks
        $\phi_x(\;\cdot \;; \vec{W}_{\phi_x}) : \RR^\dimObsX \to \RR^\dimRedX$ and 
        $\phi_u(\;\cdot \;; \vec{W}_{\phi_u}) : \RR^\dimU \to \RR^\dimRedU$, yielding
        their low-dimensional latent-space equivalent $\vec{z}_x(t) \in \RR^\dimRedX$ and
        $\vec{z}_u(t) \in \RR^\dimRedU$, respectively.
    \item \textbf{Discrete \SindyC{} in the latent space:} Given a set of $d$ dictionary functions
        (e.g. polynomials, trigonometric functions, etc.) the dictionary 
        $\vec{\Theta}(\vec{z}_x(t),\vec{z}_u(t))= \vec{\Theta}( \phi_x(\vec{x}_t; 
        \vec{W}_{\phi_x}), \phi_u(\vec{u}_t;
        \vec{W}_{\phi_u})) \in \RR^d$ is evaluated
        and multiplied by $\vec{\Xi} \in \RR^{d \times \dimRedX}$, i.e., the coefficients to be fit.
    \item \textbf{Decoding:} The result $\vec{z}_x(t+1) = \vec{\Theta}(\vec{z}_x(t), 
        \vec{z}_u(t)){\vec{\Xi}} \in \RR^\dimRedX$ is then decoded to obtain a prediction for the
        next state of the high-dimensional system by using the state-decoder
        $\psi_x(\;\cdot\;; \vec{W}_{\psi_x}) : \RR^\dimRedX \to \RR^\dimObsX$. In order to train the
        AE, the compressed state and action are also decoded and fed into the loss function
        by using the state-decoder and the control-decoder
        $\psi_u(\;\cdot\;; \vec{W}_{\psi_u}) : \RR^\dimRedU \to \RR^\dimObsU$.
\end{enumerate}
Overall, we obtain the following loss function:
\begin{align}
    \begin{aligned}
\min_{\vec{W}_{\phi_x}, \vec{W}_{\phi_u}, \vec{W}_{\psi_x}, \vec{W}_{\psi_u}, \vec{\Xi}}
    \quad &\underbrace{\norm{\vec{x}_{t+1}  - \psi_x\left(\vec{\Theta}( \phi_x(\vec{x}_t; \vec{W}_{\phi_x}),
            \phi_u(\vec{u}_t; \vec{W}_{\phi_u})) \cdot {\vec{\Xi}};
                {\vec{W}_{\psi_x}}\right)}_2^2}_{\text{forward \SindyC{} prediction loss}}\\
    &+ \lambda_1 \underbrace{\norm{\vec{x}_t - \psi_x(\phi_x(\vec{x}_t; \vec{W}_{\phi_x}); 
         {\vec{W}_{\psi_x}})}_2^2}_{\text{autoencoder loss state}}\\
     & + \lambda_1 \underbrace{
     \norm{\vec{u}_t - 
   \psi_u( \phi_u(\vec{u}_t; \vec{W}_{\phi_u}); {\vec{W}_{\psi_u}})}_2^2
   }_{\text{autoencoder loss control}}\\
    &+ \lambda_2 \underbrace{\norm{{\vec{\Xi}}}_1}_{\text{sparsity regularization}},
    \end{aligned}
    \label{eq:AutoencoderLossFunction}
\end{align}
representing the loss of the decoded forward prediction in the \SindyC{} latent space formulation, i.e.,
part (ii) and (iii), the classical AE loss and also the regularization loss to promote sparsity in the
SINDy-coefficients. Since our entire pipeline is implemented in the \emph{PyTorch} library
\cite{paszke2017automatic}, we train
\cref{eq:AutoencoderLossFunction} by using the automatic differentiation framework and thus do not rely
on STLS. We use \cite{kaptanoglu2021pysindy} to create the set of $d$ dictionary functions once
in the beginning of the pipeline.
The parameters $\lambda_1, \lambda_2 \in \RR_{>0}$ are hyperparameters to individually weight
the contribution of each term.
%For alternative ideas on defining the loss function, we refer to \cref{sec:FutureWork}.

\paragraph{Online Deployment} Following the training stage, where the parameters 
$\vec{W}_{\phi_x}$, $\vec{W}_{\phi_u}$, $\vec{W}_{\psi_x}$, $\vec{W}_{\psi_u}$, and $\vec{\Xi}$
are learned, the trained network can be deployed as a Dyna-style environment approximation to
train the DRL agent. The surrogate environment enables extremely fast inferences and,
if well-trained, provides accurate predictions of the system dynamics. Given a current state $\vec{x}_t \in \RR^\dimObsX$ (either observed or simulated)
and a control $\vec{u}_t \sim
\pi(\;\cdot\; \vert \vec{x}_t)$, the prediction of the next state is computed as follows:
\begin{align*}
    \hat{\vec{x}}_{t+1} = \psi_x\left(\vec{\Theta}( \phi_x(\vec{x}_t; \vec{W}_{\phi_x}),
    \phi_u(\vec{u}_t; \vec{W}_{\phi_u})) \cdot {\vec{\Xi}};
        {\vec{W}_{\psi_x}}\right) \approx f(\vec{x}_t, \vec{u}_t),
\end{align*}
which involves only one matrix multiplication, the evaluation of $d$ dictionary 
functions, and three NN forward passes, resulting in exceptionally-low inference times.

\subsection{DRL Training Procedure}
We describe the usage of \ourAlgo{} %\textcolor{red}{casting it? what do you mean exactly?} 
in a Dyna-style MBRL, see 
\cref{alg:MBRLAlgorithmScheme}. The most important point is \cref{algoLine:FitFreq}, where the
agent in trained $\kDyn - 1$ times on the surrogate model, before new data is collected by
interacting with the full-order environment. This procedure ensures optimal data usage and drastically
reduces the required number of full-order interactions. Our experiments show that the choice of
the hyperparamter $\kDyn$ strongly influences the stability,
convergence speed and data usage of the DRL training.
We use the PPO algorithm \cite{PPO2017} as a state-of-the-art actor-critic policy. As in \cite{zolman_sindy-rl_2024}, we also use the PPO gradients to update the parameters of the autoencoder surrogate model.

To also correctly emulate the partial observability, i.e., $\dimObsX < \dimX$, in the reward function, we project the target state into the lower dimensional observation space and compute the projected reward.
Namely, for the Burgers' equation we compute
$(\vec{x}_t^\observed - C\cdot \vec{x}_\text{ref})^\top Q^{\mathrm{Proj}} (\vec{x}_t^\observed - C\cdot \vec{x}_\text{ref})$, where $Q^{\mathrm{Proj}}$ is
a scaled identity matrix in the lower dimensional observation space. This procedure makes the training more challenging, since we only obtain partial information via the reward function. In the case of $\dimObsX = \dimX$ and for the evaluation of all final models, we use the closed form of the full-order reward function.

\begin{algorithm}
    \caption{Dyna-style MBRL using \ourAlgo{}}
    \label{alg:MBRLAlgorithmScheme}
    \begin{algorithmic}[1]
    \Require $\mathcal{E}, \pi_0, \vec{\Theta}, N_{\text{off}},
        N_{\text{collect}}, n_{\text{batch}}, \kDyn$
    \Ensure Optimized Policy $\pi$
    \Function{OptimizePolicy}{$\mathcal{E}, \pi_0, \mathcal{A}$}
        \State $D_{\text{off}} = \text{CollectData}(\mathcal{E}, \pi_0, N_{\text{off}})$
            \Comment{Collect off-policy data from full-order environment, $\mathcal{E}$}
        \State $D = \text{InitializeDatastore}(D_{\text{off}})$
        \State $\hat{\mathcal{E}} = \text{\ourAlgo{}}(D, \vec{\Theta})$ \Comment{Fit surrogate environment}
        \State $\pi = \text{InitializePolicy}()$
        \While{not done}
            \For{$\kDyn - 1$ steps } \label{algoLine:FitFreq}
                    \Comment{Train the agent $\kDyn - 1$ times using the surrogate environment}
                \State $\pi = \mathrm{PPO}.\mathrm{update}(\hat{\mathcal{E}}, \pi, n_{\text{batch}})$
                    \Comment{Update DRL policy using surrogate environment, $\hat{\mathcal{E}}$}
            \EndFor
            \State $D_{\text{on}} = \text{CollectData}(\mathcal{E}, \pi, N_{\text{collect}})$
                \Comment{Collect on-policy data}
            \State $D = \text{UpdateStore}(D, D_{\text{on}})$
            \State $\hat{\mathcal{E}} = \text{\ourAlgo{}}(D, \vec{\Theta})$ \Comment{Update surrogate environment}
            \label{algoLine:SurrogateFitEnvironment}
        \EndWhile
        \State \Return $\pi$
    \EndFunction
    \end{algorithmic}
\end{algorithm}
\section{Numerical Experiments}
\label{sec:NumericalExperiments}
To validate our approach, we study two control problems related with either the Burgers' equation or the Navier-Stokes equations, on two test benchmark implementations provided by the well-established \emph{ControlGym} library \cite{zhang_controlgym_2023} and the \emph{PDEControlGym} library \cite{bhan2024pde} respectively. In both cases, we compare our Dyna-style MBRL \cref{alg:MBRLAlgorithmScheme}
with the model-free PPO baseline which only interacts with the full-order environment. The Burgers' equation serves as an 
initial example to highlight the data efficiency of the algorithm, explore both partially and fully 
observable cases, examine robustness to noisy observations, and experiment with discovering control 
dimensions in the latent space. Additionally, we emphasize the method's strengths in sample efficiency,
out-of-distribution generalization for the initial condition, interpretability of the learned surrogate
dynamics, and details about the autoencoder's training, which are covered in \cref{subsec:BurgersEquation}.
In contrast, the Navier-Stokes equations offer a more challenging example, with only boundary controls and a much higher-dimensional state space, underscoring \ourAlgo{}'s scalability. This case will showcase how the method can be applied to complex, high-dimensional systems, discussed in 
\cref{subsec:NavierStokesEquation}.

Since \ourAlgo{} serves as a Dyna-style MBRL algorithm, we analyze the efficiency of the proposed
framework with respect to the number of full-order model interactions. For all our experiments, we use the \rlLib{} package \cite{RayRlLib2017}; 
\ourAlgo{} is implemented in \emph{PyTorch} \cite{paszke2017automatic}. All details about
the PDE environments parameters are
available in \cref{sec:AppendixHyperparameters}.
For further details on the DRL training we refer to \cref{sec:AppendixDRL}; instead, specific details about the AE training can be found in \cref{sec:AppendixAutoencoder}.

\subsection{Burgers' Equation (ControlGym)}
\label{subsec:BurgersEquation}
The viscous Burgers equation can be interpreted as a one-dimensional, simplified version of the 
Navier-Stokes equations, describing fluid dynamics and capturing key phenomena
such as shock formations in water waves and gas dynamics.
Solving the Burgers' equation is particularly challenging
for RL and other numerical methods due to its nonlinearity and the presence of sharp
gradients and discontinuities.
Given a spatial domain $\Omega = [0, L] \subset \RR$ and a time horizon $T$, we consider
the evolution of continuous fields $\vec{x}(x,t) : \Omega \times [0, T] \to \RR$ under a
the temporal dynamics:
\begin{align}
    \begin{aligned}
        \partial_t \vec{x}(x,t) + \vec{x}(x,t) \partial_x \vec{x}(x,t) - \nu \partial_{xx}^2 \vec{x}(x,t) &= \vec{u}(x,t)\\
        \vec{x}(x,0) &= \vec{x}_\text{init}(x)
    \end{aligned}\;,
    \label{eq:BurgersEquation}
\end{align}
given an initial state $\vec{x}_\text{init}: \Omega \to \RR$, a constant diffusivity $\nu > 0$, a source term (also called forcing function) $\vec{u}(x, t) : \Omega \times [0,T] \to \RR$ and periodic boundary conditions.
Internally, \emph{ControlGym} discretizes
the PDE in space and time, assuming to rely on uniformly spatially distributed control inputs $\vec{u}_t$, which are 
piecewise constant over time. This results in a discrete-time finite-dimensional nonlinear system of the form:
\begin{align}
    \vec{x}_{t+1} = f(\vec{x}_t, \vec{u}_t; \vec{w}_k),
\end{align}
including (optional) Gaussian noise $\vec{w}_k$.\footnote{To be precise, the discrete dynamics $f$ also depend on the mesh size of the discretization.}
In this work, the internally used solution
parameters are set to their default and for more details we refer to
\cite[p.6]{zhang_controlgym_2023} and \cref{sec:AppendixHyperparameters}. 
As previously mentioned, since our new encoding-decoding scheme does not rely on
derivatives, we explicitly want to test the robustness of \ourAlgo{} with respect to
observation noise. Thus, we follow the observation process by \emph{ControlGym}
given by
\begin{align*}
    \vec{x}_t^\observed = C\cdot \vec{x}_t + \vec{v}_t,
\end{align*}
where the matrix $C \in \RR^{\dimObsX \times \dimX}$ is as described in 
\cref{fig:TikzSchemeRLSurrogateModel} and $\vec{v}_t \sim \mathcal{N}(0, \Sigma_v)$
is zero-mean Gaussian noise with symmetric positive definite covariance matrix $\Sigma_v = \sigma^2 \cdot \mathrm{Id}_{\dimObsX} \in \RR^{\dimObsX \times \dimObsX}, \sigma^2 > 0$. For both problems, we define a target state $\vec{x}_\text{ref}$ and use the following objective function to be maximized:
\begin{align*}
    \mathcal{J}(\vec{x}_t, \vec{u}_t) = -\mathbb{E} \left\{ \sum_{t=0}^{K-1} \left[ (\vec{x}^\observed_t -
        \vec{x}_\text{ref})^\top Q (\vec{x}^\observed_t - \vec{x}_\text{ref}) + \vec{u}_t^\top R \vec{u}_t \right] \right\},
\end{align*}
%\textcolor{red}{not clear how the random $\vec{x}_t^\observed$ affects the objective. Moreover, why minus in front of it? -> we want to maximize + fixed a typo}
with $K$ discrete time steps and positive definite matrices
$Q \in \RR^{\dimObsX \times \dimObsX}$
and $R \in \RR^{\dimObsU \times \dimObsU}$ balancing the control effort and the tracking performance. As reference state and controls, we both use the zero-vector. The inherent solution manifold of \cref{eq:BurgersEquation} is two-dimensional, representing both space and time. 

In this work, the diffusivity constant $\nu = 1.0$ is fixed and we choose $\Omega = (0,1)$ as spatial domain, that is, we select $L=1$. We propose $\dimRedX = 2$, since the inherent dimension of the solution manifold is two, and
$\dimRedU = 2$ for the reason of symmetry. The target state of $\vec{x}_\text{ref} \equiv \mathbf{0} \in \RR^\dimObsX$ is used. In \cref{sec:DynamicsSurrogateSpaceBurgers} we will see
how \ourAlgo{} is independently able to compress the controls even further.

For the Burgers' equation, we trained \ourAlgo{} with different 
full-order model update frequencies $\kDyn = 5, 10, 15$ to analyze the sample efficiency
and the sensitivity of the surrogate model. In the case  $\kDyn = 15$ the internal modeling
bias was too big for a successful DRL training, which is the reason why this case is excluded from the
analysis. This appears to be a natural limitation since the number of full-order interactions 
was too low and thus not sufficient to adequately model the system dynamics under investigation. 

\subsubsection{Sample Efficiency and Speed-Ups}
\label{sec:5AnalysisSampleEfficiency}
\begin{figure}[htb]
	\centering
	\includegraphics[width=.95\textwidth]{
		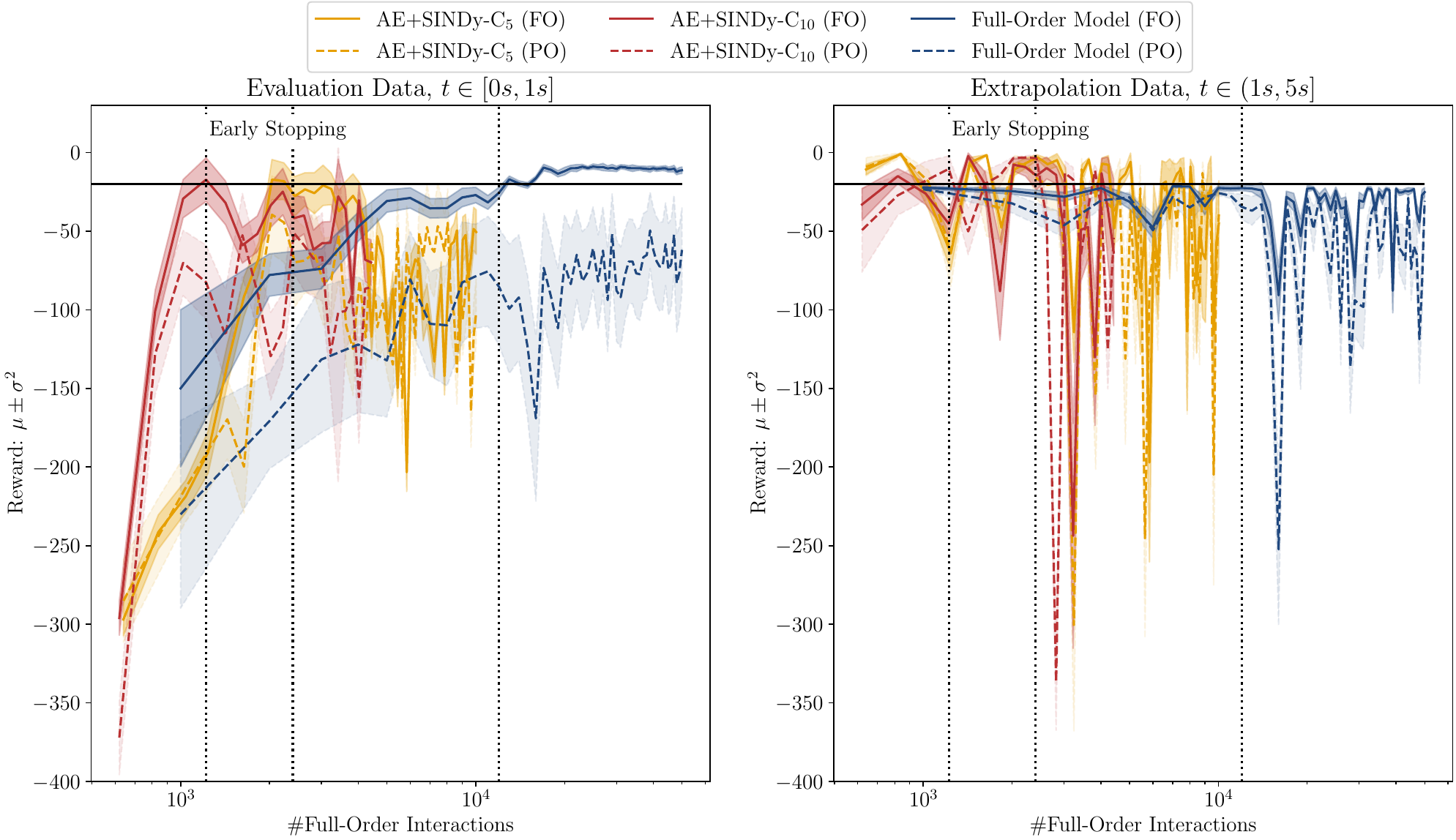
	}
	\caption{Sample efficiency of the Dyna-style \ourAlgo{} method for the \textbf{Burgers'} equation.
	 	We test
		$\kDyn = 5, 10$ against the full-order baseline for the \textbf{fully observable (solid line)}
		and \textbf{partially observable (dashed line)}. The \textbf{dashed vertical lines} indicate the
		point of early stopping for each of the model  classes (FO + PO) after \textbf{100 epochs}
		and represent the models
		which are evaluated in detail in \cref{sec:RandomInitialConditionStateControlBurgers}.
        For the evaluation the performance over \textbf{five fixed random seeds} is used.}
	\label{fig:BurgerFOMInteractionsPerformance}
\end{figure}

As in the training, the initial
condition is given by a uniform $\Uniform$ distribution, for more details
we refer to \cref{sec:RandomInitialConditionStateControlBurgers}.
We use an evaluation time horizon of $T_\text{eval} = 1\text{s}$ and let 
the agent extrapolate over time for four more seconds, i.e.,
$T_\text{extrap} = 5\text{s}$, resulting in $T=5\mathrm{s}$, \textcolor{black}{4 times more than the training horizon}.
Fig. \ref{fig:BurgerFOMInteractionsPerformance} shows
the average reward ($\pm$ one standard deviation) given the number of full-order interactions
each model has observed. For a quantitative overview, we refer to \cref{tab:ComparisonBurgerNumbers} where we display the results with further detail.
We separately illustrate the performance on the evaluation horizon
$[0s, 1s]$, i.e. the time horizon the agent interacts for during the training, and the
extrapolation horizon $(1s, 5s]$. The vertical dashed lines in 
\cref{fig:BurgerFOMInteractionsPerformance}
represent the models after 100 epochs and are deployed to evaluate the agent's performance,
i.e.\cref{tab:ComparisonBurgerNumbers}.
%\textcolor{red}{models in which sense? -> I extracted the trained
%agents/models and evaluated them to obtain the graphs} which are analyzed in detail in
%\cref{sec:RandomInitialConditionStateControlBurgers}.
We save all intermediate models and stop the training for all models after exactly 100 epochs,
which is approximately the point where the extrapolation performance drops.
In the case of the \ourAlgo{} algorithm this behavior is also evident in the evaluation data,
caused by the high sensitivity of the online training of the autoencoder.

In general, we can see that all of the models overfit the dynamics.
While the performance on the evaluation time interval still improves, the model becomes
very unstable when extrapolating in time, and at the end of the training interval 
almost all model solutions diverge.
As expected, generally the model fully observing the PDE outperforms the model which only
observes partial measurements. Additionally, the partially observable cases exhibit a higher
standard deviation and thus higher variability in their  performance.

\paragraph{Partially observable (PO)} Generally, the models only partially observing the system exhibit worse performances and suffer under higher variability. At the end of the training horizon, in both
cases the \ourAlgo{} method overfits the dynamics model, resulting in divergent
solutions when extrapolating in time. The highest performance for each of the PO cases is almost similar, although \ourAlgo{} 
with $\kDyn = 10$ requires nearly 10x less data compared to the full-order model, and with
$\kDyn = 5$ close to 5x less data respectively (cf. \cref{tab:ComparisonBurgerNumbers})
-- matching the intuition of sample efficiency of \cref{alg:MBRLAlgorithmScheme} based on $\kDyn$.

\paragraph{Fully observable (FO)} As already mentioned, in a direct comparison all of the trained
models in the FO case outperform their PO counterparts, quantitatively and qualitatively.
As visualized in \cref{tab:ComparisonBurgerNumbers}, and similarly to the PO case,
we achieve nearly the same level of performance with 10x, respectively 5x, less data.

\begin{table}[h]
	\centering
	\scalebox{0.59}{
	\begin{tabular}{lllllll}
	\toprule
											& \multicolumn{2}{c}{Baseline}                          & \multicolumn{2}{c}{\ourAlgo{}, $\kDyn = 5$} & \multicolumn{2}{c}{\ourAlgo{}, $\kDyn = 10$}      \\
											\cmidrule(rl){2-3} \cmidrule(rl){4-5} \cmidrule(rl){6-7} 
											& PO $48\times 8$          & FO $256 \times 8$        & PO $48\times 8$                        & FO $256 \times 8$        & PO $48\times 8$          & FO $256 \times 8$        \\
	\midrule
	\#FOM Interactions                     & 12000                    & 12000                    & 2400                                   & 2400                     & \bf{1200}                     & \bf{1200}                     \\
	$\quad$Off-policy                              & -                      & -                      & 200                                    & 200                      & 200                      & 200                      \\
	$\quad$On-poliy                        & 12000                    & 12000                    & 2200                                   & 2200                     & 100                      & 1000                     \\
	Reward $\mathcal{R}$ & & & & & & \\
	$\quad$Random Init ($\mu \pm \sigma^2$)        &                          &                          &                                        &                          &                          &                          \\
	$\qquad t \in [0\mathrm{s},1\mathrm{s}]$                           & $-85.42 \pm 37.13$       & $\bf{-8.52 \pm 3.47}$         & $-51.95 \pm 25.03$                     & $-17.24 \pm 12.81$       & $-82.52 \pm 24.77$       & $-17.22 \pm 12.39$       \\
	$\qquad t \in (1\mathrm{s}, 5\mathrm{s}]$                          & $-14.52 \pm 8.09$        & $\bf{-2.61 \pm 1.51}$         & $-6.50 \pm 4.54$                       & $-10.62 \pm 3.86$        & $-10.39 \pm 8.22$        & $-45.29 \pm 7.65$        \\
	$\quad$Bell Function Init ($\mu \pm \sigma^2$) &                          &                          &                                        &                          &                          &                          \\
	$\qquad t \in [1\mathrm{s},2\mathrm{s}]$                           & $\bf{-31551.70 \pm 10772.44}$ & $-32817.89 \pm 11513.03$ & $-44054.61 \pm 13382.10$    & $-42032.25 \pm 13418.30$ & $-47417.93 \pm 14071.58$ & $-45027.72 \pm 13936.04$ \\
	$\qquad t \in (2\mathrm{s}, 6\mathrm{s}]$                          & $\bf{-2642.00 \pm 1055.71}$   & $-2778.67 \pm 1313.07$   & $-23165.13 \pm 8099.89$ 	   & $-15335.65 \pm 5967.48$  & $-35431.61 \pm 10610.13$ & $-25302.46 \pm 8719.55$  \\
	Total \#parameters                      & \bf{13705}                    & 66953                    & 14905                                  & 78727                    & 14905                    & 78727                    \\
	$\quad\ourAlgo{}$                            & -                        & -                        & 1200                                   & 11774                    & 1200                     & 11774                    \\
	$\quad$Actor + Critic                          & 13705                    & 66953                    & 13705                                  & 66953                    & 13705                    & 66953                 \\
	\bottomrule
	\end{tabular}}
\vspace{4pt}	
 \caption{Performance comparison of the Dyna-style \ourAlgo{} method for the \textbf{Burgers'} equation.
	We test $\kDyn = 5, 10$ against the full-order baseline for the partially observable (PO) and
	fully observable (FO) case. The models correspond to the dashed vertical lines in
	\cref{fig:BurgerFOMInteractionsPerformance} and represent all models after \textbf{100 epochs}.
 	We compare the number of full-order model (FOM) interactions,
	the performance for a random initialization (cf. \cref{sec:RandomInitialConditionStateControlBurgers}),
	the bell-shape initialization and $\nu = 0.01$ (cf. \cref{sec:BellShapeBurgersGeneralization}),
	and the total number of parameters. \textbf{Best performances} (bold) are highlighted \textbf{row-wise}.
	For the evaluation the performance over \textbf{five fixed random seeds} is used.}
	\label{tab:ComparisonBurgerNumbers}
\end{table}
\subsubsection{Generalization: Variation of Initial Condition and Diffusivity Constant}
\label{sec:BellShapeBurgersGeneralization}
Compared to the $\Uniform$ distribution we used during the training phase and the detailed
analysis in \cref{sec:RandomInitialConditionStateControlBurgers}, we are also interested in the
generalization capabilities of the agents trained with \ourAlgo{} on out-of-distribution
initial conditions with more regularity.
Given $N_\text{spatial}$ discretization points in space, we modify the default initial
condition of the \emph{ControlGym} library and define the initial state
\begin{align}
	\begin{aligned}
		(\vec{x}_{\mathrm{init}})_i &= 5 \cdot \frac{1}{\cosh\left(10 \left(\Delta_i - \alpha \right) \right)} \in [0,5] \\
		\Delta_i &= \frac{i}{N_\text{spatial}}, \quad i=0,\ldots, N_\text{spatial}\\
		\alpha &\sim \mathcal{U}[0.25, 0.75]
	\end{aligned},
	\label{eq:BellShapeInitialCondition}
\end{align}
for our second test case, exhibiting a higher degree of regularity. The term $\alpha \sim \mathcal{U}[0.25, 0.75]$ randomly
moves the peak of the curve, enabling the ablation studies presented in \cref{tab:ComparisonBurgerNumbers}.
For the plots reported in \cref{fig:ComparionBellBurgersPartiallyObservableStateControl} and 
\ref{fig:ComparisonBellBurgersFullyObservableStateControl}, we centered the function in the domain by taking $\alpha = 0.5$. With a non-negative domain of
$[0,5]$, this initial state clearly exceed the training domain of $[-1, 1]$. To make the test case
more interesting and more realistic, we also change the diffusivity constant to $\nu = 0.01$, i.e.
two orders of magnitude smaller than on the training data, and we let the PDE evolve uncontrolled
for one second before the controller starts actuating for five more seconds, i.e. a delayed start
and a total observation horizon of $T=6\mathrm{s}$ of which the
controller is activated in $[1\mathrm{s},6\mathrm{s}]$.

Clearly, this task is much more challenging than the one previously considered, and the baseline model is the only agent to perfectly regulate
the state towards zero at the end of the extrapolation horizon. In none of the two cases (PO and FO)
one of the \ourAlgo{} agents is able to push the system towards the zero-state within the evaluation time horizon
of one second. As in the first test setting, the fully observable case seems to be overall easier
for all of the models. Most importantly, as discussed in \cref{sec:RandomInitialConditionStateControlBurgers},
working with a regular initial state, the agent also generates regular control trajectories.
Although we did not apply actuation bounds, the order of magnitude of the controls is almost the same as
in the previous test case, indicating a very robust generalization of the policy.
\paragraph{Partially Observable (PO)} In the PO case 
(cf. \cref{fig:ComparionBellBurgersPartiallyObservableStateControl}), the baseline model clearly
outperforms the
\ourAlgo{} method. While for both values of $\kDyn = 5,10$ the agent is able to stabilize the PDE
towards a steady state and slowly decreases it is constant, only the baseline agent fully converges towards
to the desired zero-state. Interestingly, the \ourAlgo{} controllers seem to focus on channel-wise 
high and low control values, while the full-order baseline agents exhibit a switching behavior
after circa three seconds for seven out of eight actuators. Similar to the case of the uniform
initial condition, the \ourAlgo{} with $\kDyn = 10$ does not seem to rely on actuator two.
\begin{figure}[h!]
	\centering
	\begin{subfigure}[t]{0.39\textwidth}
		\centering
		\includegraphics[width=\linewidth, keepaspectratio]{
			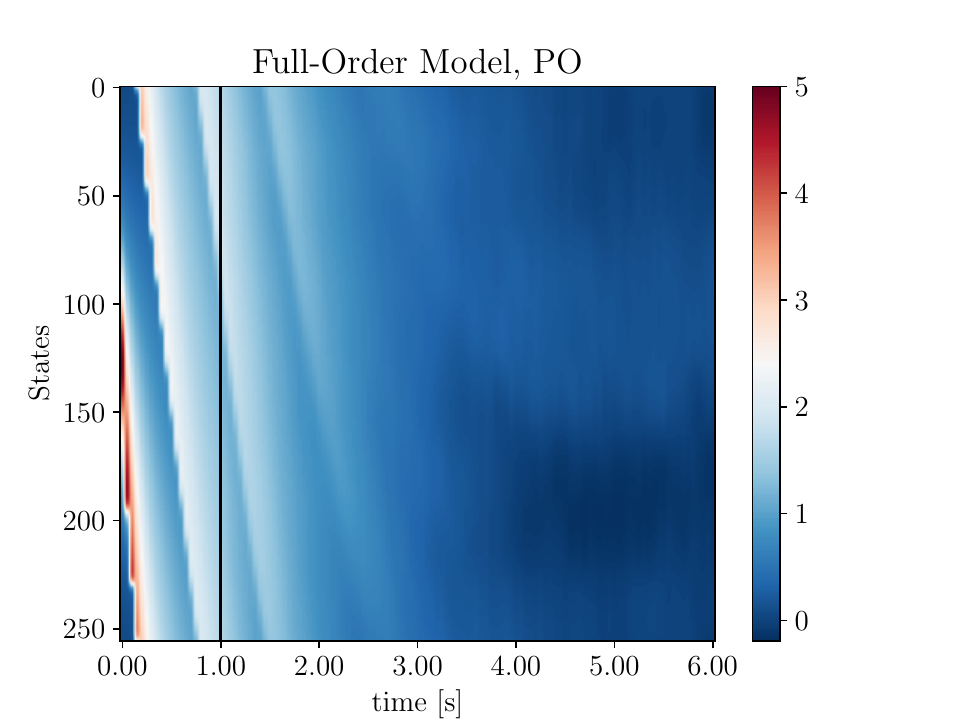
		}
		\caption{FOM: states, partially observable}
		\label{fig:BellBurgersFullOrderModelpartiallyObservableState}
	\end{subfigure}
	\hfill
	\begin{subfigure}[t]{0.39\textwidth}
		\centering
		\includegraphics[width=\linewidth, keepaspectratio]{
			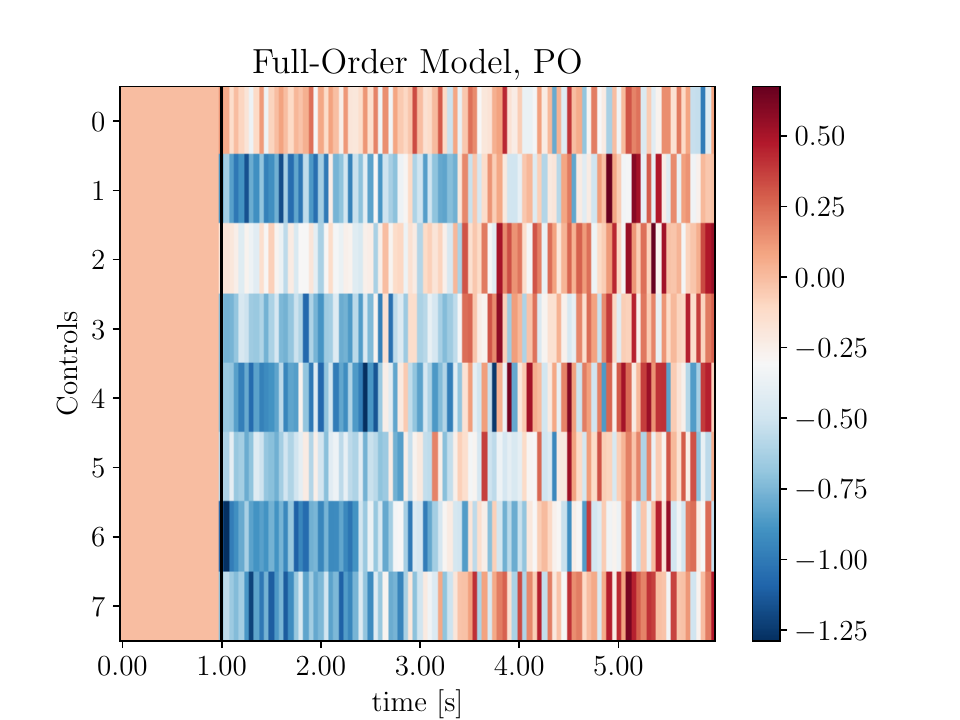
		}
		\caption{FOM: controls, partially observable}
		\label{fig:BellBurgersFullOrderModelpartiallyObservableControl}
	\end{subfigure}
    \hfill
    \begin{subfigure}[t]{0.39\textwidth}
	%	\centering
		\includegraphics[width=\linewidth,keepaspectratio]{
			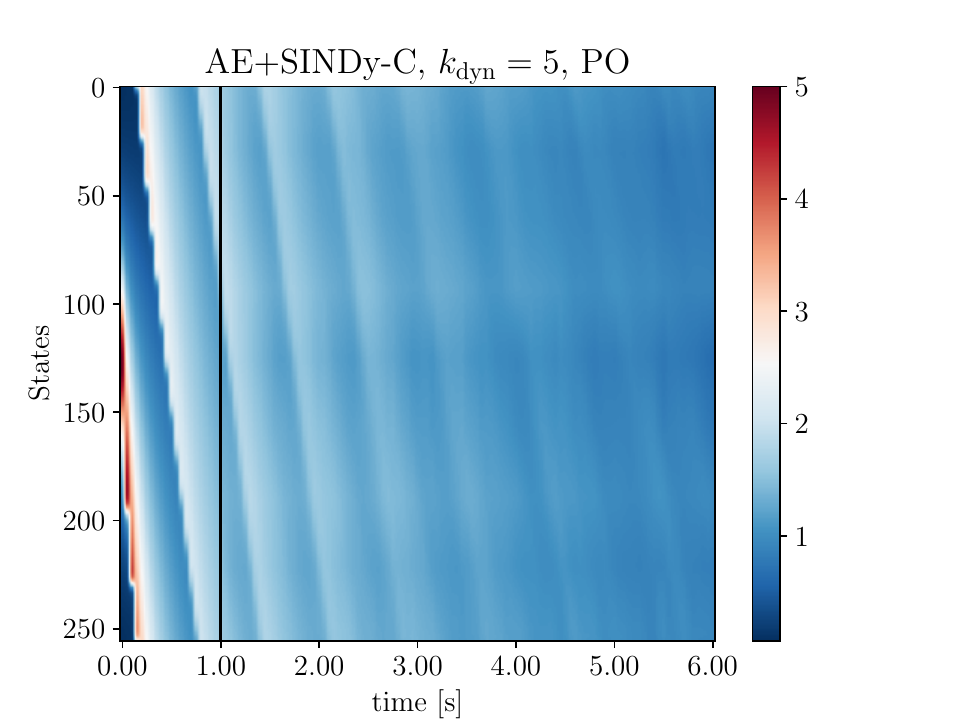
		}
		\caption{\ourAlgo{} with $\kDyn = 5$: states, partially observable}
		\label{fig:BellBurgersAutoencoderpartiallyObservableStateFit5}
	\end{subfigure}
	\hfill
	\begin{subfigure}[t]{.39\textwidth}
		\centering
		\includegraphics[width=\linewidth,keepaspectratio]{
			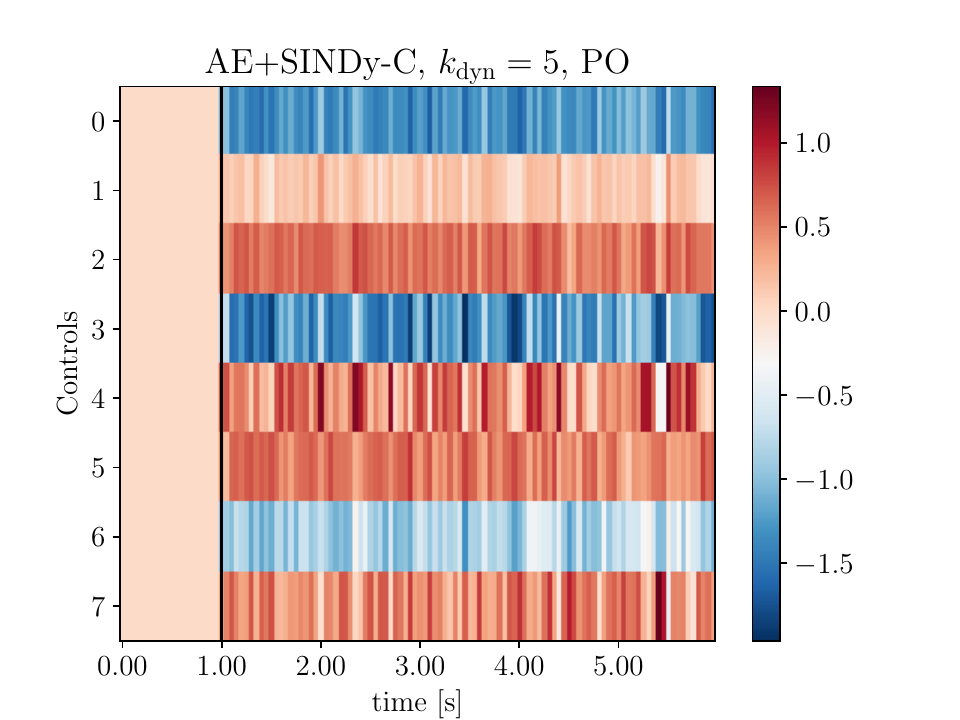
		}
		\caption{\ourAlgo{} with $\kDyn = 5$: controls, partially observable}
		\label{fig:BellBurgersAutoencoderPartiallyObservableControlFit5}
	\end{subfigure}
    \hfill
    \begin{subfigure}[t]{0.39\textwidth}
		\centering
		\includegraphics[width=\linewidth,keepaspectratio]{
			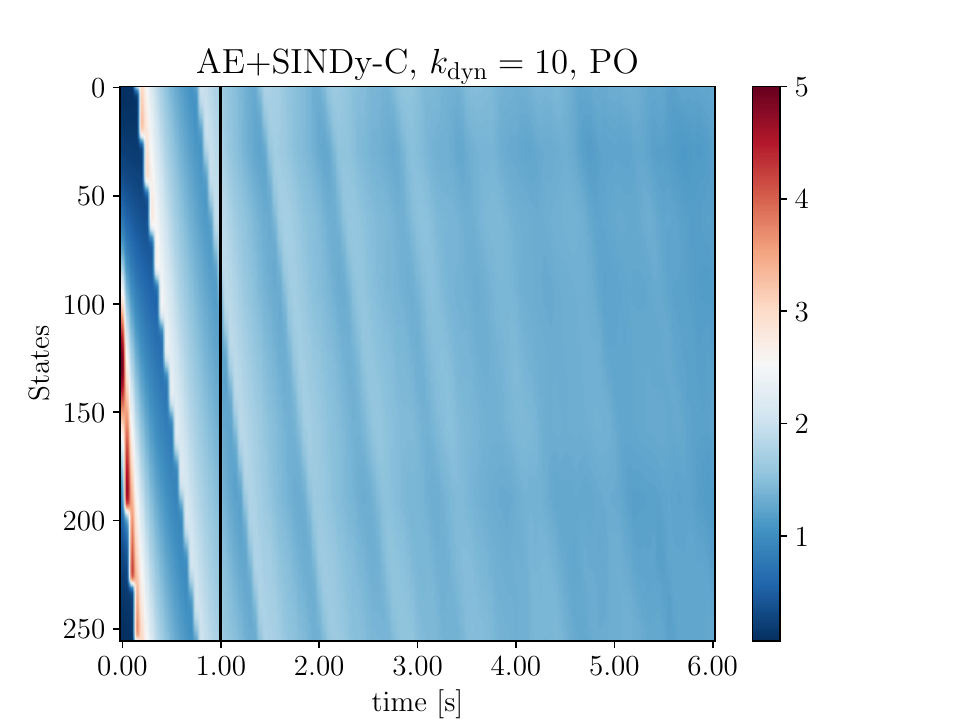
		}
		\caption{\ourAlgo{} with $\kDyn = 10$: states, partially observable}
		\label{fig:BellBurgersAutoencoderpartiallyObservableStateFit10}
	\end{subfigure}
	\hfill
	\begin{subfigure}[t]{.39\textwidth}
		\centering
		\includegraphics[width=\linewidth,keepaspectratio]{
			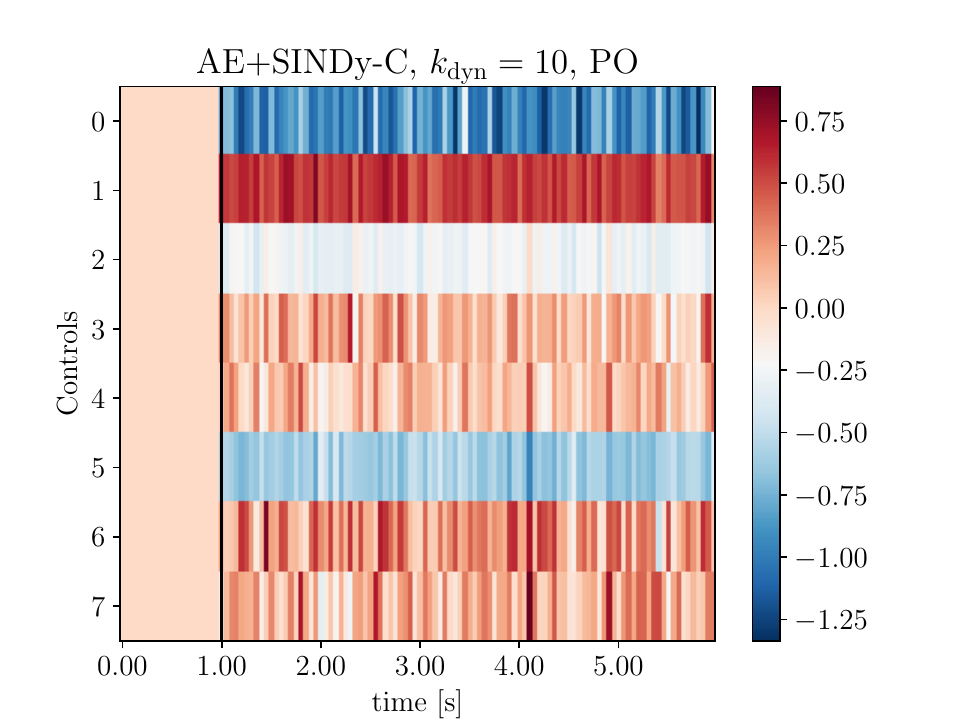
		}
		\caption{\ourAlgo{} with $\kDyn = 5$: controls, partially observable}
		\label{fig:BellBurgersAutoencoderPartiallyObservableControlFit10}
	\end{subfigure}
	\caption{State and control trajectories for the \textbf{Burgers'} equation in the
	\textbf{partially observable} (PO) case. The initial condition is a \textbf{bell-shape} hyperbolic
	cosine (\cref{eq:BellShapeInitialCondition} with $\alpha=0.5$ fixed), we use $\nu = 0.01$
	(two orders of magnitude smaller compared to the training phase),
	and the black \textbf{solid line} indicates the timestep $t$ when the \textbf{controller}
	is \textbf{activated}.}
	\label{fig:ComparionBellBurgersPartiallyObservableStateControl}
    %\hfill
    %\caption{\textcolor{red}{Here a caption! :-)}}
    %\label{fig:ComparionBellBurgersFullyObservableStateControl}
\end{figure}%
%\note{TODO(Florian): there is sth off 
%with the figure order} \textcolor{blue}{Andrea: Should be fixed by moving it upper.}
\iffalse
\begin{figure}[h!]\ContinuedFloat
	%\hfill
	\begin{subfigure}{0.48\textwidth}
		\centering
		\includegraphics[width=\linewidth,keepaspectratio]{
			./data/paper_evaluation/burgers/fom_interact/plots/BELL_AE_PO_10_states_heatmap.pdf
		}
		\caption{\ourAlgo{} with $\kDyn = 10$: states, partially observable}
		\label{fig:BellBurgersAutoencoderpartiallyObservableStateFit10}
	\end{subfigure}
	\hfill
	\begin{subfigure}{.48\textwidth}
		\centering
		\includegraphics[width=\linewidth,keepaspectratio]{
			./data/paper_evaluation/burgers/fom_interact/plots/BELL_AE_PO_10_controls_heatmap.pdf
		}
		\caption{\ourAlgo{} with $\kDyn = 5$: controls, partially observable}
		\label{fig:BellBurgersAutoencoderPartiallyObservableControlFit10}
	\end{subfigure}
	\caption{State and control trajectories for the \textbf{Burgers'} equation in the
	\textbf{partially observable} (PO) case. The initial condition is a \textbf{bell-shape} hyperbolic
	cosine (\cref{eq:BellShapeInitialCondition} with $\alpha=0.5$ fixed), we use $\nu = 0.01$
	(two orders of magnitude smaller compared to the training phase),
	and the black \textbf{solid line} indicates the timestep $t$ when the \textbf{controller}
	is \textbf{activated}.}
	\label{fig:ComparionBellBurgersPartiallyObservableStateControl}
\end{figure}
\fi
\paragraph{Fully Observable (FO)} As in the PO case, the FO test case (cf.
\cref{fig:ComparisonBellBurgersFullyObservableStateControl}) shows a similar pattern. The
baseline model outperforms both the \ourAlgo{} agents. In contrast to the PO case,
now also the two agents trained with the surrogate model are close to the zero-state
target, with $\kDyn = 5$ slightly outperforming the $\kDyn = 10$ agent. Notably,
parts of the controls in all models exhibit zero-like actuators, the baseline model again
showing a switching behavior after circa three and a half seconds. It seems like having access to
more state measurements requires the usage of less actuators, resulting in a higher reward,
as displayed in \cref{tab:ComparisonBurgerNumbers}. For real-world applications, this can be
really useful when designing position and size of controllers for an actual system.

\begin{figure}[h!]
	\centering
	\begin{subfigure}[t]{0.39\textwidth}
		\centering
		\includegraphics[width=\linewidth, keepaspectratio]{
			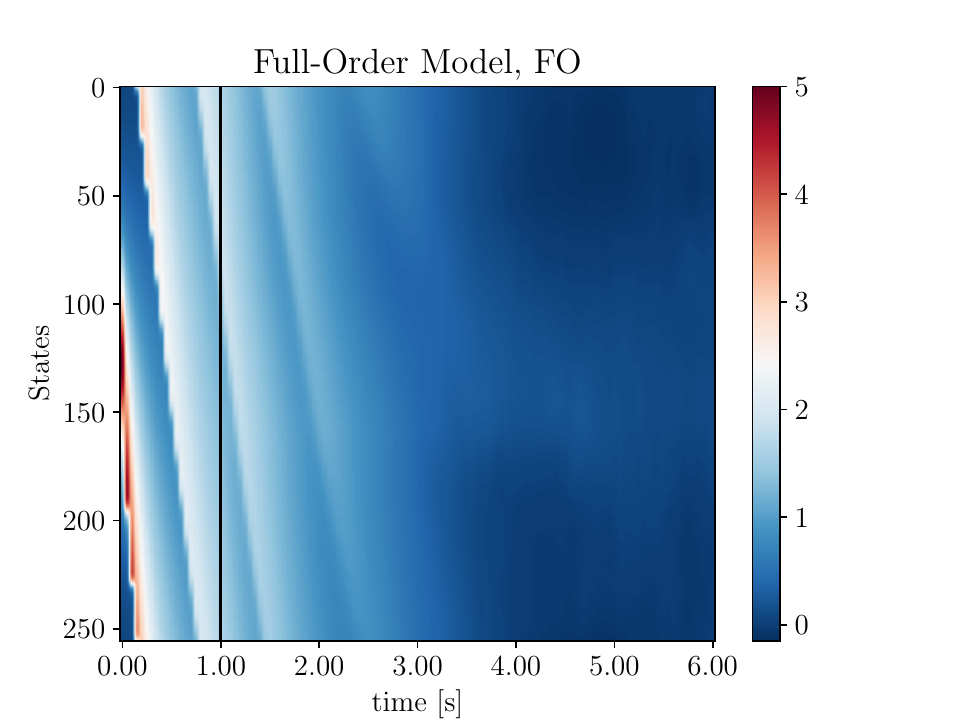
		}
		\caption{FOM: states, fully observable}
		\label{fig:BellBurgersFullOrderModelFullyObservableState}
	\end{subfigure}
	\hfill
	\begin{subfigure}[t]{0.39\textwidth}
		\centering
		\includegraphics[width=\linewidth, keepaspectratio]{
			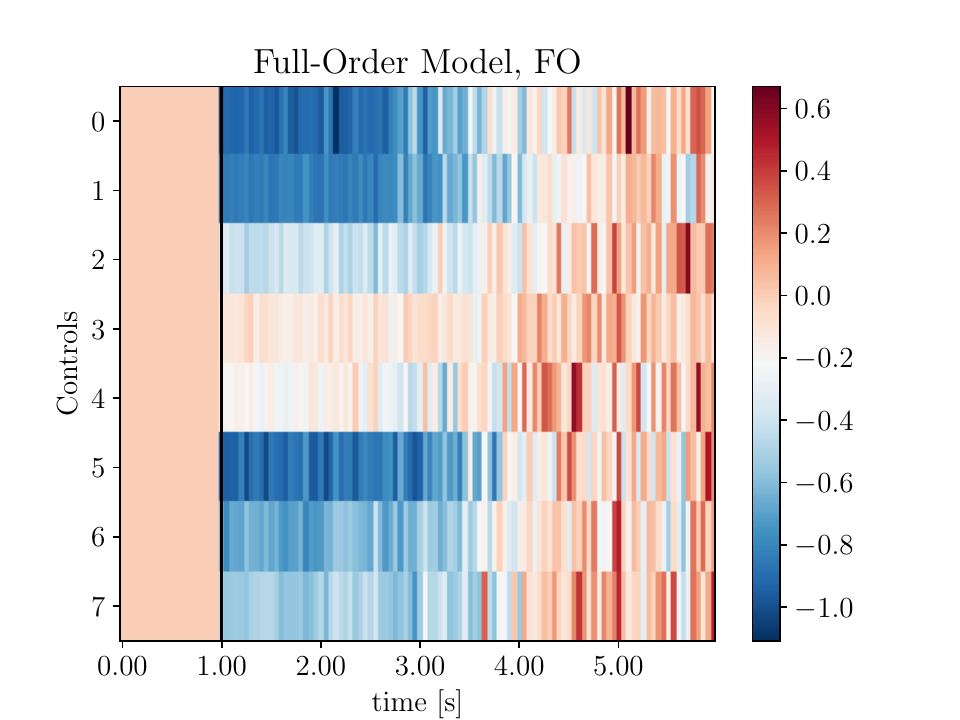
		}
		\caption{FOM: controls, fully observable}
		\label{fig:BellBurgersFullOrderModelFullyObservableControl}
	\end{subfigure}
    %\end{figure}%
    %\begin{figure}[ht]\ContinuedFloat
	\hfill
	\begin{subfigure}{0.39\textwidth}
		\centering
		\includegraphics[width=\linewidth,keepaspectratio]{
			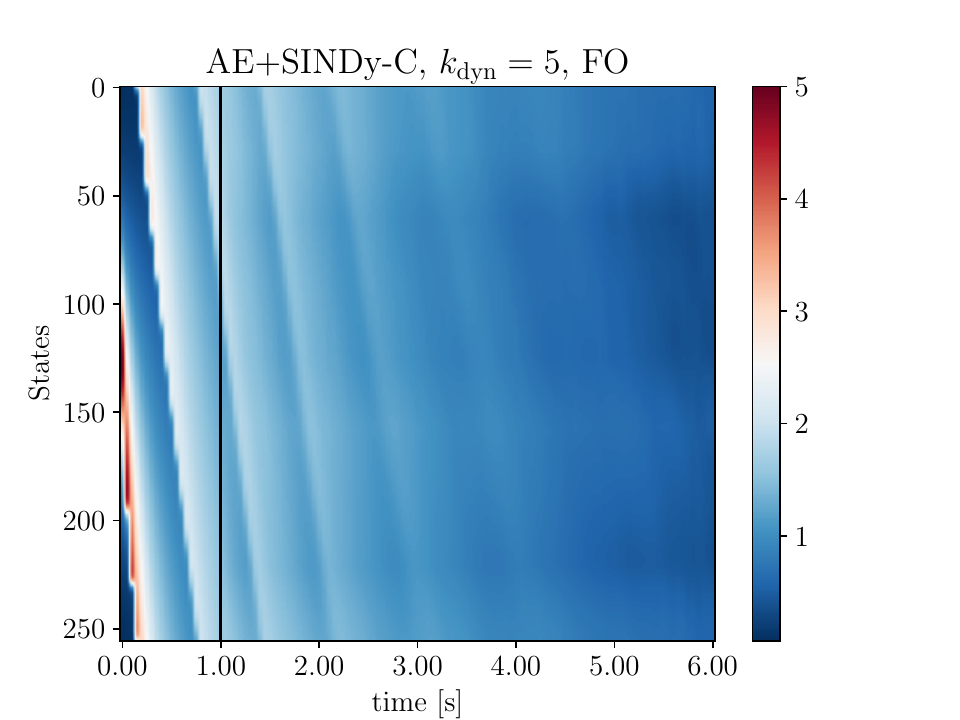
		}
		\caption{\ourAlgo{} with $\kDyn = 5$: states, fully observable}
		\label{fig:BellBurgersAutoencoderFullyObservableStateFit5}
	\end{subfigure}
	\hfill
	\begin{subfigure}{.39\textwidth}
		\centering
		\includegraphics[width=\linewidth,keepaspectratio]{
			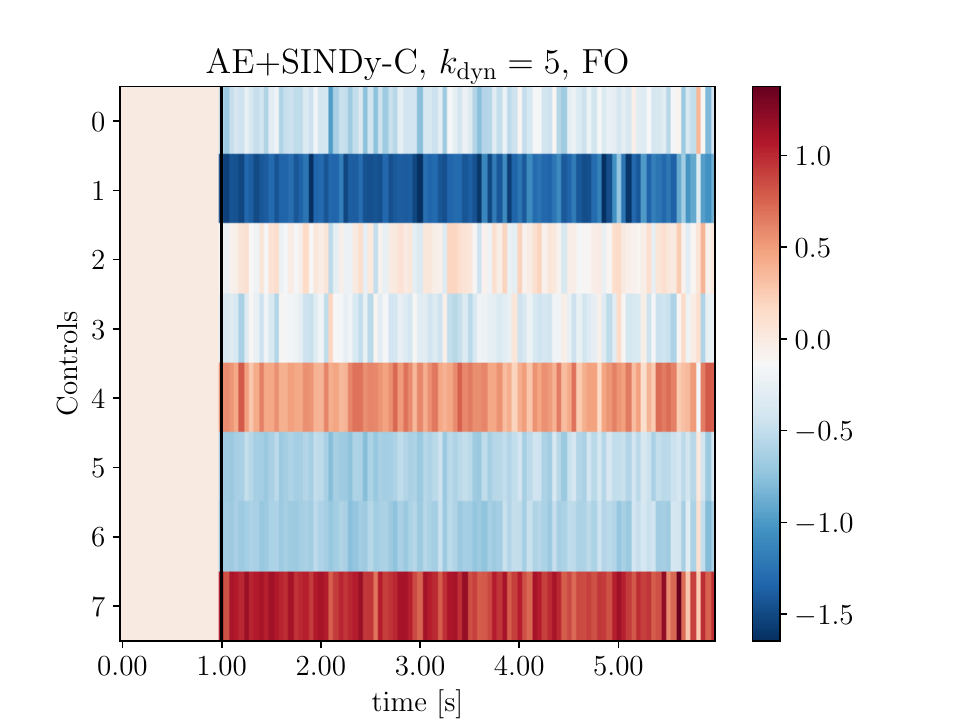
		}
		\caption{\ourAlgo{} with $\kDyn = 5$: controls, fully observable}
		\label{fig:BellBurgersAutoencoderFullyObservableControlFit5}
	\end{subfigure}
    \hfill	
 \begin{subfigure}{0.39\textwidth}
		\centering
		\includegraphics[width=\linewidth,keepaspectratio]{
			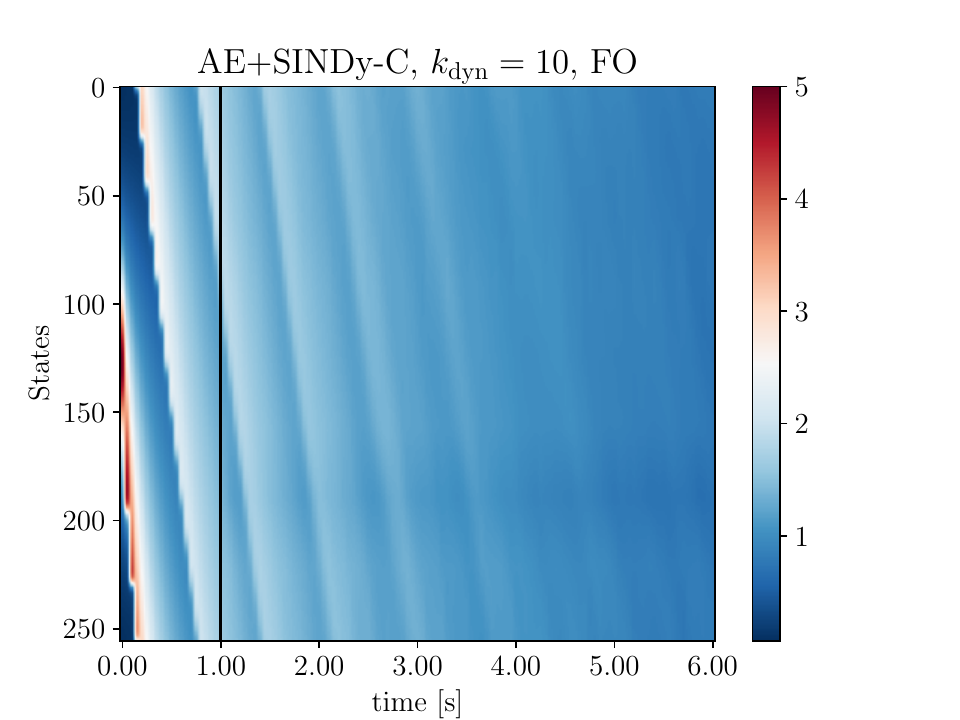
		}
		\caption{\ourAlgo{} with $\kDyn = 10$: states, fully observable}
		\label{fig:BellBurgersAutoencoderFullyObservableStateFit10}
	\end{subfigure}%
	\hfill
	\begin{subfigure}{.39\textwidth}
		\centering
		\includegraphics[width=\linewidth,keepaspectratio]{
			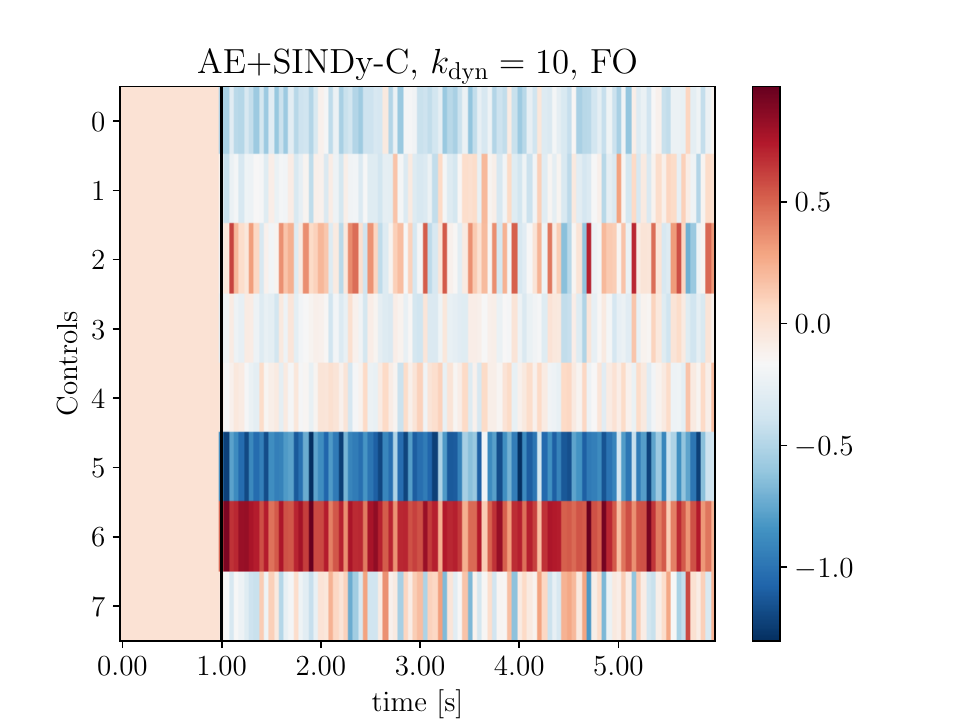
		}
		\caption{\ourAlgo{} with $\kDyn = 10$: controls, fully observable}
		\label{fig:BellBurgersAutoencoderFullyObservableControlFit10}
	\end{subfigure}%
	\caption{State and control trajectories for the \textbf{Burgers'} equation in the
	\textbf{fully observable} (FO) case. The initial condition is a \textbf{bell-shape} hyperbolic
	cosine (\cref{eq:BellShapeInitialCondition} with $\alpha=0.5$ fixed), we use $\nu = 0.01$
	(two orders of magnitude smaller compared to the training phase),
	and the black \textbf{solid line} indicates the timestep $t$ when the \textbf{controller}
	is \textbf{activated}.}
	\label{fig:ComparisonBellBurgersFullyObservableStateControl}
\end{figure}

\subsubsection{Dynamics in the Latent Space} %Surrogate Space}
\label{sec:DynamicsSurrogateSpaceBurgers}
An important aspect of the \ourAlgo{} method is the compression of high-dimensional PDE
discretization into a low-dimensional surrogate space, resulting in a closed-form and
interpretable representation and the potential to discover unknown system dynamics.
The coefficient matrices $\vec{\Xi} \in \RR^{d \times \dimObsX}$ are visualized in
\cref{fig:BurgersXiMatrixComparison}  for the Burgers' case, % \textcolor{red}{(which of the two?)}, 
highlighting the coefficients as a heatmap as well as
some key points of interests (e.g. sparsity).

Based on the coefficients $\vec{\Xi}$, we compute the closed-form representation of the
\SindyC{} dynamics model in the surrogate space. Since our algorithm is not trained
with sequential thresholding, we chose a threshold of $0.15$ to cut-off non-significant
contributions and improve readability. We obtain the following surrogate space dynamics:
\begin{itemize}
	\item Partially observable, $\kDyn = 5$:
		\begin{align*}
			\vec{z}_{x,1}(t+1) = &- 0.688 \cdot \vec{z}_{x,1}(t) \vec{z}_{x,2}(t) + 0.497 \cdot \vec{z}_{x,1}(t)^3\\
				&- 0.195 \cdot \vec{z}_{x,1}(t)^2 \vec{z}_{x,2}(t) + 0.288 \cdot \vec{z}_{u,1}(t)\\
			\vec{z}_{x,2}(t+1) = &+ 0.297 \cdot \vec{z}_{x,2}(t) - 0.180 \cdot \vec{z}_{x,1}(t)^2 - 0.245 \cdot \vec{z}_{x,1}(t) \vec{z}_{x,2}(t)\\
				&- 0.381 \cdot \vec{z}_{x,2}(t)^2 + 0.304 \cdot \vec{z}_{x,1}(t)^2 \vec{z}_{x,2}(t) + 0.487 \cdot \vec{z}_{u,1}(t).
		\end{align*}
	\item Partially observable, $\kDyn = 10$:
		\begin{align*}
			\vec{z}_{x,1}(t+1) = &+ 0.470 \cdot \vec{z}_{x,1}(t) + 0.286 \cdot \vec{z}_{x,2}(t) - 0.295 \cdot \vec{z}_{x,1}(t)^3\\
				&- 0.257 \cdot \vec{z}_{x,1}(t) \vec{z}_{x,2}(t)^2 + 0.267 \cdot \vec{z}_{u,2}(t)\\
			\vec{z}_{x,2}(t+1) = &+ 0.689 \cdot \vec{z}_{x,2}(t) - 0.278 \cdot \vec{z}_{x,1}(t)^2 \vec{z}_{x,2}(t)\\
				&- 0.303 \cdot \vec{z}_{x,2}(t)^3 + 0.153 \cdot \vec{z}_{u,2}(t).
		\end{align*}
	\item Fully observable, $\kDyn = 5$:
		\begin{align*}
			\vec{z}_{x,1}(t+1) = &+ 0.795 \cdot \vec{z}_{x,1}(t)^2 + 0.420 \cdot \vec{z}_{x,1}(t) \vec{z}_{x,2}(t) - 0.676 \cdot \vec{z}_{x,1}(t)^3\\
				&+ 0.224 \cdot \vec{z}_{x,1}(t)^2 \vec{z}_{x,2}(t) + 0.159 \cdot \vec{z}_{u,2}(t)\\
			\vec{z}_{x,2}(t+1) = &+ 0.310 \cdot \vec{z}_{x,1}(t)^3 - 0.454 \cdot \vec{z}_{x,1}(t)^2 \vec{z}_{x,2}(t)\\
				&+ 0.655 \cdot \vec{z}_{x,1}(t) \vec{z}_{x,2}(t)^2 + 0.426 \cdot \vec{z}_{x,2}(t)^3.
		\end{align*}
	\item Fully observable, $\kDyn = 10$:
		\begin{align*}
			\vec{z}_{x,1}(t+1) = &+ 0.328 \cdot \vec{z}_{x,1}(t) + 0.402 \cdot \vec{z}_{x,2}(t) - 0.220 \cdot \vec{z}_{x,2}(t)^2 \\
				&+ 0.532 \cdot \vec{z}_{x,1}(t)^2 \vec{z}_{x,2}(t) - 0.429 \cdot \vec{z}_{x,2}(t)^3 + 0.304 \cdot \vec{z}_{u,2}(t)\\
			\vec{z}_{x,2}(t+1) = &+ 0.566 \cdot \vec{z}_{x,1}(t) + 0.253 \cdot \vec{z}_{x,1}(t) \vec{z}_{x,2}(t) + 0.197 \cdot \vec{z}_{x,1}(t)^3\\
				&- 0.640 \cdot \vec{z}_{x,1}(t)^2 \vec{z}_{x,2}(t) + 0.344 \cdot \vec{z}_{x,2}(t)^3 + 0.278 \cdot \vec{z}_{u,2}(t).
		\end{align*}
\end{itemize}
Denoting by $\vec{z}_{x,i}$ and $\vec{z}_{u,i}$  the $i$-th component of the state and the control
control in the latent space, respectively, SINDy individually compressed both state and action into a two-dimensional latent space representation. While for the state the corresponding AE requires both dimensions, the control AE correctly compressed the control into a one-dimensional representation. Overall, each of the
representations is different, which can be explained by (a) a different training procedure, (b) different
full-order interactions observed by the models, and (c) the non-uniqueness of the representation.

\begin{figure}[H]
	\centering
	\begin{subfigure}[t]{0.48\textwidth}
		\centering
		\includegraphics[width=\linewidth,keepaspectratio]{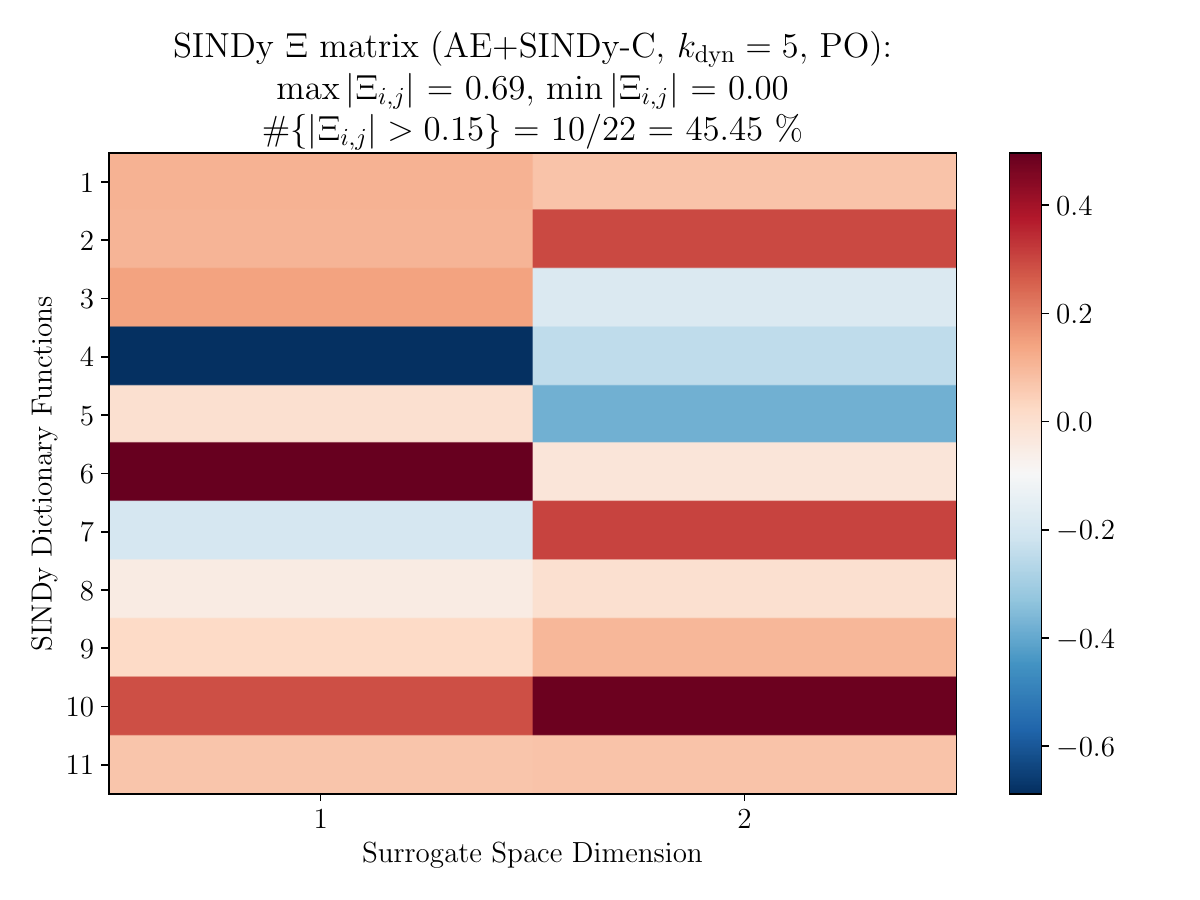}
		\caption{Partially observable case, $\kDyn = 5$}
		\label{fig:BurgersXiMatrixPartiallyObservableFit5}
	\end{subfigure}
	\hfill
	\begin{subfigure}[t]{0.48\textwidth}
		\centering
		\includegraphics[width=\linewidth,keepaspectratio]{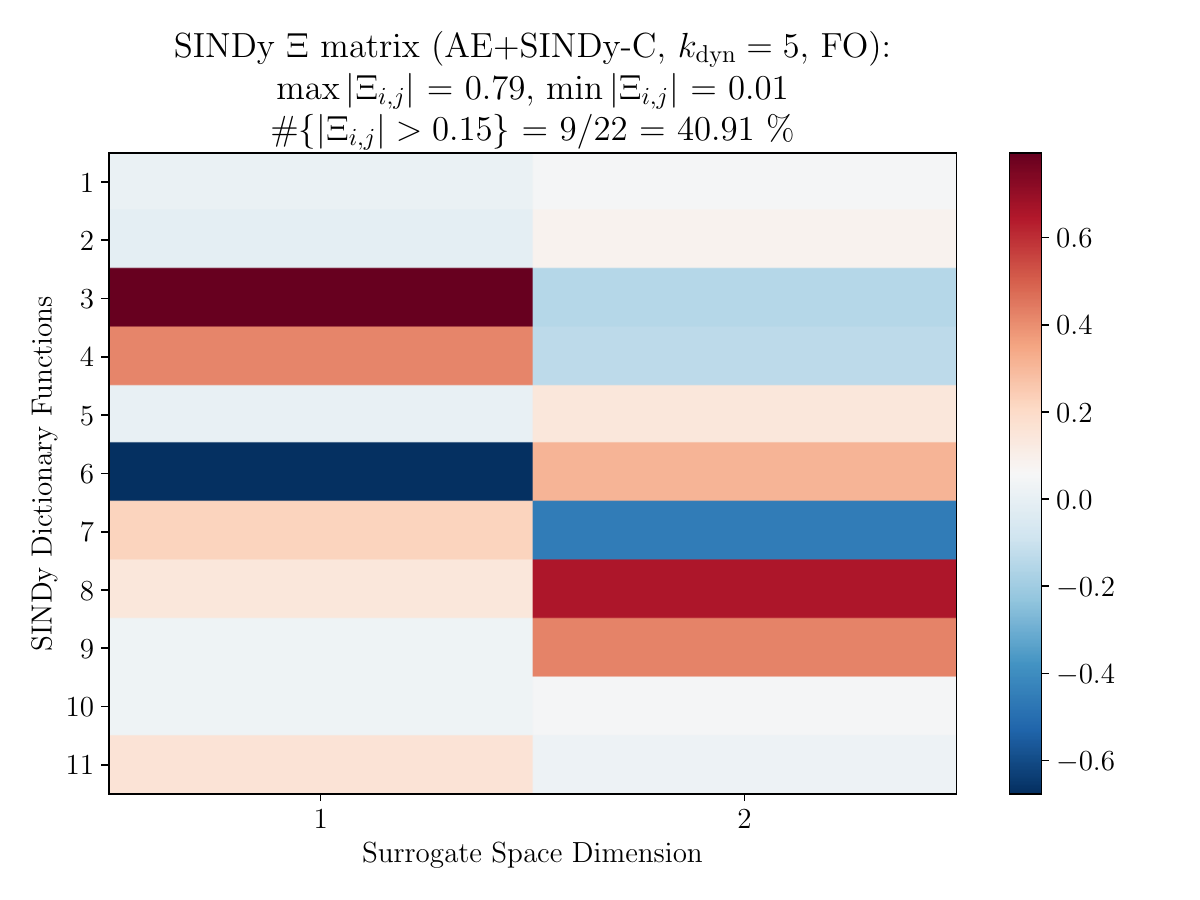}
		\caption{Fully Observable Case, $\kDyn = 5$}
		\label{fig:BurgersXiMatrixFullyObservableFit5}
	\end{subfigure}
	%\hfill
    \end{figure}%
    \begin{figure}[ht]\ContinuedFloat
	\begin{subfigure}[t]{0.48\textwidth}
		\centering
		\includegraphics[width=\linewidth,keepaspectratio]{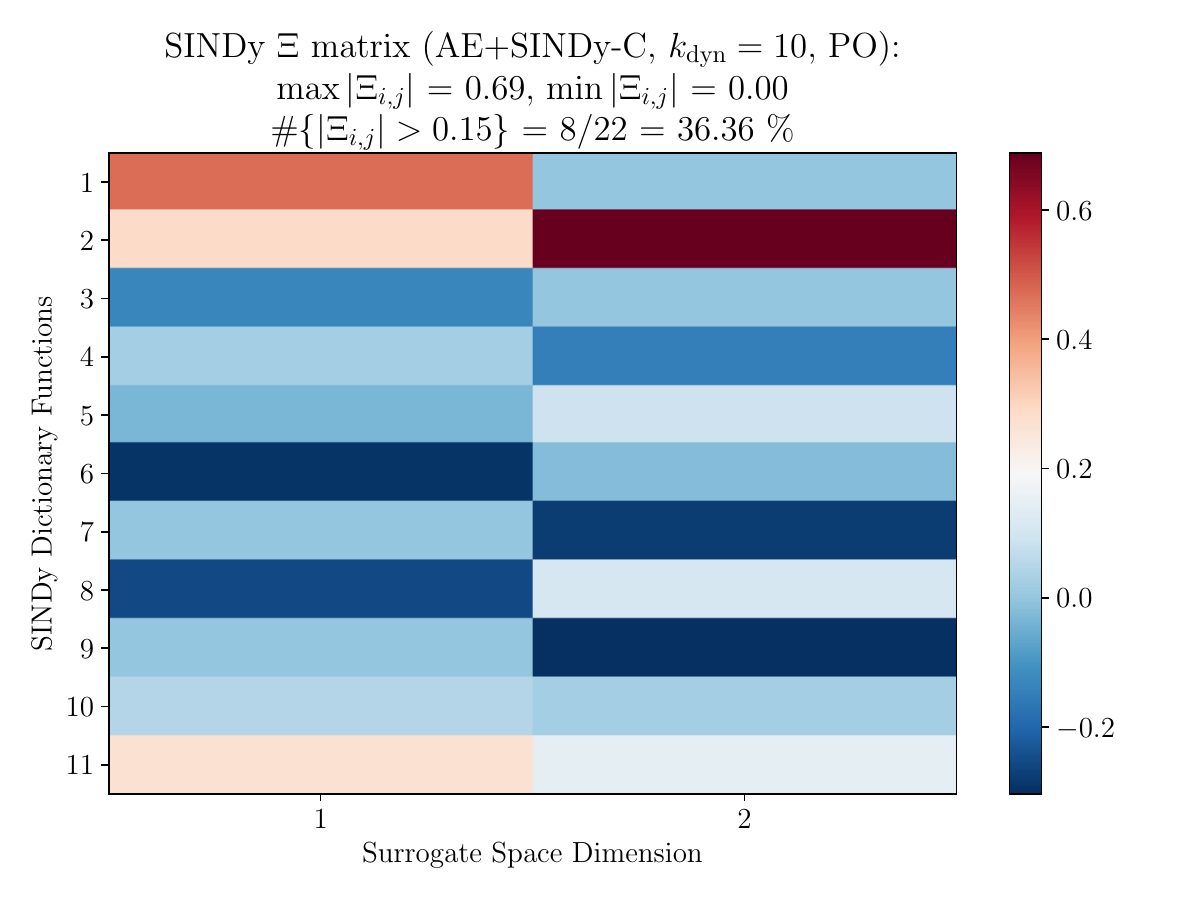}
		\caption{Partially observable case, $\kDyn = 10$}
		\label{fig:BurgersXiMatrixPartiallyObservableFit10}
	\end{subfigure}
	\hfill
	\begin{subfigure}[t]{0.48\textwidth}
		\centering
		\includegraphics[width=\linewidth,keepaspectratio]{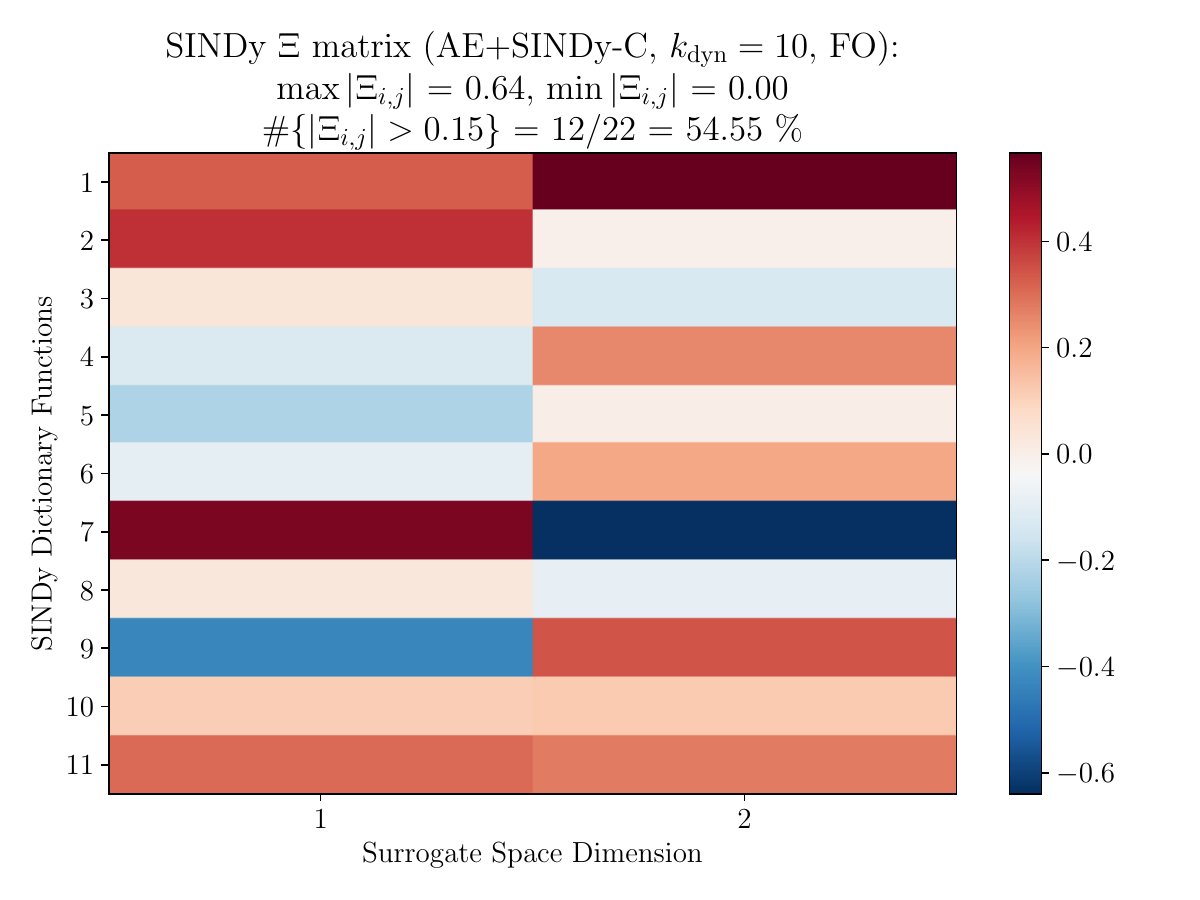}
		\caption{Fully Observable Case, $\kDyn = 10$}
		\label{fig:BurgersXiMatrixFullyObservableFit10}
	\end{subfigure}
	\caption{Analysis of the coefficient matrix $\vec{\Xi} \in \RR^{d \times \dimObsX}$ for the
	\textbf{Burgers'} equation.}
	\label{fig:BurgersXiMatrixComparison}
\end{figure}

\subsubsection{Training of \ourAlgo{}}
\label{sec:5AnalysisTrainingAutoencoder}
To analyze the goodness of fit of the internal surrogate model, we provide in 
\cref{tab:BurgersTrainingTimeErrorAnalysis} the average training and validation error, i.e. \cref{eq:AutoencoderLossFunction}, of the last iteration of each surrogate model update, i.e. step
\cref{algoLine:SurrogateFitEnvironment}.
We clearly see that keeping the training epochs of \ourAlgo{} low helps to reduce overfitting. Additionally,
the order of magnitude for each of the four methods is very
similar.\footnote{Internally, we use a 80/20 splitting for training
and validation and train for 100 epochs. Debugging plots show that in all
cases an upper bound of 30 epochs would be enough to train the \ourAlgo{}
representation.}
Additionally, \cref{tab:BurgersTrainingTimeErrorAnalysis} also displays
the average duration of a surrogate model update, i.e. step
\cref{algoLine:SurrogateFitEnvironment}, which is due yo the small number of
parameters.

\begin{table}[hb]
	\centering
	\scalebox{0.8}{
\begin{tabular}{lllll}
\toprule
                                         & \multicolumn{2}{c}{\ourAlgo{}, $\kDyn = 5$}                            & \multicolumn{2}{c}{\ourAlgo{}, $\kDyn = 10$}                         \\
                                         \cmidrule(rl){2-3} \cmidrule(rl){4-5}
                                         & PO $48 \times 8$                           & FO $256 \times 8$                          & PO $48 \times 8$                          & FO $256 \times 8$                         \\
\midrule
Training Time $[s]$ $(\mu \pm \sigma^2)$ & $3.80 \pm 1.21$                            & $3.80 \pm 2.55$                            & $4.55 \pm 0.71$                           & $4.11 \pm 0.41$                           \\
Loss \cref{eq:AutoencoderLossFunction} $(\mu \pm \sigma^2)$           &                                            &                                            &                                           &                                           \\
$\quad$ Training                         & $7.53 \cdot 10^{-3} \pm 1.01\cdot 10^{-4}$ & $7.27 \cdot 10^{-3} \pm 1.33\cdot 10^{-4}$ & $8.01\cdot 10^{-3} \pm 3.21\cdot 10^{-4}$ & $7.81\cdot 10^{-3} \pm 1.34\cdot 10^{-4}$ \\
$\quad$ Validation                       & $8.43 \cdot 10^{-3} \pm 2.88\cdot 10^{-4}$ & $7.44 \cdot 10^{-3} \pm 1.64\cdot 10^{-4}$ & $7.94\cdot 10^{-3} \pm 6.24\cdot 10^{-4}$ & $9.33\cdot 10^{-3} \pm 5.22\cdot 10^{-4}$\\
\bottomrule
\end{tabular}
}
	\caption{Training time and overview of the loss distribution of
    \cref{eq:AutoencoderLossFunction} during the training phase of 
    the Dyna-style \ourAlgo{} method for the \textbf{Burgers'} equation.
    Mean and variance are computed over all training epochs.
    The training was performed on a MacBook M1 (2021, 16GB RAM).
}
	\label{tab:BurgersTrainingTimeErrorAnalysis}
\end{table}

\subsection{Incompressible Navier-Stokes Equations (PDEControlGym)}
\label{subsec:NavierStokesEquation}
This second example focuses on a much
more challenging equation and control setting. To highlight the framework's capabilities
to scale to high-dimensional spaces, we consider the two-dimensional incompressible Navier-Stokes equations.
The Navier-Stokes equations play a fundamental role in fluid dynamics with numerous applications in areas
such as aerodynamics, weather forecasting, oceanography, and industrial processes.
They describe the motion of fluid substances and are used to model phenomena ranging from airflow
over an aircraft wing to blood flow in arteries and the formation of weather patterns. 

The temporal dynamics of the 2D velocity field $\vec{x}(x_1, x_2, t) : \Omega \times [0, T] \to \RR^2$ and the pressure field $p(x_1, x_2, t) : \Omega \times [0, T] \to \RR$ is given by
\begin{align}
    \begin{aligned}
        &\nabla \cdot \vec{x} = 0,\\
        &\partial_t \vec{x} + \vec{x} \cdot \nabla\vec{x} = - \frac{1}{\rho} \nabla p + \nu \nabla^2 \vec{x},
    \end{aligned}
\end{align}
in $\Omega \times [0,T]$, denoting the kinematic viscosity of the
fluid by $\nu$, and its density by $\rho$. For the experiments, the equation is fully
observable, we consider Dirichlet boundary conditions with velocities set to 0, and we control
along the top boundary $u_t = u_t(x_1) = \vec{x}(x_1, 1, t)$, for $x_1 \in \Omega$, i.e. $x_2 = 1$ 
is fixed.
Following the benchmark of \cite[section 5.3]{bhan2024pde}, the domain
$\Omega = (0,1)^2$ and the time interval $T = 0.2\mathrm{s}$ are used, parameters for the solution procedure
are kept. The equation is noise-free fully observed. The DRL agents are trained with a discretized version
of the following objective function
\begin{equation*}
    \mathcal{J}(u_t, \vec{x}) = -\frac{1}{2} \int_0^T \int_{\Omega}
        \norm{\vec{x}(x_1, x_2, t) - \vec{x}_{\mathrm{ref}}(x_1, x_2, t)}^2 \dd (x_1, x_2) \dd t +\frac{\gamma}{2} \int_0^T  \norm{u_t - u^{\mathrm{ref}}_t}^2 \dd t,
\end{equation*}
where the reference solution $\vec{x}_{\mathrm{ref}}$ is given by the resulting
velocity field when applying the controls
$u : [0,T] \to \RR, t \mapsto 3 - 5t$. For the reference controls in the objective function $u^{\mathrm{ref}}_t \equiv 2.0$ is used. 
%\textcolor{red}{Not very clear: what is $u^{\mathrm{ref}}_t?$ $3 - 5t$ or $2$?}
%The initial velocity field and the pressure field are constant and equal to 1, scaled
%by an individually uniformly sampled scalar with range $(-5, 5)$. \textcolor{red}{in which sense? Why to rescale?}

Based on the superior results in the extensive study of the previous chapter,
we focus on $\kDyn = 5$ for the Navier-Stokes equations example and compare
the Dyna-style MBRL scheme with the model-free benchmark. We want to experiment
with the dimension of the surrogate space and ideally estimate the dimension at which
the agent is not possible to effectively learn a policy anymore. For this reason,
we train \ourAlgo{} with an architecture of $(882, 84,4)\times (1,6,2)$, i.e. a
six-dimensional latent space, $(882, 52, 3)\times (1,4,1)$, i.e. a four-dimensional latent
space, for the state and controls respectively, as well as with $(882, 30, 1) \times (1,4,1)$,
i.e. a two-dimensional latent space. Less degrees of freedom in the surrogate space decrease the
goodness of fit in the surrogate space and in the case of a two-dimensional surrogate space
the agent was not able to successfully learn a policy which is why this case is excluded
from the analysis.

\subsubsection{Sample Efficiency and Scalability}
\label{sec:5AnalysisScalability}
\Cref{fig:NavierStokesFOMInteractionsPerformance} clearly highlights the sample efficiency
of \ourAlgo{} compared to the model-free baseline in the case of an eight-dimensional surrogate space.
Not only does \ourAlgo{} need approximately 5x
less data, but simultaneously outperforms the baseline -- a clear indication that the internal
dynamics model in the latent space helps the agent to understand the underlying system dynamics.
The bumps in performances of \ourAlgo{} at approximately 12k and 20k FOM interactions represent the
point where the dynamics model correctly represents the directions of the flow. Before that,
\ourAlgo{} was able to correctly control the magnitude of the flow field and adjust the controls,
but the orientation of the vector field was reversed. The four-dimensional version is not
outperforming the baseline and the final model exhibits high uncertainty,
also suffering under very fuzzy controls (cf. \cref{fig:NavierStokesControlsAESmall}).
Since the corresponding velocity-field cannot be considered
a valid solution (cf. \cref{fig:NavierStokesFlowFieldAESmall}),
it is thus excluded from the following analysis.
The plateau at around
15k interactions corresponds to very fuzzy controls and only with a big delay the baseline
model can stabilize the controls, missing at the end the correct magnitude of the flow field
(cf. \cref{fig:NavierStokesFlowFieldFOM}). Interestingly, the fuzzy overfitting at the end
of the training procedure for the Burgers' equation does not appear, which is most probably due
to simpler initial conditions and thus more regular systems dynamics, although the dynamics in
general are harder to capture, i.e. more full-order interactions are needed.

\subsubsection{Velocity Field and Control Analysis}
\label{sec:5AnalysisHigherPenalties}
\begin{figure}[h]
	\centering
	\begin{subfigure}[t]{0.59\textwidth}
		\centering
		\includegraphics[width=\linewidth,keepaspectratio]{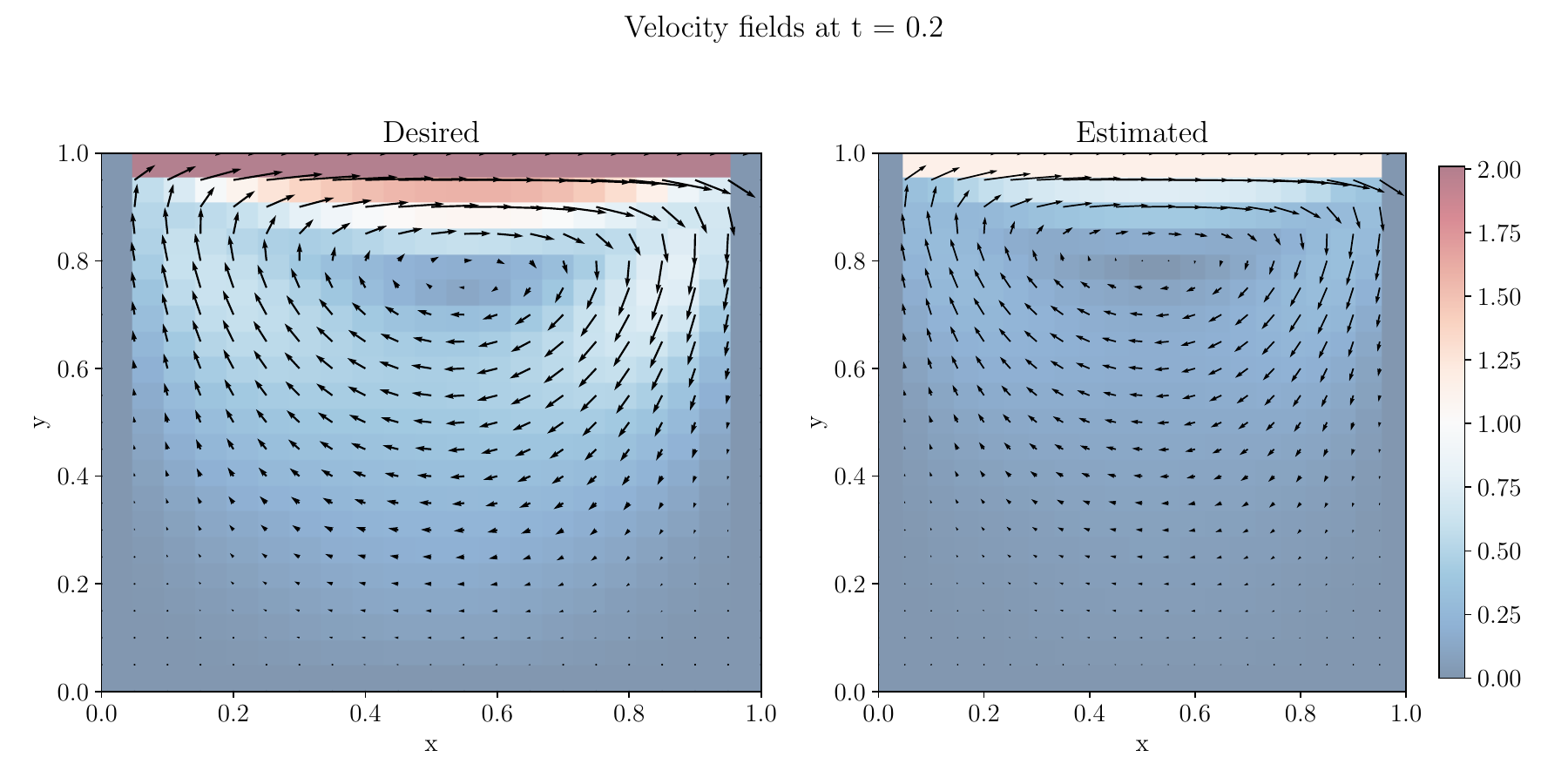}
		\caption{Full-order baseline model, Velocity field at $t=0.2$.}
		\label{fig:NavierStokesFlowFieldFOM}
	\end{subfigure}
	\hfill
	\begin{subfigure}[t]{0.39\textwidth}
		\centering
		\includegraphics[width=\linewidth,keepaspectratio]{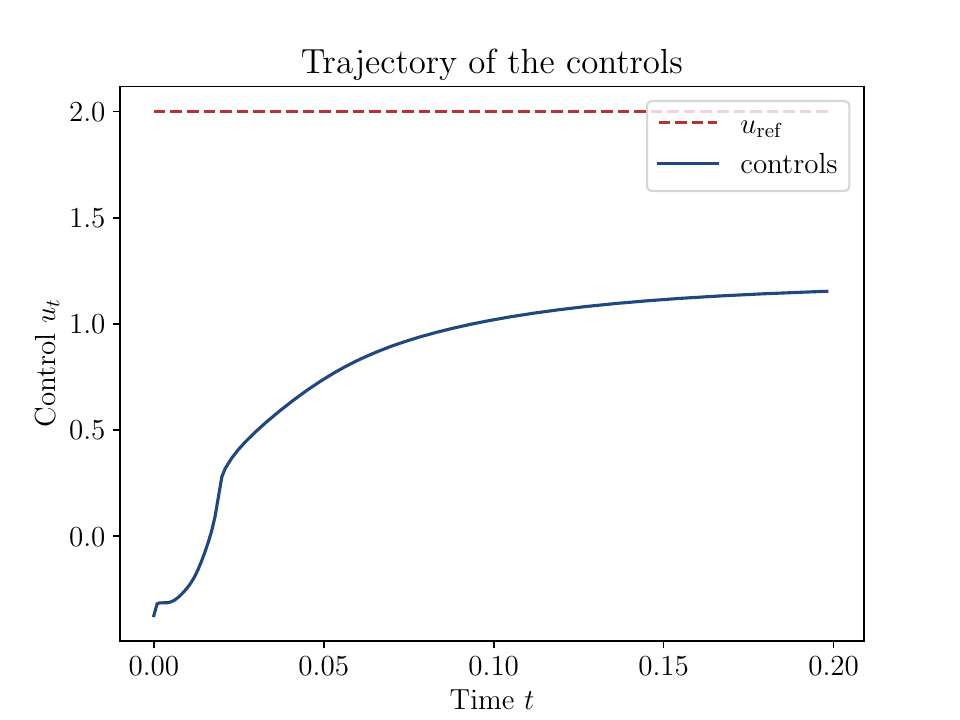}
		\caption{Full-order baseline model, control trajectory.}
		\label{fig:NavierStokesControlsFOM}
	\end{subfigure}
	%\end{figure}%
    %\begin{figure}[ht]\ContinuedFloat
    \hfill
	\begin{subfigure}[t]{0.59\textwidth}
		\centering
		\includegraphics[width=\linewidth,keepaspectratio]{
			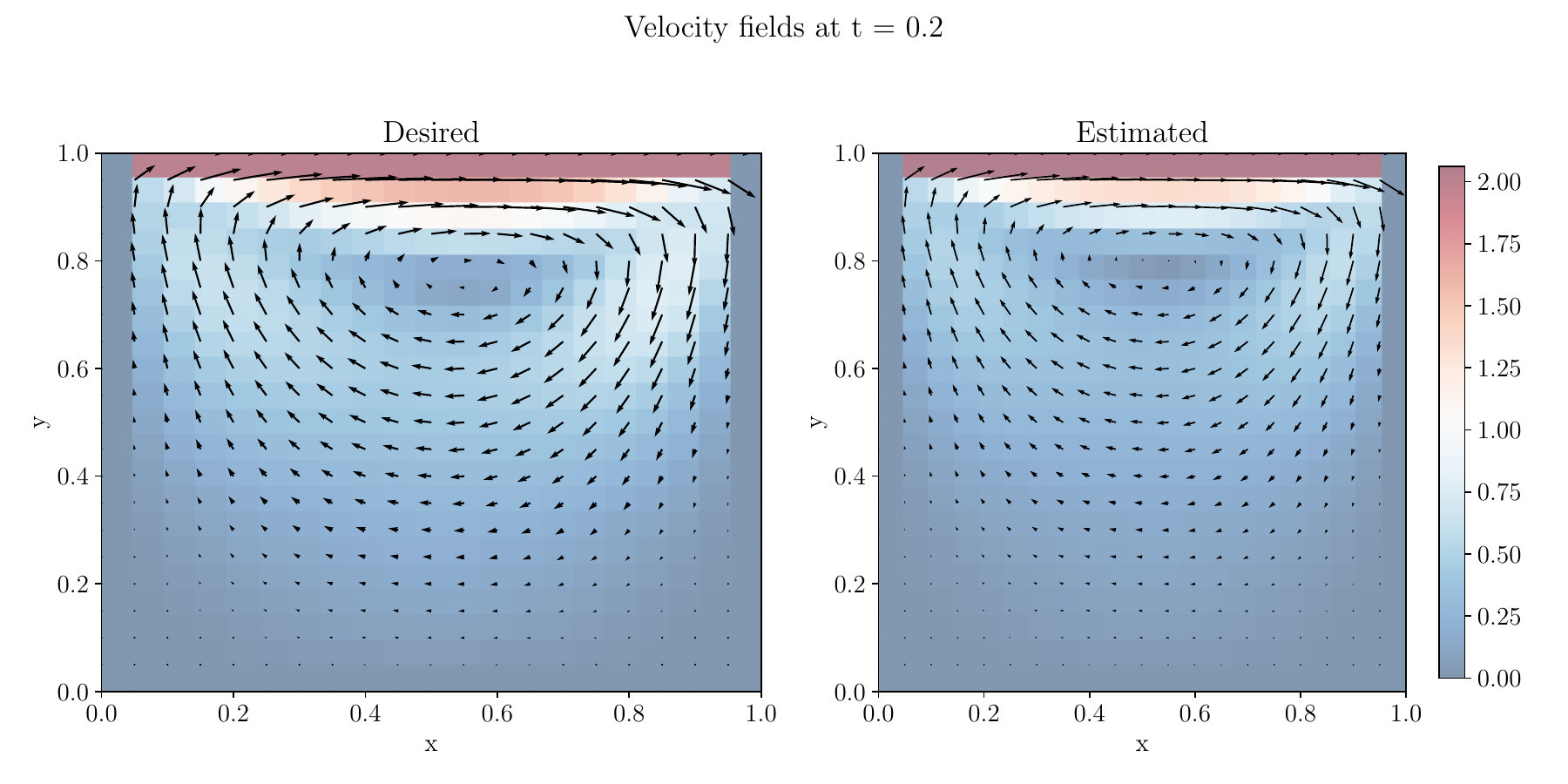}
		\caption{\ourAlgo{} with $\kDyn = 5$ and an 8-dimensional latent space, velocity field at $t=0.2$.}
		\label{fig:NavierStokesFlowFieldAE}
	\end{subfigure}
	\hfill
	\begin{subfigure}[t]{0.39\textwidth}
		\centering
		\includegraphics[width=\linewidth,keepaspectratio]{
			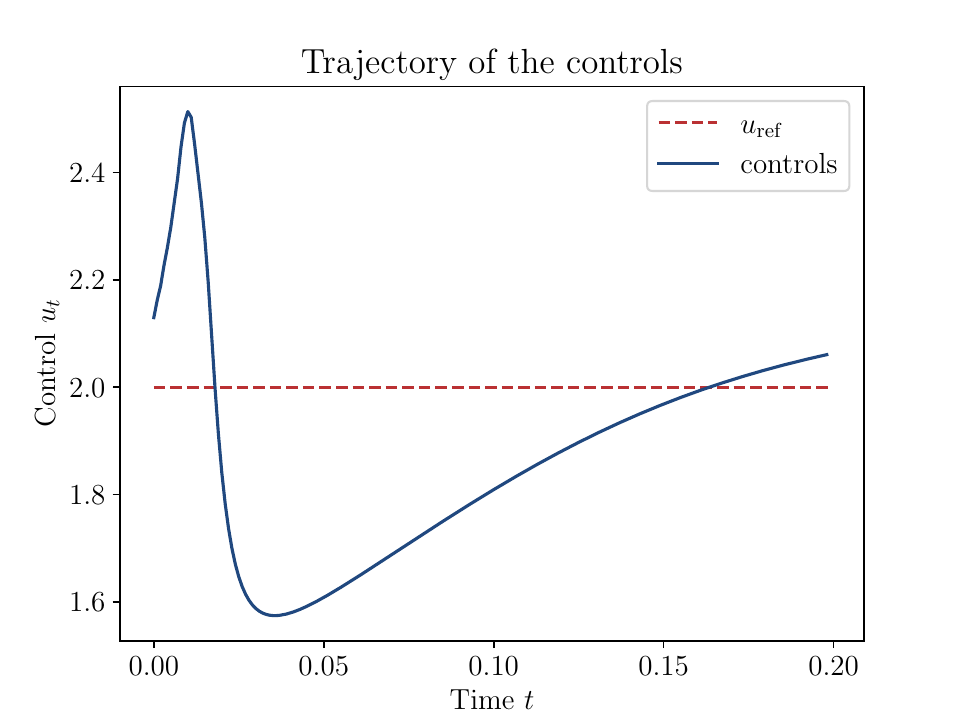}
		\caption{\ourAlgo{} with $\kDyn = 5$ and an 8-dimensional latent space, control trajectory.}
		\label{fig:NavierStokesControlsAE}
	\end{subfigure}
	\hfill
	\begin{subfigure}[t]{0.59\textwidth}
		\centering
		\includegraphics[width=\linewidth,keepaspectratio]{
			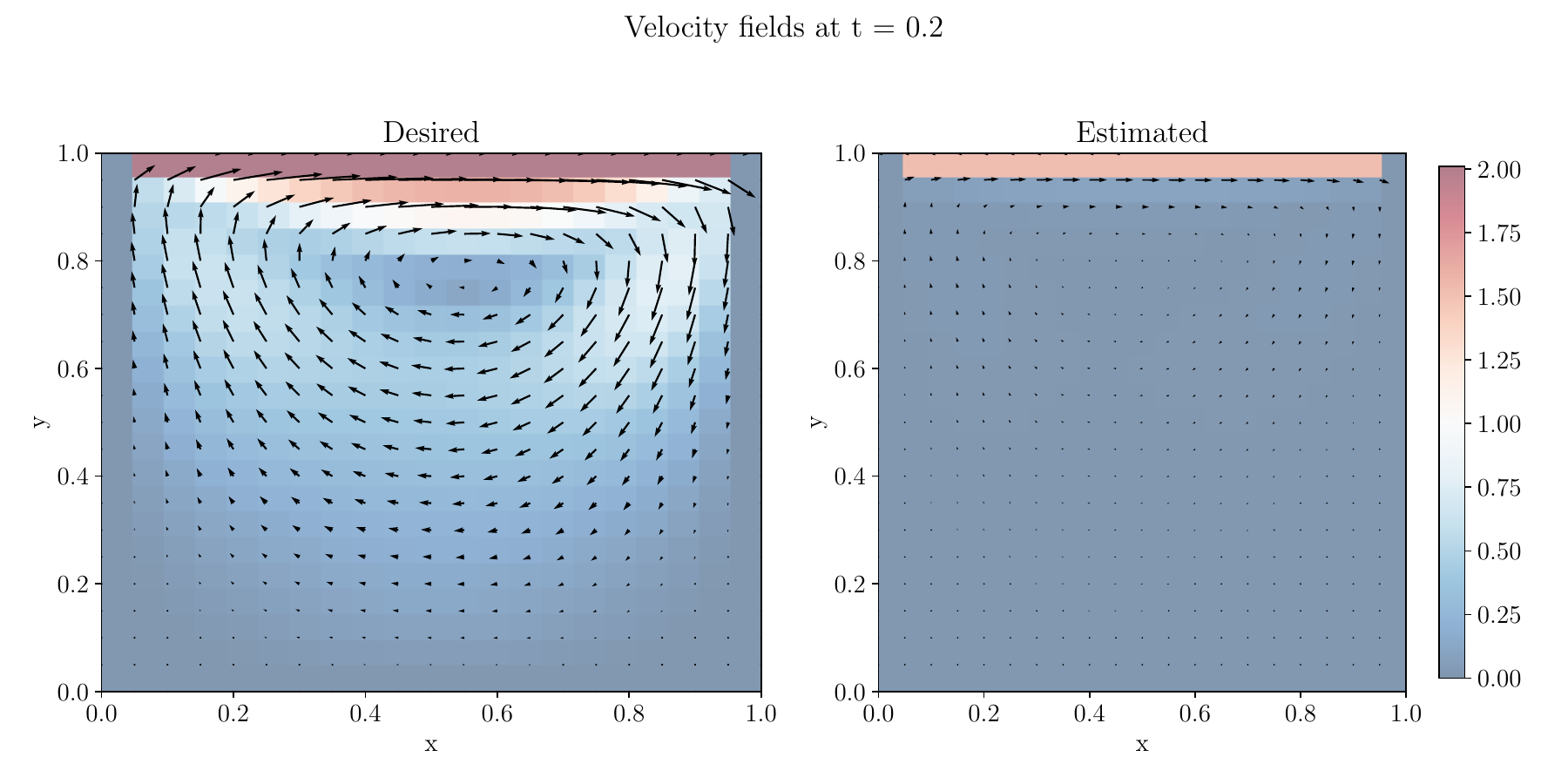}
		\caption{\ourAlgo{} with $\kDyn = 5$ and an 4-dimensional latent space, velocity field at $t=0.2$.}
		\label{fig:NavierStokesFlowFieldAESmall}
	\end{subfigure}
	\hfill
	\begin{subfigure}[t]{0.39\textwidth}
		\centering
		\includegraphics[width=\linewidth,keepaspectratio]{
			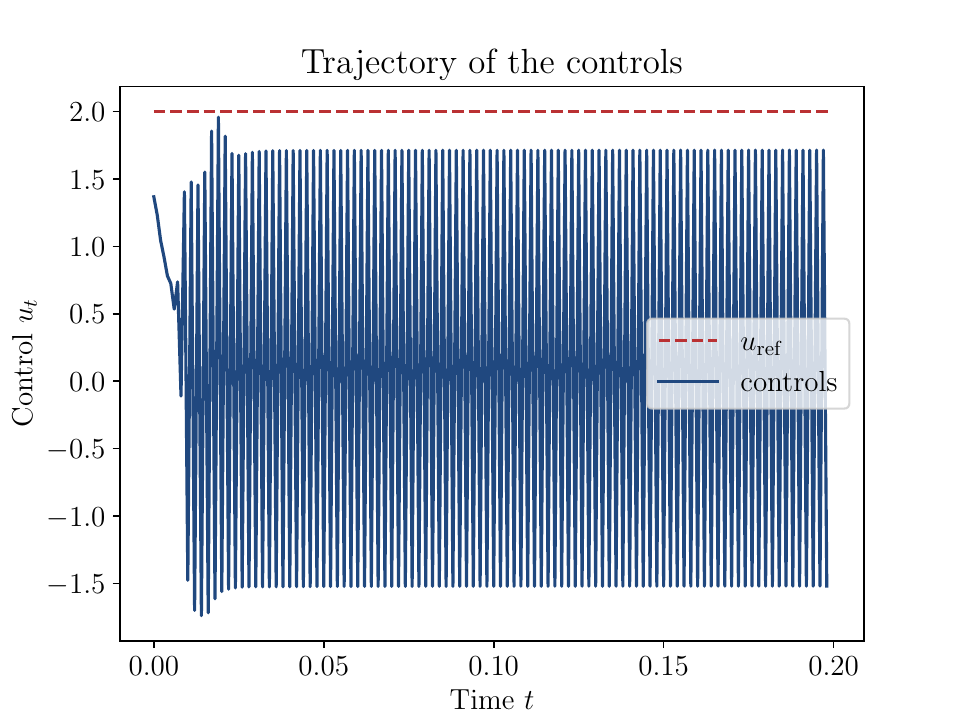}
		\caption{\ourAlgo{} with $\kDyn = 5$ and an 4-dimensional latent space, control trajectory.}
		\label{fig:NavierStokesControlsAESmall}
	\end{subfigure}
	\caption{Velocity field and control trajectories for the model-free baseline and \ourAlgo{}
		for the \textbf{Navier-Stokes} equations. Black arrows represent the velocity fields and
		the background color the magnitude of the velocity vector.} 
	\label{fig:NavierStokesOverviewControlFlowField}
\end{figure}

In \cref{fig:NavierStokesOverviewControlFlowField} the resulting velocity fields at the end of the 
observation horizon are visualized \cite[cf. figure 4]{bhan2024pde}. It is evident that \ourAlgo{}
nearly perfectly matches the desired flow and clearly outperforms
the model-free baseline which has difficulties finding the correct magnitude of the velocity field,
although the general directions are correct.\footnote{Since we use with 
\rlLib{}~\cite{RayRlLib2017} instead
of \emph{Stable-Baselines3} \cite{stable-baselines3} a different RL-engine and we run PPO with
different settings, we are not slightly able to achieve the performance presented in
\cite[table 3]{bhan2024pde} with our baseline model. But, our \ourAlgo{} outperforms the
presented baseline algorithms.} Thus, imposing a Dyna-style dynamics model does not only
significantly decrease reducing the number of full-order model interactions, however it helps
the agent to grasp the general dynamics faster.
While the controls of the agent \ourAlgo{} stay close to the reference value, the
baseline algorithm has difficulties in smoothly increasing the controls (previous epochs exhibit
strongly oscillating behavior, similar to bang-bang controls).

\begin{minipage}{\textwidth}
	\begin{minipage}[b]{0.49\textwidth}
	  \centering
	  \includegraphics[scale=.5]{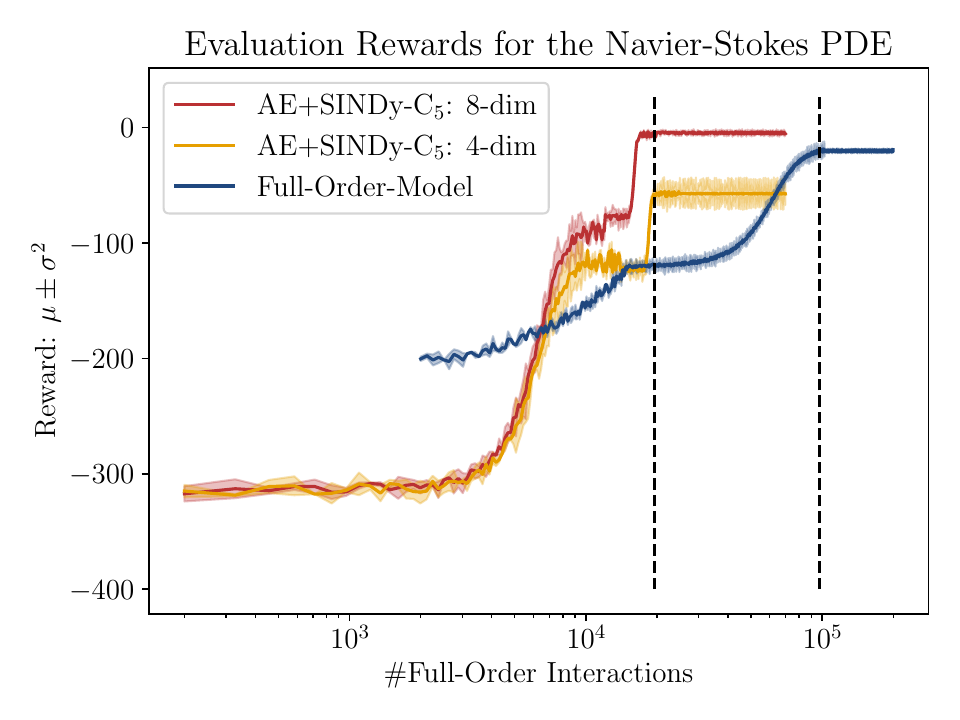}
	  \captionof{figure}{Sample efficiency of the Dyna-style \ourAlgo{} method for the \textbf{Navier-Stokes} equations.
	  We test $\kDyn = 5$ against the full-order model-free baseline. The \textbf{dashed vertical lines}
	  indicate the point of early stopping for
	  each of the model after \textbf{750 epochs} and represent the models which are evaluated 
	  in detail in \cref{sec:5AnalysisHigherPenalties}.
	  For the evaluation the performance over \textbf{five fixed random seeds is used}.\\\;}
	\label{fig:NavierStokesFOMInteractionsPerformance}
	\end{minipage}
	\hfill
	\begin{minipage}[b]{0.49\textwidth}
	  \centering
	 \scalebox{0.65}{
		\begin{tabular}{llll}
    \toprule
                                              & Baseline $(882 \times 1)$ & \multicolumn{2}{c}{$\ourAlgo{}, \kDyn = 5, (882 \times 1)$}\\
                                              \cmidrule(rl){3-4}
    Surrogate Space Dim                       &                           & 8 & 4 \\
    \midrule
    \# FOM Interactions                       & 97280                     & \textbf{19456}                          & \textbf{19456}\\
    $\quad$Off-policy                         & -                         & 2000                                    & 2000  \\
    $\quad$On-policy                          & 97280                     & 17456                                   & 17456 \\
    Reward $\mathcal{R}$ $(\mu \pm \sigma^2)$ & $-24.88 \pm 5.77$         & $\bf{-3.61 \pm 1.64}$                   & $-60.09 \pm 12.11$\\
    Total \# parameters                       & \textbf{226306}           & 377601                                  & 319421 \\
    $\quad\ourAlgo{}$                         & -                         & 151295                                  & 93115 \\
    $\quad$Actor + Critic                     & 226306                    & 226306                                  & 226306 \\
    \bottomrule
\end{tabular}

		}
        \vspace{.85cm}
		\captionof{table}{
	Performance comparison of the Dyna-style \ourAlgo{} method for the \textbf{Navier-Stokes} equations.
	We test $\kDyn = 5$ against the full-order baseline (both $\dimObsX \times \dimU = 882 \times 1$).
	The models correspond to the dashed vertical lines in
	\cref{fig:NavierStokesFOMInteractionsPerformance} and represent all models after \textbf{750 epochs}.
	 We compare the number of full-order model (FOM) interactions, the reward
	using \textbf{five fixed random seeds} and the total number of parameters. 
	\textbf{Best performances} (bold) are highlighted \textbf{row-wise}.}
	\label{tab:ComparisonNavierStokesNumbers}
	\end{minipage}
\end{minipage}

\subsubsection{Dynamics in the Latent Space} % Surrogate 
\begin{figure}[H]
	\centering
	\begin{subfigure}[t]{0.4\textwidth}
		\centering
		
		\includegraphics[width=\linewidth,keepaspectratio]{
			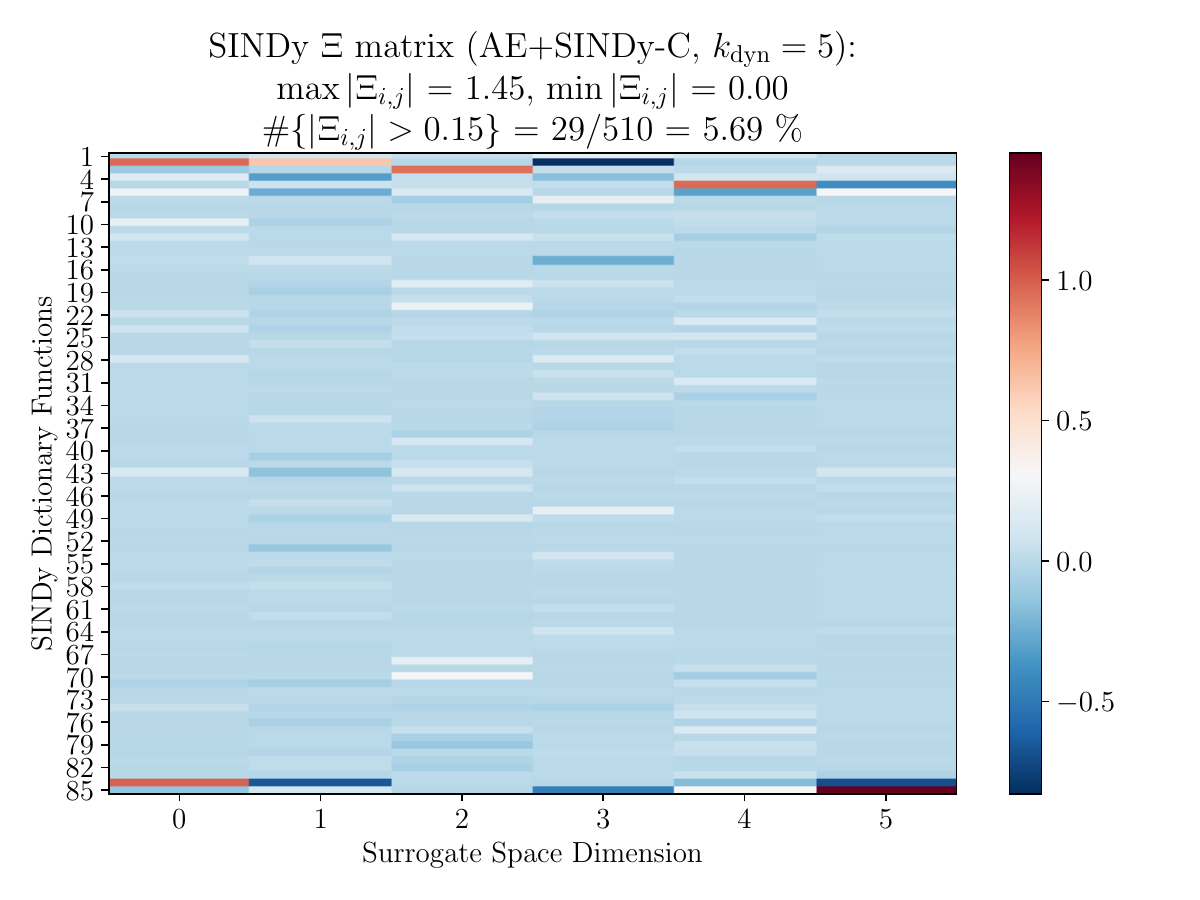
		}
	\caption{8-dimensional Latent space.}
	\label{fig:NavierStokesXiHeatmapLarge}
	\end{subfigure}
	\hfill
	\begin{subfigure}[t]{0.4\textwidth}
		\centering
		\includegraphics[width=\linewidth,keepaspectratio]{
			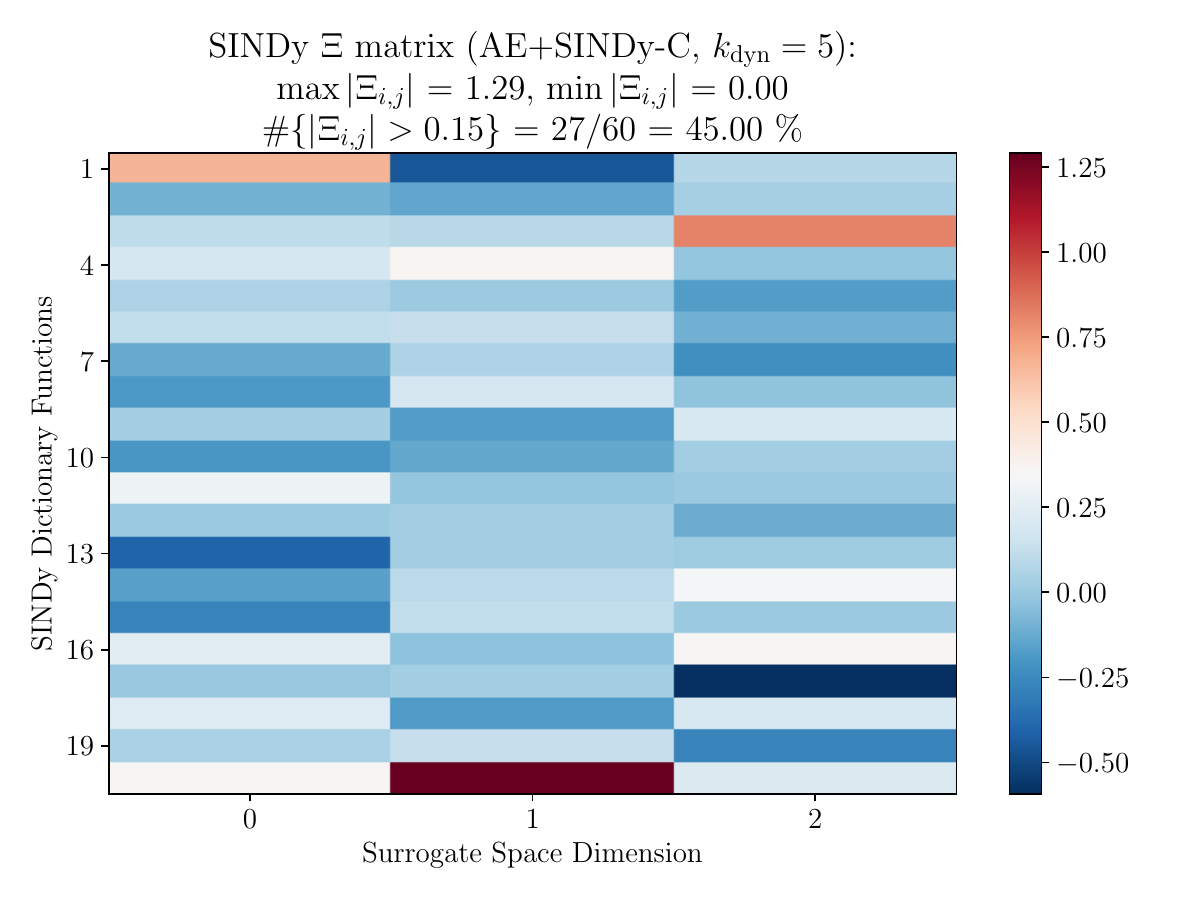
		}
	\caption{4-dimensional Latent space.}
	\label{fig:NavierStokesXiHeatmapSmall}
	\end{subfigure}
	\caption{Analysis of the coefficient matrix $\vec{\Xi} \in \RR^{d \times \dimObsX}$ for the
	\textbf{Navier-Stokes} equations.} 
	\label{fig:NavierStokesXiHeatmap}
\end{figure}

As in the Burgers' equation example, we visualize the coefficient matrix $\vec{\Xi} \in \RR^{d \times \dimRedX}$ for the eight-dimensional latent space version.
\Cref{fig:NavierStokesXiHeatmap} shows the sparse nature of the dynamics model.
We also compute the closed-form representation of the \SindyC{}
dynamics model in the surrogate space and cut off non-significant values with an absolute value
below 0.15.%. \textcolor{red}{why? a better motivation should be inserted.} 
Interestingly, with 29, and respectively 27, 
coefficients with absolute values above the threshold, both representations need
almost the same amount of basis functions, indicating the amount
of information carried by the state and action of the PDE.
We obtain the following dynamics equations:

\begin{itemize}
	\item 8-dimensional Latent space:
		\begin{align*}
		\vec{z}_{x,1}(t+1) = &+ 0.960 \cdot \vec{z}_{x,2}(t) + 0.165 \cdot \vec{z}_{x,4}(t) + 0.251 \cdot \vec{z}_{x,6}(t) \\
			&+ 0.228 \cdot \vec{z}_{x,1}(t) \vec{z}_{x,4}(t) + 0.981 \cdot \vec{z}_{u,1}(t)\\
		\vec{z}_{x,2}(t+1) = &+ 0.631 \cdot \vec{z}_{x,2}(t) - 0.308 \cdot \vec{z}_{x,4}(t) 
			- 0.245 \cdot \vec{z}_{x,6}(t) - 0.652 \cdot \vec{z}_{u,1}(t)\\
		\vec{z}_{x,3}(t+1) = &+ 0.941 \cdot \vec{z}_{x,3}(t) + 0.166 \cdot \vec{z}_{x,3}(t)^2 
			+ 0.265 \cdot \vec{z}_{x,3}(t) \vec{z}_{x,6}(t) + 0.190 \cdot \vec{z}_{x,3}(t) \vec{z}_{x,4}(t)^2\\
			&+ 0.286 \cdot \vec{z}_{x,3}(t) \vec{z}_{x,4}(t) \vec{z}_{x,6}(t)\\
		\vec{z}_{x,4}(t+1) = &+ 0.208 \cdot \vec{z}_{x,1}(t) - 0.828 \cdot \vec{z}_{x,2}(t) - 0.158 \cdot \vec{z}_{x,4}(t) + 0.216 \cdot \vec{z}_{x,1}(t)^2 \\
			&- 0.235 \cdot \vec{z}_{x,2}(t) \vec{z}_{x,4}(t) + 0.175 \cdot \vec{z}_{x,1}(t) \vec{z}_{x,6}(t)^2 - 0.464 \cdot \vec{z}_{u,2}(t)\\
		\vec{z}_{x,5}(t+1) = &+ 0.949 \cdot \vec{z}_{x,5}(t) - 0.290 \cdot \vec{z}_{x,6}(t) - 0.177 \cdot \vec{z}_{u,1}(t) + 0.310 \cdot \vec{z}_{u,2}(t)\\
		\vec{z}_{x,6}(t+1) = &- 0.396 \cdot \vec{z}_{x,5}(t) + 0.314 \cdot \vec{z}_{x,6}(t) - 0.695 \cdot \vec{z}_{u,1}(t) + 1.451 \cdot \vec{z}_{u,2}(t)
		\end{align*}
	\item 4-dimensional Latent space:
	\begin{align*}
		\vec{z}_{x,1}(t+1) = &+ 0.668 \cdot \vec{z}_{x,1}(t) + 0.183 \cdot \vec{z}_{x,1}(t)^2 - 0.189 \cdot \vec{z}_{x,2}(t) \vec{z}_{x,3}(t) - 0.195 \cdot \vec{z}_{x,1}(t)^3\\
			&+ 0.299 \cdot \vec{z}_{x,1}(t)^2 \vec{z}_{x,2}(t) - 0.405 \cdot \vec{z}_{x,1}(t) \vec{z}_{x,2}(t)^2 - 0.160 \cdot \vec{z}_{x,1}(t) \vec{z}_{x,2}(t) \vec{z}_{x,3}(t)\\
			& - 0.282 \cdot \vec{z}_{x,1}(t) \vec{z}_{x,3}(t)^2 + 0.246 \cdot \vec{z}_{x,2}(t)^3 + 0.230 \cdot \vec{z}_{x,2}(t) \vec{z}_{x,3}(t)^2 + 0.361 \cdot \vec{z}_u(t)\\
		\vec{z}_{x,2}(t+1) = &- 0.448 \cdot \vec{z}_{x,1}(t) + 0.370 \cdot \vec{z}_{x,1}(t)^2 + 0.192 \cdot \vec{z}_{x,2}(t) \vec{z}_{x,3}(t) - 0.174 \cdot \vec{z}_{x,3}(t)^2 \\
			&- 0.183 \cdot \vec{z}_{x,2}(t) \vec{z}_{x,3}(t)^2 + 1.291 \cdot \vec{z}_u(t)\\
		\vec{z}_{x,3}(t+1) = &+ 0.817 \cdot \vec{z}_{x,3}(t) - 0.175 \cdot \vec{z}_{x,1}(t) \vec{z}_{x,2}(t) - 0.226 \cdot \vec{z}_{x,2}(t)^2 + 0.203 \cdot \vec{z}_{x,3}(t)^2\\
			&+ 0.333 \cdot \vec{z}_{x,1}(t) \vec{z}_{x,2}(t) \vec{z}_{x,3}(t) + 0.359 \cdot \vec{z}_{x,2}(t)^3 - 0.593 \cdot \vec{z}_{x,2}(t)^2 \vec{z}_{x,3}(t)\\
			&+ 0.195 \cdot \vec{z}_{x,2}(t) \vec{z}_{x,3}(t)^2 - 0.281 \cdot \vec{z}_{x,3}(t)^3 + 0.215 \cdot \vec{z}_u(t)
	\end{align*}
\end{itemize}

\subsubsection{Training of \ourAlgo{}}
For the sake of completeness, the goodness of fit in terms of the average training and validation error
are provided in \cref{tab:NavierStokesTrainingTimeError} for the Navier-Stokes case. Compared to the Burgers' equation, the training and validation
loss are of the same order of magnitude. Since the \ourAlgo{} model is much larger in this case, the
training takes longer. Interestingly, while in the Burgers' equation case our internal logging metrics show
that usually even 20-30 epochs would be enough, in the more complex Navier-Stokes equations case on average
at least 70-80 epochs are needed before the loss converges.

\begin{table}[H]
    \centering
\begin{tabular}{lll}
\toprule
                                                                             & \multicolumn{2}{c}{\ourAlgo{}, $\kDyn = 5, (882\times1)$}              \\
                                                                             \cmidrule(rl){2-3}
Surrogate Dimension                                                          & 8                                          & 4                                          \\
\midrule
Training Time $[s]$ $(\mu \pm \sigma^2)$                                     & $12.65 \pm 5.23$                           & $9.81 \pm 2.92$                            \\
Loss \cref{eq:AutoencoderLossFunction} $(\mu \pm \sigma^2)$ &                                            &                                            \\
$\quad$ Training                                                             & $1.53 \cdot 10^{-4} \pm 3.01\cdot 10^{-4}$ & $7.57 \cdot 10^{-3} \pm 3.36\cdot 10^{-4}$ \\
$\quad$ Validation                                                           & $9.33 \cdot 10^{-4} \pm 3.82\cdot 10^{-4}$ & $9.11 \cdot 10^{-3} \pm 8.43\cdot 10^{-4}$\\
\bottomrule
\end{tabular}
    \caption{Analysis of the internal \textbf{loss distribution} for the training and
	validation data during the AE training
	as well as the \textbf{training time} for the \textbf{Navier-Stokes} equations,
	trained on a MacBook M1 (2021, 16GB RAM).}
    \label{tab:NavierStokesTrainingTimeError}
\end{table}
\vspace{-0.8cm}
\section{Discussion and Conclusion}
\label{sec:DiscussionAndFutureWork}
We propose a data-efficient and interpretable Dyna-style Model-Based Deep Reinforcement Learning 
(MBRL) framework for controlling distributed systems, such as those governed by 
Partial Differential Equations. Combining \SindyC{} with an
autoencoder framework not only scales to high-dimensional systems but also provides a low-dimensional
learned representation as a dynamical system in the latent space. The proposed controllers have proven
to be effective and robust in both partially and fully observable cases. Additionally, we showed 
how the proposed framework can be used to estimate an approximate lower bound for a low-dimensional
surrogate representation of the dynamics.

A clear limitation of our method is the training of an autoencoder online. Problems
such as overfitting \cite{Goodfellow-et-al-2016, Zhang2021Understanding} and catastrophic
forgetting \cite{Kirkpatrick2017Overcoming, McCloskey1989Catastrophic} lead to decreasing performance
if trained for too long. Potential options to overcome this issue could include the usage of shorter roll-outs
\cite{janner_when_2021}, ensemble methods \cite{zolman_sindy-rl_2024},
or training the autoencoder using sequential thresholding to enforce zero-valued
coefficients. The Navier-Stokes PDE example showed that, under the availability
of enough regularity in the initial condition this problem can be lessened. On the other hand, the second
example highlights well the importance of choosing the proper latent space representation and experiment
with the dimensionality of the surrogate space in order to avoid a poor quality of fitting.

Further research should address these limitations to enhance the robustness and 
generalizability of our framework across diverse PDE scenarios. Beyond the current 
improvements, integrating \SindyC{} with autoencoders opens new avenues in multiple
fields. In future work, we plan to extend the latent space representation to include 
parameter dependencies, as seen in \cite{conti_reduced_2023}, enabling more 
effective learning of parametric systems, such as parameterized PDEs.

\emph{The code for \ourAlgo{} will be made publicly available upon acceptance.}

\section*{Acknowledgments}
FW was supported by a scholarship from the Italian 
Ministry of Foreign Affairs and International Cooperation.
AM acknowledges the Project “Reduced Order Modeling and Deep Learning for the 
real- time approximation of PDEs (DREAM)” (Starting Grant No. FIS00003154), 
funded by the Italian Science Fund (FIS) - Ministero dell'Università e della 
Ricerca and the project FAIR (Future Artificial Intelligence Research), 
funded by the NextGenerationEU program within the PNRR-PE-AI scheme (M4C2, 
Investment 1.3, Line on Artificial Intelligence). AM is member of the Gruppo 
Nazionale Calcolo Scientifico-Istituto Nazionale di Alta Matematica (GNCS-
INdAM). 
% \textcolor{red}{Anyone else has to acknowledge projects?}

\section*{Declarations}
\textbf{Conflict of interest} We declare that we have no financial and personal 
relationships with other people or
organizations that can inappropriately influence our work. There is no professional 
or other personal interest
of any nature or kind in any product, service and/or company that 
could be construed as influencing the position
presented in, or the review of, the manuscript entitled.

% Print bibliography
\printbibliography
%\input{./AE_SINDyC_Wolf_Manzoni.bbl}
%-------------------------------------------
% Appendix
%-------------------------------------------
\appendix
%\newpage
% Change equation numbering format to be sequential within sections in the appendix
\renewcommand\theequation{\Alph{section}\arabic{equation}} % Redefine equation numbering format
\counterwithin*{equation}{section} % Number equations within sections
\renewcommand\thefigure{\Alph{section}\arabic{figure}} % Redefine equation numbering format
\counterwithin*{figure}{section} % Number equations within sections
\renewcommand\thetable{\Alph{section}\arabic{table}} % Redefine equation numbering format
\counterwithin*{table}{section} % Number equations within sections

\begin{appendices}
\section{Burgers' Equation: Additional Evaluation}
\label{sec:Appendix}
\subsection{Random Initial Condition: State and Control Trajectories}
\label{sec:RandomInitialConditionStateControlBurgers}
\begin{figure}[H]
	\centering
	\begin{subfigure}[t]{0.49\textwidth}
		\centering
		\includegraphics[width=\linewidth, keepaspectratio]{
			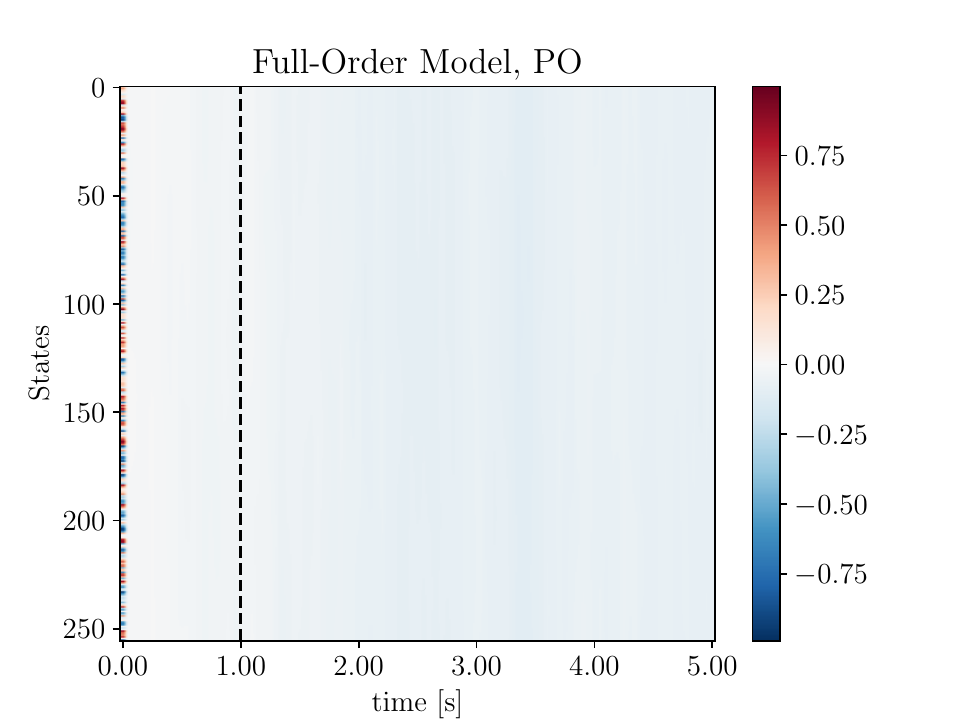
		}
		\caption{FOM: states, partially observable}
		\label{fig:BurgersFullOrderModelpartiallyObservableState}
	\end{subfigure}
	\hfill
	\begin{subfigure}[t]{0.49\textwidth}
		\centering
		\includegraphics[width=\linewidth, keepaspectratio]{
			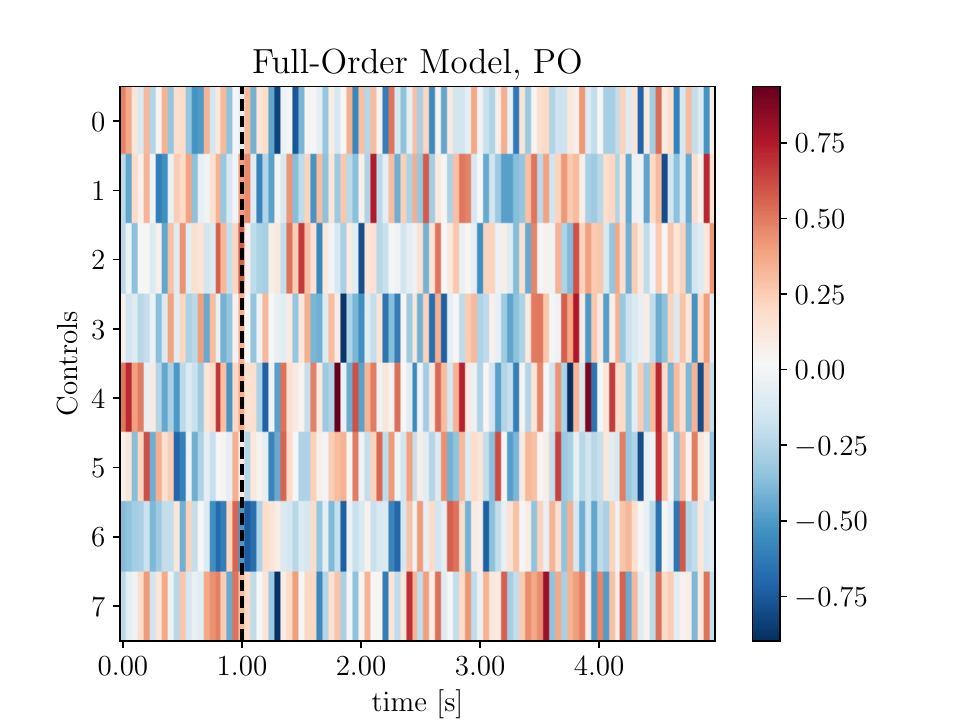
		}
		\caption{FOM: controls, partially observable}
		\label{fig:BurgersFullOrderModelpartiallyObservableControl}
	\end{subfigure}
    \hfill
	\begin{subfigure}{0.49\textwidth}
		\centering
		\includegraphics[width=\linewidth,keepaspectratio]{
			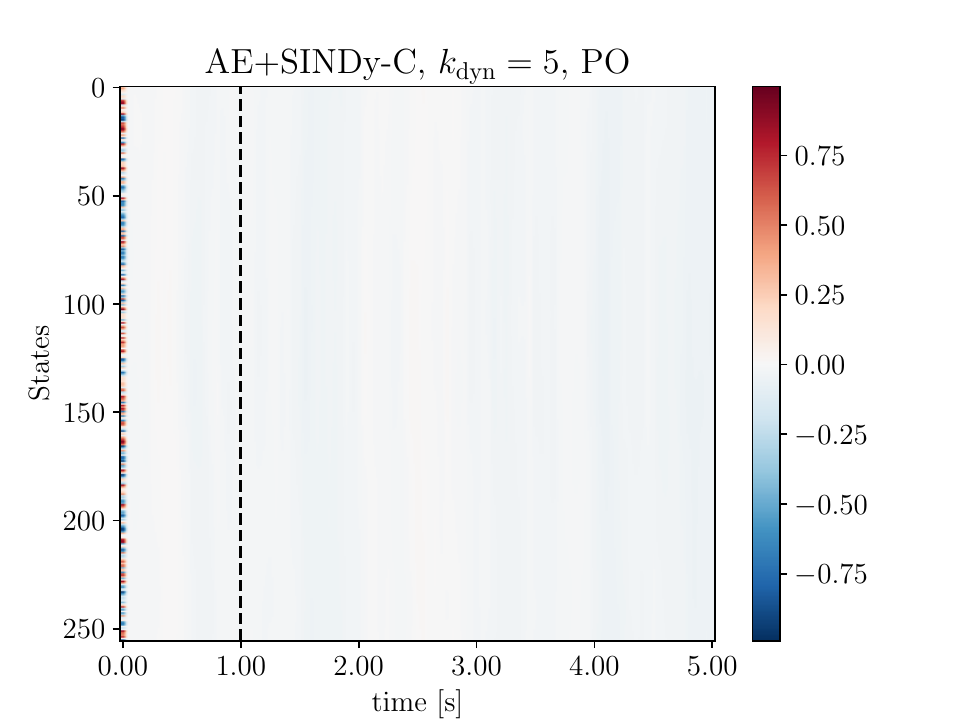
		}
		\caption{\ourAlgo{} with $\kDyn = 5$:: states, partially observable}
		\label{fig:BurgersAutoencoderpartiallyObservableStateFit5}
	\end{subfigure}
	\hfill
	\begin{subfigure}{.49\textwidth}
		\centering
		\includegraphics[width=\linewidth,keepaspectratio]{
			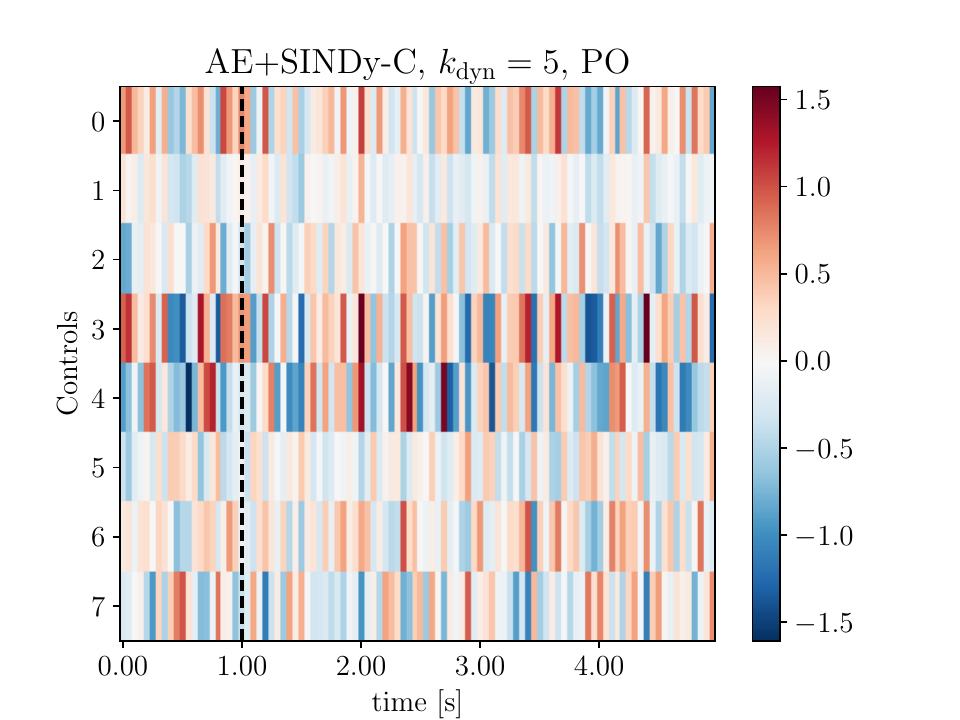
		}
		\caption{\ourAlgo{} with $\kDyn = 5$: controls, partially observable}
		\label{fig:BurgersAutoencoderPartiallyObservableControlFit5}
	\end{subfigure}
	\hfill
	\begin{subfigure}{0.49\textwidth}
		\centering
		\includegraphics[width=\linewidth,keepaspectratio]{
			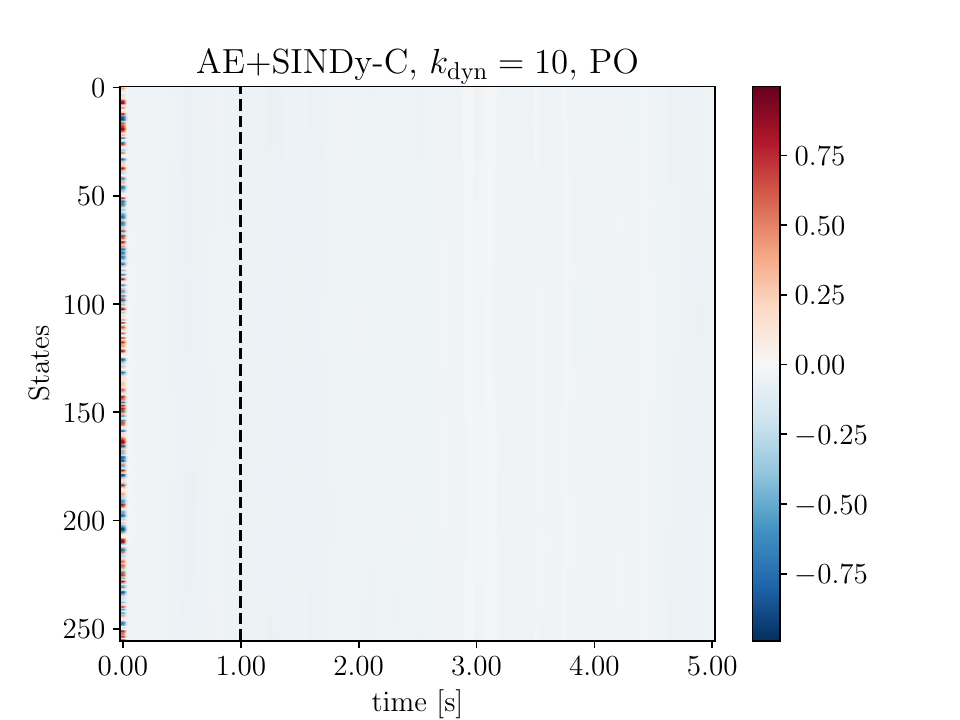
		}
		\caption{\ourAlgo{} with $\kDyn = 10$: states, partially observable}
		\label{fig:BurgersAutoencoderpartiallyObservableStateFit10}
	\end{subfigure}
	\hfill
	\begin{subfigure}{.49\textwidth}
		\centering
		\includegraphics[width=\linewidth,keepaspectratio]{
			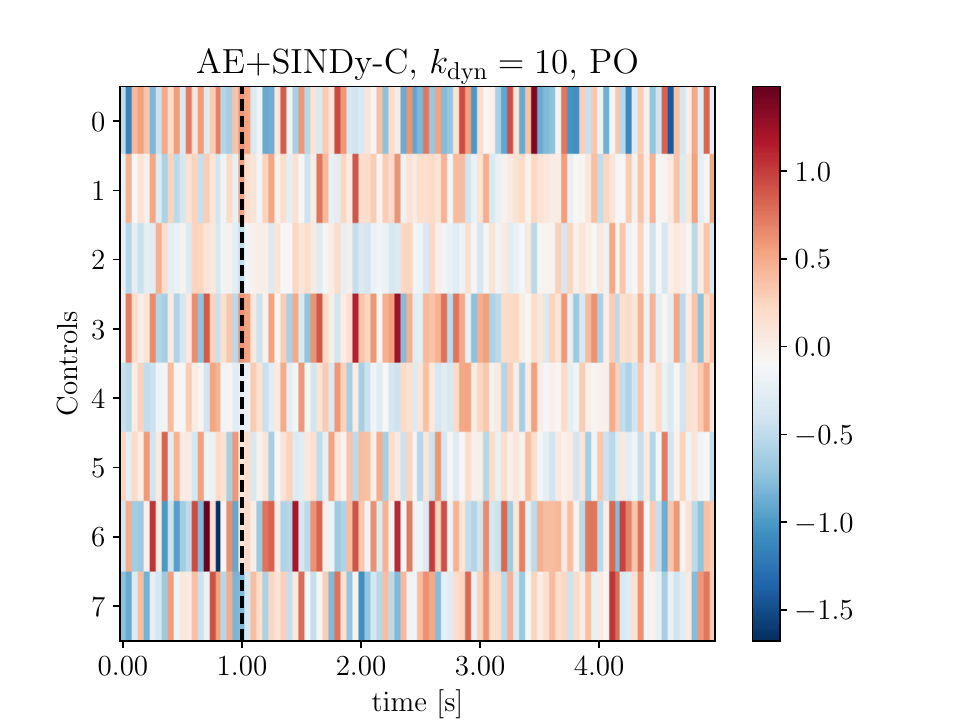
		}
		\caption{\ourAlgo{} with $\kDyn = 5$: controls, partially observable}
		\label{fig:BurgersAutoencoderPartiallyObservableControlFit10}
	\end{subfigure}
	\caption{State and control trajectories for the \textbf{Burgers'} equation in the
	\textbf{partially observable} (PO) case. The initial condition is a \textbf{uniform} distribution
	$\Uniform$ and the black \textbf{dashed line} indicates the timestep $t$ of
	\textbf{extrapolation in time}.}
	\label{fig:ComparionBurgersPartiallyObservableStateControl}
\end{figure}

\begin{figure}[H]
	\centering
	\begin{subfigure}[t]{0.49\textwidth}
		\centering
		\includegraphics[width=\linewidth, keepaspectratio]{
			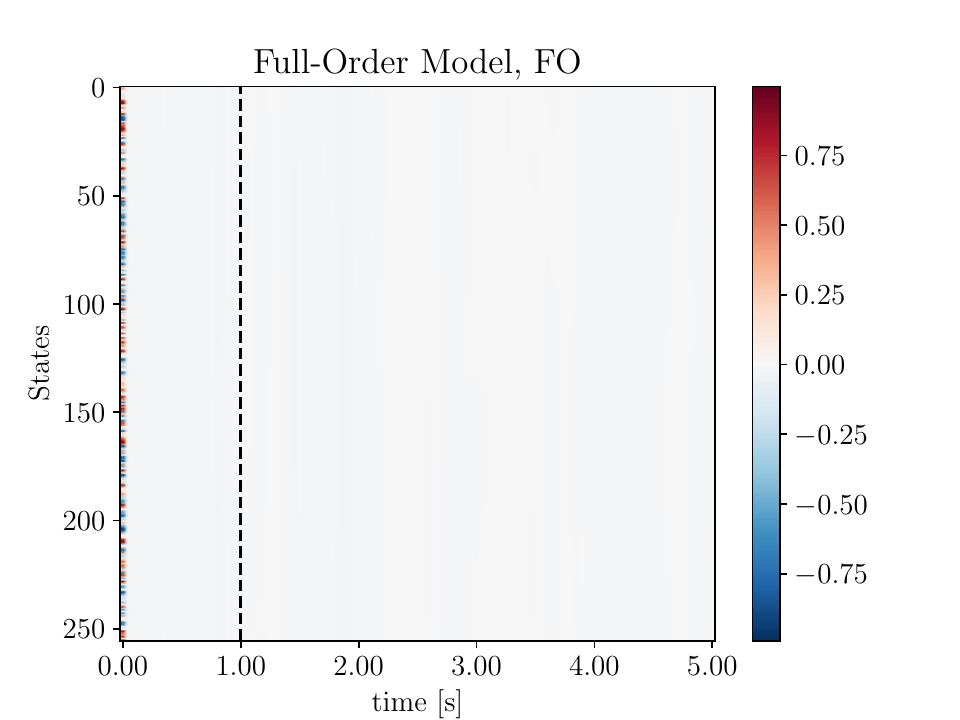
		}
		\caption{FOM: states, fully observable}
		\label{fig:BurgersFullOrderModelFullyObservableState}
	\end{subfigure}
	\hfill
	\begin{subfigure}[t]{0.49\textwidth}
		\centering
		\includegraphics[width=\linewidth, keepaspectratio]{
			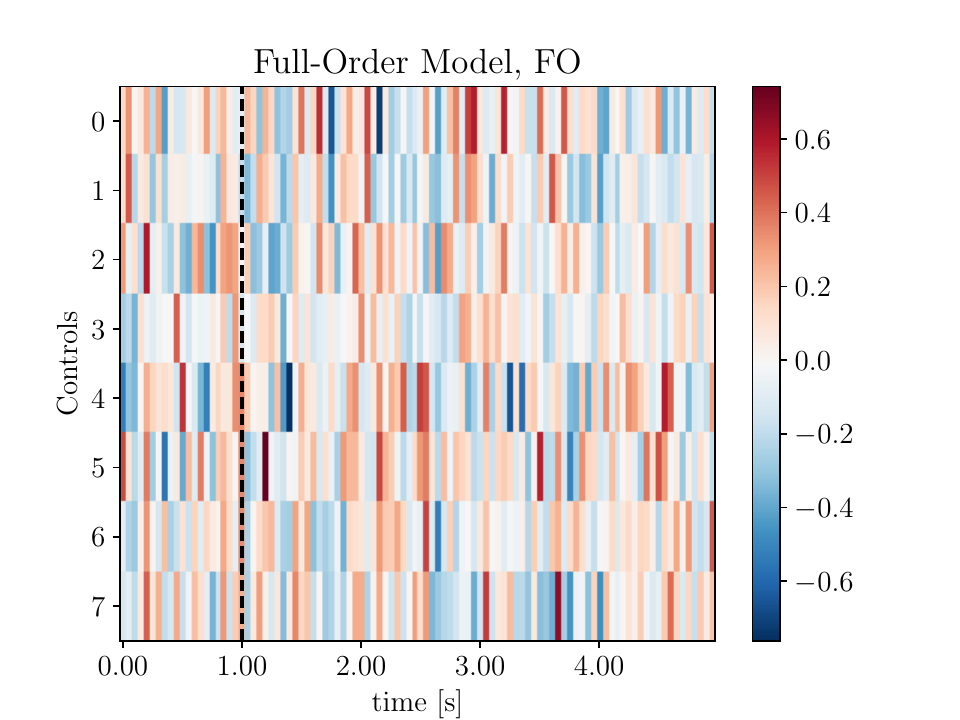
		}
		\caption{FOM: controls, fully observable}
		\label{fig:BurgersFullOrderModelFullyObservableControl}
	\end{subfigure}
    \hfill
	\begin{subfigure}{0.49\textwidth}
		\centering
		\includegraphics[width=\linewidth,keepaspectratio]{
			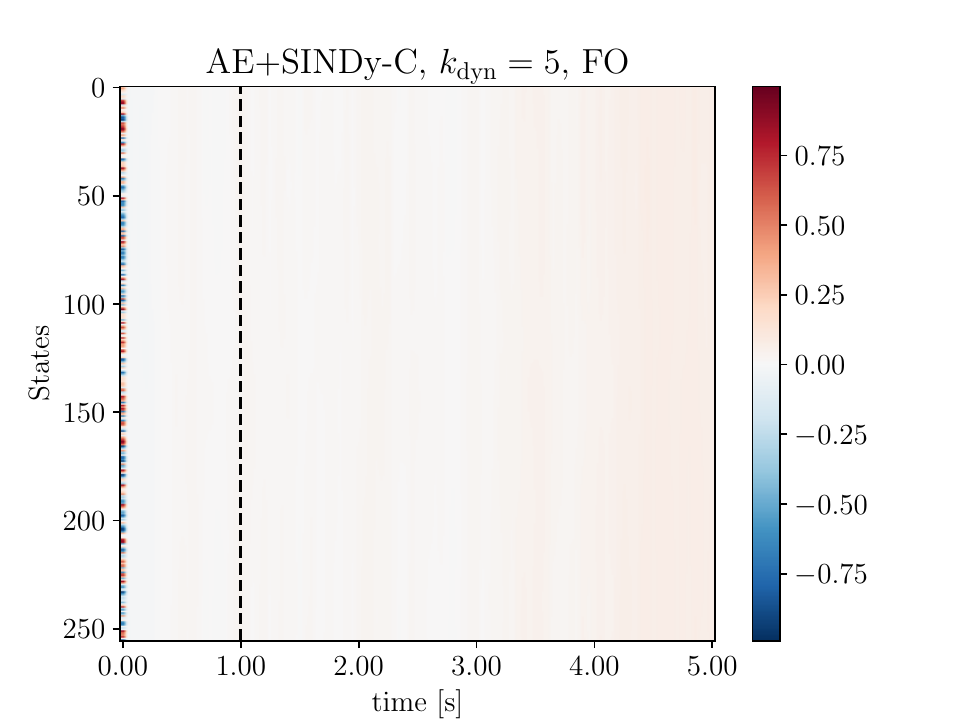
		}
		\caption{\ourAlgo{} with $\kDyn = 5$: states, fully observable}
		\label{fig:BurgersAutoencoderFullyObservableStateFit5}
	\end{subfigure}
	\hfill
	\begin{subfigure}{.49\textwidth}
		\centering
		\includegraphics[width=\linewidth,keepaspectratio]{
			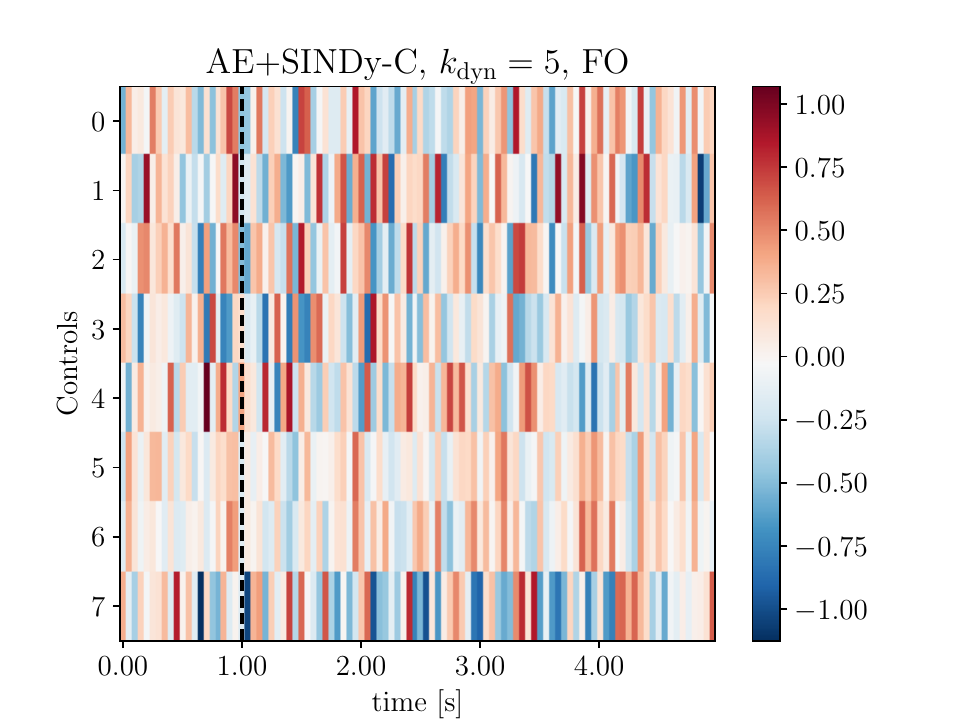
		}
		\caption{\ourAlgo{} with $\kDyn = 5$: controls, fully observable}
		\label{fig:BurgersAutoencoderFullyObservableControlFit5}
	\end{subfigure}
	\hfill
	\begin{subfigure}{0.49\textwidth}
		\centering
		\includegraphics[width=\linewidth,keepaspectratio]{
			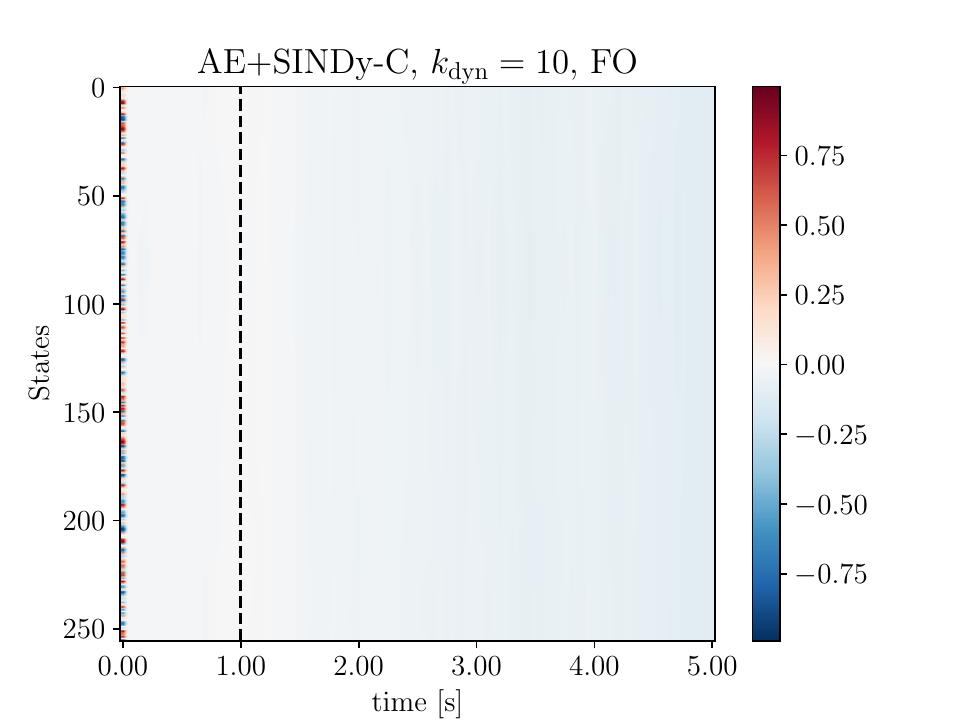
		}
		\caption{\ourAlgo{} with $\kDyn = 10$: states, fully observable}
		\label{fig:BurgersAutoencoderFullyObservableStateFit10}
	\end{subfigure}%
	\hfill
	\begin{subfigure}{.49\textwidth}
		\centering
		\includegraphics[width=\linewidth,keepaspectratio]{
			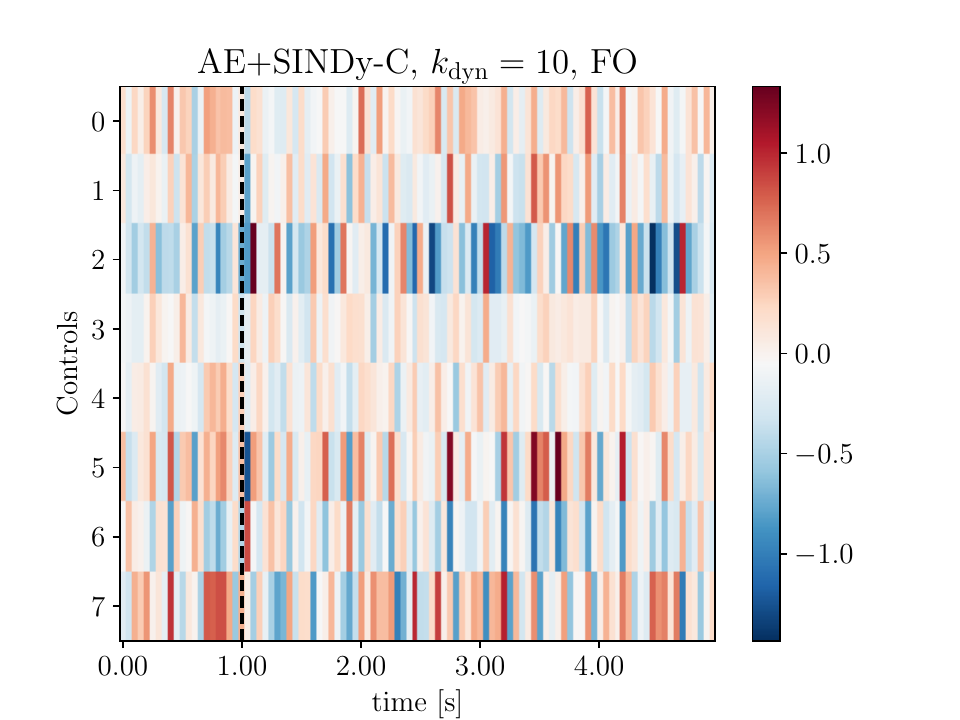
		}
		\caption{\ourAlgo{} with $\kDyn = 10$: controls, fully observable}
		\label{fig:BurgersAutoencoderFullyObservableControlFit10}
	\end{subfigure}%
	\caption{State and control trajectories for the \textbf{Burgers'} equation in the
	\textbf{fully observable} (FO) case. The initial condition is a \textbf{uniform} distribution
	$\Uniform$ and the black \textbf{dashed line} indicates the timestep $t$ of
	\textbf{extrapolation in time}.}
	\label{fig:ComparisonBurgersFullyObservableStateControl}
\end{figure}

We analyze the models indicated by the dashed line in \cref{fig:BurgerFOMInteractionsPerformance}
with the same fixed random seed for one specific initial condition. As in the training, the initial
condition is drawn from a uniform $\Uniform$ distribution. The idea of training the agent in this way
might seem counterintuitive at first due to the non-regularity of the initial state,
but turns out to be a very effective strategy, as we will see
in \cref{sec:BellShapeBurgersGeneralization}.
We impose low penalties on the controls (cf. \cref{sec:AppendixHyperparameters}), resulting in very
aggressive control strategies. Fig. 
\ref{fig:ComparionBurgersPartiallyObservableStateControl} and
\cref{fig:ComparisonBurgersFullyObservableStateControl} visualize the results for the
partially respectively fully observable case. Both of them confirm the results we have already
seen from \cref{fig:BurgerFOMInteractionsPerformance}.

Overall, the controls show a very chaotic behavior and are not regular.
This can clearly be explained by the non-regularity of the initial state distribution and the resulting state trajectories, 
as well as the low penalty on the control itself. 
The issue of regularity has been discussed in detail in \cref{sec:BellShapeBurgersGeneralization},
when we consider regular, but out-of-distribution, initial conditions. With a regular initial
condition, the problem of chaotic controls does not appear.

In a model-by-model comparison, the FO case is overall solved more effectively and with less
variation between different random seeds -- confirming the intuition that, in general, more
measurement points increase the performance of the agent. Both cases will be individually
analyzed in detail in the following two paragraphs.

\paragraph{Partially Observable (PO)} Compared to the FO case, the PO case
(cf. \cref{fig:ComparionBurgersPartiallyObservableStateControl}) is as expected more challenging
for all of the methods. All three controllers struggle 
to correctly capture the system dynamics and effectively regulate the system -- represented
by lower rewards in general and higher standard deviations
(cf. \cref{tab:ComparisonBurgerNumbers}). In the PO case, the baseline model is outperformed
by the \ourAlgo{} method, although taking into account the standard deviation, the difference
is not significant. Interestingly, while the baseline model seems to rely on all of the controls,
both, the $\kDyn = 5$ and $\kDyn = 10$ exhibit one $u_i$ with close to zero controls over the
entire time horizon (cf. \cref{fig:BurgersAutoencoderPartiallyObservableControlFit5} index one
and \cref{fig:BurgersAutoencoderFullyObservableControlFit10} index two). Overall, the PDE is
aggressively controlled and successfully regulated towards the zero-state.

\paragraph{Fully Observable (FO)} In the FO case
(cf. \cref{fig:ComparisonBurgersFullyObservableStateControl}) the baseline model outperforms the \ourAlgo{} method.
Nevertheless, \cref{tab:ComparisonBurgerNumbers} highlights that the differences between the models
are marginal although the baseline model specifically stands out by a much lower standard deviation
and thus can be trusted more. The performance of \ourAlgo{} for $\kDyn = 5$ and $\kDyn = 10$ are similar, also
regarding their standard deviations, even though at the end of the extrapolation horizon in the case of
$\kDyn = 5$ the DRL agent seems to slightly overshoot the target while in the case of $\kDyn = 10$ 
the agent undershoots the target (see \cref{fig:BurgersAutoencoderFullyObservableStateFit5} 
and \cref{fig:BurgersAutoencoderFullyObservableStateFit10} respectively).

%--- Section ---%
\section{Environments: Hyperparameters}
\label{sec:AppendixHyperparameters}
\begin{table}[H]
	\centering
	\begin{tabular}{ccc}
	\toprule
								  & Partially observable & Fully observable \\
	\midrule
	Domain Size                   & 1.0                  & 1.0              \\
	Spatial Discretization        & 256                  & 256              \\
	Observable States $N_s$       & 48                   & 256              \\
	Time: Train + Extrapolation   & 1s + 4s              & 1s + 4s          \\
	Control Discretization $N_u$  & 8                    & 8                \\
	Control support width         & 0.125                & 0.125            \\
	Additive Gaussian Noise (FOM) $\sigma$ &  0.25      &   0.25 \\
	Penalty State $Q_w$           & 100.0                & 100.0            \\
	Penalty Control $R_w$         & 0.01                 & 0.01             \\
	\bottomrule
	\end{tabular}
	\caption{Environment details for \emph{ControlGym}'s implemention of the
	\textbf{Burgers'} equation, the diffusivity constant  $\nu = 1.0$ is fixed.}
\end{table}

\begin{table}[H]
	\centering
	\begin{tabular}{cc}
	\toprule
								  & Fully observable \\
	\midrule
	Spatial Domain 						  & $[0,1]^2$          \\
	Spatial discretization step $\Delta_x$       & 0.05                  \\
	Spatial discretization step $\Delta_y$       & 0.05                  \\
	Time discretization step $\Delta_t$       & 0.001                  \\
	Time: Train + Extrapolation   & 0.2s              \\
	Penalty $\gamma$ controls     & 0.1                \\
	\bottomrule
	\end{tabular}
	\caption{Environment details for \emph{PDEControlGym}'s implemention of the
	\textbf{Navier-Stokes} equations.}
\end{table}

\section{Deep Reinforcement Learning: Hyperparameters}
\label{sec:AppendixDRL}
\begin{table}[H]
	\centering
	\begin{tabular}{ccc}
	\toprule
							 & Partially observable & Fully observable    \\
	\midrule
	Network Class            & PPO 		            & PPO                 \\
	Batch Size               & 256                  & 256                 \\
	Hidden Layer Size        & 128                  & 128                 \\
	Learning Rate            & $3.0 \cdot 10^{-4}$  & $3.0 \cdot 10^{-4}$ \\
	GAE $\lambda$            & 0.95                 & 0.95                \\
	Discount Factor $\gamma$ & 0.99                 & 0.99                \\
	Gradient Clipping        & 0.5                  & 0.5                 \\
	\bottomrule
	\end{tabular}
	\caption{DRL algorithm configuration details for the \textbf{Burgers'} equation experiment.
	We use \rlLib{} \cite{RayRlLib2017} to train our models. The \emph{PPO} \cite{PPO2017}
	policy is trained by using the \emph{Adam} algorithm \cite{AdamKingBa15}.}
\end{table}

\begin{table}[H]
	\centering
	\begin{tabular}{ccc}
	\toprule
							 & Partially observable \\
	\midrule
	Network Class            & PPO 		            \\
	Batch Size               & 128                  \\
	Hidden Layer Size        & 128                  \\
	Learning Rate            & $3.0 \cdot 10^{-4}$  \\
	GAE $\lambda$            & 0.95                 \\
	Discount Factor $\gamma$ & 0.99                 \\
	Gradient Clipping        & 0.5                  \\
	\bottomrule
	\end{tabular}
	\caption{DRL algorithm configuration details for the \textbf{Navier-Stokes} equations experiment.
	We use \rlLib{} \cite{RayRlLib2017} to train our models. The \emph{PPO} \cite{PPO2017}
	policy is trained by using the \emph{Adam} algorithm \cite{AdamKingBa15}.}
\end{table}

\section{Autoencoder: Hyperparameters and Training Details}
\label{sec:AppendixAutoencoder}
Internally, a $80/20$ splitting is used for training and validation and once new data is
available, the surrogate model is trained for 100 epochs. In all of the cases, we computed
the number of neurons of the hidden layer such that the ratio between the size of the
input and the hidden layer is the same as the ratio between the size of the hidden and the output
layer. Only for the controls of the Navier-Stokes equations we went for an increase in Neurons
to find an effective latent space representation (cf. \cref{tab:DetailsAETrainingNavierStokes}).

\iffalse
\ourAlgo{} is trained by optimizing the loss function
\begin{align*}
    \begin{aligned}
    \min_{\vec{W}_{\phi_x}, \vec{W}_{\phi_u}, \vec{W}_{\psi_x}, \vec{W}_{\psi_u}, \vec{\Xi}}
    \quad &\underbrace{\norm{\vec{x}_{t+1}  - \psi_x\left(\vec{\Theta}( \phi_x(\vec{x}_t; \vec{W}_{\phi_x}),
            \phi_u(\vec{u}_t; \vec{W}_{\phi_u})) \cdot {\vec{\Xi}};
                {\vec{W}_{\psi_x}}\right)}_2^2}_{\text{Forward \SindyC{} Prediction Loss}}\\
    &+ \lambda_1 \underbrace{\norm{(\vec{x}_t, \vec{u}_t) - (\psi_x(\phi_x(\vec{x}_t; \vec{W}_{\phi_x}); 
         {\vec{W}_{\psi_x}}),
       \psi_u( \phi_u(\vec{u}_t; \vec{W}_{\phi_u}); {\vec{W}_{\psi_u}}))}_2^2}_{\text{Autoencoder Loss}}\\
    &+ \lambda_2 \underbrace{\norm{{\vec{\Xi}}}_1}_{\text{Promote Sparsity}}
    \end{aligned}\;,
\end{align*}
see also \cref{eq:AutoencoderLossFunction}. Compared to SINDy-RL and \AeSindy{}, we use the library
\emph{PySINDy} only once to create the set of dictionary functions, but not to optimize the
coefficients. Instead, the optimization of the coefficient matrix $\vec{\Xi}$ is embedded in the
general training framework of the autoencoder, allowing (a) to optimize the $\norm{\cdot}_1$-loss
directly without relying on sequential thresholding and (b) simultaneously optimize the 
coefficients of the internal dynamics model and the compression of the AE framework
(cf. \cref{fig:TikzArchitectureAutoEncoder}) using \emph{PyTorch}'s automatic 
differentiation framework.
\fi
\begin{table}[H]
	\centering
	\begin{tabular}{ccc}
	\toprule
								   & Partially observable                  & Fully observable                       \\
	\midrule
	Layer shapes (state, control)   & $(48, 8) \times (10, 4) \times (2,2)$ & $(256, 8) \times (22, 4) \times (2,2)$ \\
	$\#$Parameters: AE + $\Xi$-matrix & 1178 + 22                             & 11752 + 22                             \\
	Off-policy buffer size         & 200 & 200                                  \\
	On-policy buffer size        & 2400 & 2400                                  \\
	SINDy Polynomial degrees (state, control)        & (3,1) & (3,1)                                  \\
	SINDy dictionary size $d$        & 11 & 11                                  \\
	FOM data update frequency       & 5 \& 10                              & 5 \& 10                                \\
	Internal Epochs                & 100                                   & 100                                    \\
	Adam Learning Rate             & $10^{-3}$                             & $10^{-3}$                              \\
	Batch Size                     & 64                                    & 64                                     \\
	Activation Function            & Softmax                               & Softmax                                \\
	Loss Function                  & $\lambda_i = 1.0, i=1, 2$           & $\lambda_i = 1.0, i=1, 2$            \\
	Clip Gradient Norm             & $1.0$                                 & $1.0$                                  \\
	\bottomrule
	\end{tabular}
	\caption{Details of the \ourAlgo{} surrogate model for the \textbf{Burgers'} equation.
	We use \emph{PySINDy} \cite{kaptanoglu2021pysindy} to generate the set of dictionary
	functions.}
\end{table}

\begin{table}[H]
	\centering
	\begin{tabular}{ccc}
	\toprule
								   & 8-dim Latent space          & 4-dim Latent space         \\
	\midrule
	Layer shapes (state, control)   & $(882, 1) \times (84, 4) \times (6,2)$ &  $(882, 1) \times (52, 4) \times (3,1)$ \\
	$\#$Parameters: AE + $\Xi$-matrix & 150785 + 510                   &  93055 + 60 \\
	Off-policy buffer size         & 200 & 200 \\
	On-policy buffer size        & 2400 & 2400 \\
	SINDy Polynomial degrees (state, control)        & (3,1) & (3,1) \\
	SINDy dictionary size $d$        & 85 & 20 \\
	FOM data update frequency       & 5    & 5                        \\
	Internal Epochs                & 100       & 100                            \\
	Adam Learning Rate             & $10^{-3}$ &  $10^{-3}$                    \\
	Batch Size                     & 64                    & 64                \\
	Activation Function            & Softmax                   & Softmax            \\
	Loss Function                  & $\lambda_i = 1.0, i=1, 2$      & $\lambda_i = 1.0, i=1,2$ \\
	Clip Gradient Norm             & $1.0$                              & $1.0$   \\
	\bottomrule
	\end{tabular}
	\caption{Details of the \ourAlgo{} surrogate model for the \textbf{Navier-Stokes} equations.
	We use \emph{PySINDy} \cite{kaptanoglu2021pysindy} to generate the set of dictionary
	functions.}
	\label{tab:DetailsAETrainingNavierStokes}
\end{table}

\end{appendices}

\end{document}